\documentclass{article}

\PassOptionsToPackage{numbers, sort, compress}{natbib}
\usepackage[main, final]{neurips_2026}

\usepackage[utf8]{inputenc}
\usepackage[T1]{fontenc}
\usepackage{hyperref}
\usepackage{url}
\usepackage{booktabs}
\usepackage{amsfonts}
\usepackage{nicefrac}
\usepackage{microtype}
\usepackage{xcolor}

\usepackage{wrapfig}
\usepackage{needspace}
\usepackage{amsmath}
\usepackage{multirow}
\usepackage{array}
\usepackage{graphicx}
\usepackage{pifont}
\usepackage{algorithm}
\usepackage{algorithmic}
\usepackage[normalem]{ulem}
\definecolor{cmarkgreen}{RGB}{0,150,60}\definecolor{xmarkred}{RGB}{210,30,30}
\newcommand{\cmark}{\textcolor{cmarkgreen}{\ding{51}}}

\newcommand{\xmark}{\textcolor{xmarkred}{\ding{55}}}

\title{
  MotionGrounder: Grounded Multi-Object\\
  Motion Transfer via Diffusion Transformer%
}

\author{%
  Samuel Teodoro$^{1}$ \quad
  Yun Chen$^{1}$ \quad
  Agus Gunawan$^{1}$ \quad
  \textbf{Soo Ye Kim}$^{2}$ \\
  \textbf{Jihyong Oh}$^{3,\dagger}$ \quad
  \textbf{Munchurl Kim}$^{1,\dagger}$ \\[4pt]
  $^{1}$VICLab, KAIST \quad $^{2}$Adobe Research \quad $^{3}$CMLab, Chung-Ang University \\
  \texttt{\{sateodoro, cyruby, agusgun, mkimee\}@kaist.ac.kr} \\
  \texttt{sooyek@adobe.com} \quad
  \texttt{jihyongoh@cau.ac.kr} \\[4pt]
  {\hypersetup{pdfborder={0 0 0}}\color[RGB]{45,85,150}\url{https://kaist-viclab.github.io/motiongrounder-site/}}
}

\begin{document}

\maketitle
\addtocontents{toc}{\protect\setcounter{tocdepth}{-1}}
{\renewcommand{\thefootnote}{\fnsymbol{footnote}}\footnotetext[2]{Co-corresponding authors.}}

\begin{figure}[h]
  \centering

  \vspace{-1pt}
  \includegraphics[width=1.0\linewidth]{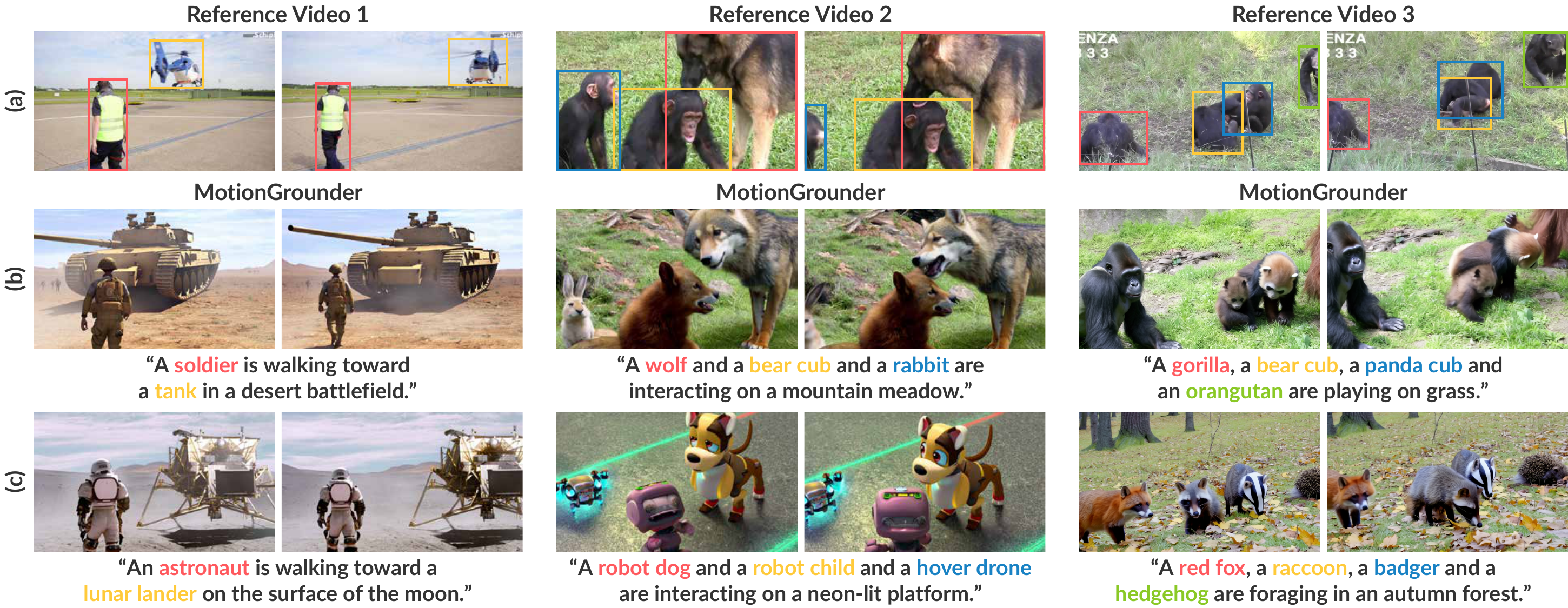}
  \vspace{-17pt}

  \caption{
    \textbf{Overview of MotionGrounder}.
    MotionGrounder transfers motion from the three reference videos in (a) to the synthesized videos in (b) and (c) with explicit object grounding, supporting structural and appearance changes in a training-free manner.
    Frames in the reference videos include color-coded bounding boxes corresponding to objects in the target captions.
  }

  \label{fig:main_figure}
\end{figure}

\begin{abstract}
Motion transfer enables controllable video generation by transferring temporal dynamics from a reference video to synthesize a new video conditioned on a target caption.
However, existing Diffusion Transformer (DiT)--based methods either transfer the motion of a single object or handle multiple objects without binding each object caption to its object, restricting fine-grained control in real-world scenes.
In this work, we introduce MotionGrounder, a \textit{training-free} DiT-based framework that is the first to jointly handle motion transfer and object grounding in multi-object scenes.
Our Flow-based Motion Signal (FMS) in MotionGrounder provides a stable motion prior for target video generation, while our Object-Caption Alignment Loss (OCAL) grounds object captions to their corresponding spatial regions.
We further propose two \textit{new} metrics: the Object Grounding Score (OGS), which jointly evaluates (i) spatial alignment between source video objects and their generated counterparts and (ii) semantic consistency between each generated object and its target caption, and the Motion Grounding Score (MGS), which evaluates whether each object moves like its source object at the correct location.
Our experiments show that MotionGrounder improves object grounding over recent baselines, with the advantage appearing once several objects are present, across quantitative, qualitative, and human evaluations.
\end{abstract}

\section{Introduction}
\label{sec:introduction}

Early text-to-video (T2V) diffusion models \mbox{%
\citep{ddpm2020ho, ldm2022rombach} }\hskip0pt%
extend U-Net-based text-to-image models with temporal layers \mbox{%
\citep{alignlatents2023blattmann, modelscope2023wang, zeroscope2023cerspense, animatediff2023guo} }\hskip0pt%
and are trained on large-scale text--video data \mbox{%
\citep{webvid2021bain}}%
. More recent models adopt diffusion transformers (DiTs) \mbox{%
\citep{dit2023peebles, cogvideox2025yang, wan2025wanopenadvancedlargescale}}%
, which better capture spatiotemporal dependencies and achieve higher visual fidelity.
However, text prompts alone offer limited control over how objects move in the generated video.

Text-guided motion transfer addresses this by transferring the motion dynamics of a source video to a target video, where the source video provides explicit motion cues and text captions define the target appearance.
Early methods \mbox{%
\citep{dmt2023yatim, moft2024xiao, motionclone2024ling, conmo2025gao, efficientmt2025cai, motionflow2026meral} }\hskip0pt%
build on U-Net T2V models \citep{modelscope2023wang, zeroscope2023cerspense}, while recent ones \mbox{%
\citep{ditflow2025pondaven, det2025shi, multimotion2025liu, ropecraft2025gokmen, followyourmotion2025ma, flowmotion2026wang} }\hskip0pt%
adopt DiTs \citep{cogvideox2025yang, wan2025wanopenadvancedlargescale} for improved visual fidelity.

\begin{wraptable}{r}{0.48\linewidth}
  \vspace{-18pt}
    \caption{
        \textbf{Comparison of motion transfer methods.} Motion signal quality is reported only for patch trajectory-based methods (PTBM).
    }
  \label{tab:motivation}
  \centering
  \fontsize{7pt}{8pt}\selectfont
  \setlength{\tabcolsep}{2pt}
    \begin{tabular}{l*{5}{c}}
    \toprule
    \multirow{2}{*}{\textbf{Methods}} & DiT- & w/o & Multi- & Obj. & \multirow{2}{*}{PTBM} \\
    & based & inv. & object & ground. & \\
    \midrule
    DMT~\citep{dmt2023yatim} & \xmark & \xmark & Limited & \xmark & \xmark \\
    MOFT~\citep{moft2024xiao} & \xmark & \cmark & Limited & \xmark & \xmark \\
    MotionClone~\citep{motionclone2024ling} & \xmark & \cmark & Limited & \xmark & \xmark \\
    MotionFlow~\citep{motionflow2026meral} & \xmark & \xmark & Limited & \xmark & \xmark \\
    ConMo~\citep{conmo2025gao} & \xmark & \xmark & \cmark & \xmark & \xmark \\
    RoPECraft~\citep{ropecraft2025gokmen} & \cmark & \cmark & Limited & \xmark & \xmark \\
    FlowMotion~\citep{flowmotion2026wang} & \cmark & \cmark & Limited & \xmark & \xmark \\
    MultiMotion~\citep{multimotion2025liu} & \cmark & \cmark & \cmark & \xmark & \cmark (Noisy) \\
    DiTFlow~\citep{ditflow2025pondaven} & \cmark & \cmark & Limited & \xmark & \cmark (Noisy) \\
    MotionGrounder (Ours) & \cmark & \cmark & \cmark & \cmark & \cmark (Stable) \\
    \bottomrule
  \end{tabular}
\end{wraptable}

Despite this progress, existing methods face limitations:
(i) most approaches, except ConMo \citep{conmo2025gao} and MultiMotion \citep{multimotion2025liu}, are restricted to single-object motion transfer despite multi-object dynamics in real videos \citep{synfmc2025shuai};
(ii) while ConMo and MultiMotion extend to multi-object scenarios via global--local motion decomposition, they lack explicit object--caption association, and ConMo relies on costly DDIM inversion \citep{ddim2022song2022}, limiting DiT compatibility; and
(iii) DiTFlow \citep{ditflow2025pondaven} and MultiMotion are zero-shot, DiT-based, inversion-free methods that improve visual fidelity but read displacement from an argmax over cross-frame attention (AMF), which is unreliable (Fig.~\mbox{%
\ref{fig:amf_vs_fms}}\hskip0pt%
), hindering precise motion control.
These limitations motivate a unified DiT-based framework that jointly enables stable motion transfer and grounded multi-object control between the spatial regions of original objects and their corresponding target captions, as summarized in Table~\ref{tab:motivation}.

Motivated by these observations, we introduce \textbf{MotionGrounder}, a novel \textit{training-free} DiT-based framework that unifies motion transfer and semantic grounding for multi-object scenes.
Fig.~\ref{fig:main_figure} shows qualitative results of our MotionGrounder, showing stable motion transfer and grounded multi-object results.
MotionGrounder transfers motion dynamics from a source video to a text-defined target video to be generated, while ensuring each generated object aligns with its corresponding target caption, without additional training, keeping the pretrained DiT frozen and optimizing only its latent.
To achieve this, we propose (i) a Flow-based Motion Signal (FMS) that derives its motion target from dense optical flow on the source frames and provides stable motion guidance during generation and (ii) an Object-Caption Alignment Loss (OCAL) that grounds each object in the caption to a spatial object region, so that each source object's motion is attributed to the correct target object.
We evaluate grounding performance with two novel metrics, the Object Grounding Score (OGS) and the Motion Grounding Score (MGS). Our grounding advantage appears once several objects are present (Table~\mbox{%
\ref{tab:per_count}}%
). Our main contributions are:
\begin{itemize}
    \item We propose \textbf{MotionGrounder}, a \textit{training-free} DiT-based framework that \textit{firstly} handles joint motion transfer and grounding for multi-object scenes;
    \item We design a \textbf{Flow-based Motion Signal (FMS)} to provide a more stable motion guidance in the latent space during video generation;
    \item We introduce an \textbf{Object-Caption Alignment Loss (OCAL)} to enforce spatial grounding between object captions and their corresponding object regions.
    \item We propose an \textbf{Object Grounding Score (OGS)} to evaluate spatial alignment between original and generated objects, and their alignment with their corresponding target captions, and a \textbf{Motion Grounding Score (MGS)} to evaluate whether each object moves like its source object at the correct location.
\end{itemize}

\section{Related Works}
\label{sec:related_works}

\noindent
\textbf{Text-to-video generation models.}
The success of diffusion models  \citep{deepunsupervisedlearning2015sohldickstein, generativemodeling2020song, ddpm2020ho, scorebasedgenerativemodeling2021song} in text-to-image (T2I) generation \citep{zeroshottexttoimagegeneration2021ramesh, ldm2022rombach, imagen2022saharia} has inspired their extension to text-to-video (T2V) synthesis \citep{imagenvideo2022ho, cogvideo2022hong, alignlatents2023blattmann, cogvideox2025yang}.
Early T2V models \mbox{%
\citep{modelscope2023wang, zeroscope2023cerspense, animatediff2023guo, controlvideo2023zhang, videocrafter22024chen} }\hskip0pt%
start from a pretrained T2I model \mbox{%
\citep{controlnet2023zhang}}%
, typically a U-Net denoiser \mbox{%
\citep{unet2015ronneberger} }\hskip0pt%
trained on large-scale image datasets \citep{imagenet2009deng, laion5b2022schuhmann}, and add temporal layers so that it can model motion across frames.
More recent T2V models \mbox{%
\citep{sora2024brooks, gentron2024chen, latte2025ma} }\hskip0pt%
replace the U-Net with a diffusion transformer (DiT) \citep{dit2023peebles}, which captures{} spatiotemporal dependencies more effectively and scales to higher visual quality.
Our method builds on such a DiT-based model and extends it to grounded motion transfer with multiple objects.

\noindent
\textbf{Zero-shot text-guided motion transfer.}
Text-guided motion transfer synthesizes new videos that follow the motion of a source video and the appearance given by text.
Recent zero-shot diffusion-based methods model motion implicitly at inference:
DMT \mbox{%
\citep{dmt2023yatim} }\hskip0pt%
via inter-frame feature discrepancies,
MOFT \mbox{%
\citep{moft2024xiao} }\hskip0pt%
by removing content-related information and irrelevant channels from video features,
MotionClone \mbox{%
\citep{motionclone2024ling} }\hskip0pt%
via sparse temporal attention weights,
MotionFlow \mbox{%
\citep{motionflow2026meral} }\hskip0pt%
via DDIM inversion and cross-attention losses on attention-thresholded object regions,
DiTFlow \mbox{%
\citep{ditflow2025pondaven} }\hskip0pt%
via Attention Motion Flow (AMF) derived from the cross-frame attention of DiTs,
RoPECraft \mbox{%
\citep{ropecraft2025gokmen} }\hskip0pt%
by warping rotary positional embeddings (RoPE) with dense optical flow,
and FlowMotion \mbox{%
\citep{flowmotion2026wang} }\hskip0pt%
by aligning latent predictions and frame-wise differences of flow-matching T2V models.
However, these works offer limited support for multiple objects.
ConMo \mbox{%
\citep{conmo2025gao} }\hskip0pt%
and MultiMotion \mbox{%
\citep{multimotion2025liu} }\hskip0pt%
target multi-object scenes: ConMo separates camera from object motion, and MultiMotion decomposes motion per object with segmentation masks through a mask-aware AMF, but neither binds captions to objects, so the model, not the user, decides which generated object inherits which caption.
Like DiTFlow, we enforce semantic correspondences in the attention maps of a DiT, but our optical flow--based motion signal is computed outside the generator, avoiding the unreliable correspondences of AMF. RoPECraft also uses optical flow, but to warp RoPE; we use it to build a motion signal for attention-based optimization.
To address the lack of semantic grounding, our explicit object--caption grounding loss binds each object caption to its own mask region, letting users set the correspondence.

\noindent
\textbf{Grounding via attention manipulation.}
Several works in image and video generation and editing achieve object grounding via attention manipulation.
Methods such as \citep{prompttoprompt2022hertz, densediffusion2023kim, attendexcite2023chefer, diptychprompting2025shin} manipulate self- and cross-attention to control spatial alignment between text tokens and visual regions.
Attention Refocusing \citep{attentionrefocusing2023phung} refines noised samples by optimizing losses to steer attention toward a target spatial layout.
In the video domain, methods \citep{videop2p2023liu, videograin2025yang} extend these techniques temporally, enabling object-level grounding and editing across frames.
Recently, DiTCtrl \citep{ditctrl2025cai} enables video editing by leveraging the U-Net--like behavior of a multi-modal DiT's 3D attention for fine-grained, training-free control.
Similarly, we manipulate 3D attention maps to jointly enforce motion transfer and object grounding, which, to the best of our knowledge, has not yet been explored for DiTs.

\section{Method}
\label{sec:method}

\subsection{Preliminary}
\label{sec:preliminary}

Text-to-video (T2V) diffusion models generate a video $V_{T}$ by iteratively denoising a sampled Gaussian noise $z_{t} \sim \mathcal{N}(0,I)$ to $z_{0} \in \mathbb{R}^{F \times C \times W \times H}$ using a denoising network $\epsilon_{\theta}$ over $t \in [0, T]$ steps \citep{ddpm2020ho}, where $F$, $C$, $H$, $W$ denote the number of frames, latent channels, height, and width, respectively.
Each denoising step is defined as:%
\begin{equation}
    z_{t-1} = \mathcal{S}(z_{t}, \epsilon_{\theta}(z_{t}, c, t))
\end{equation}%
where $c$ is a prompt describing the output video's appearance and motion, and $\mathcal{S}$ is the noise removal following a specific scheduler \citep{ddpm2020ho, ddim2022song2022}.
For more efficient synthesis, Latent Diffusion Models (LDMs) \citep{ldm2022rombach} apply diffusion in a compact latent space defined by a pre-trained autoencoder \citep{vae2022kingma} with encoder $\mathcal{E}$ and decoder $\mathcal{D}$.
The denoising network $\epsilon_{\theta}$ is usually a U-Net \citep{unet2015ronneberger} and the prompt $c$ is injected into the network via cross-attention \citep{ldm2022rombach} or direct concatenation with the video latent representation \citep{cogvideox2025yang}.
The target video $V_{T}$ is obtained by decoding $z_{0}$ as $V_{T}=\mathcal{D}(z_{0})$.

Diffusion Transformers (DiTs) \citep{dit2023peebles} instead use a transformer denoiser that encodes noisy latents as a sequence of patch tokens, similar to Vision Transformers \citep{vit2021dosovitskiy}.
The latent is split into $P \times P$ patches embedded as $d$-dimensional tokens, flattened over space and time into a sequence of length $F \cdot \frac{H}{P} \cdot \frac{W}{P}$, with positional embeddings $\rho$ conditioning the denoising process $\epsilon_{\theta}(z_t, c, t, \rho)$.

\begin{figure}[t]
    \centering
    \includegraphics[width=1.0\linewidth]{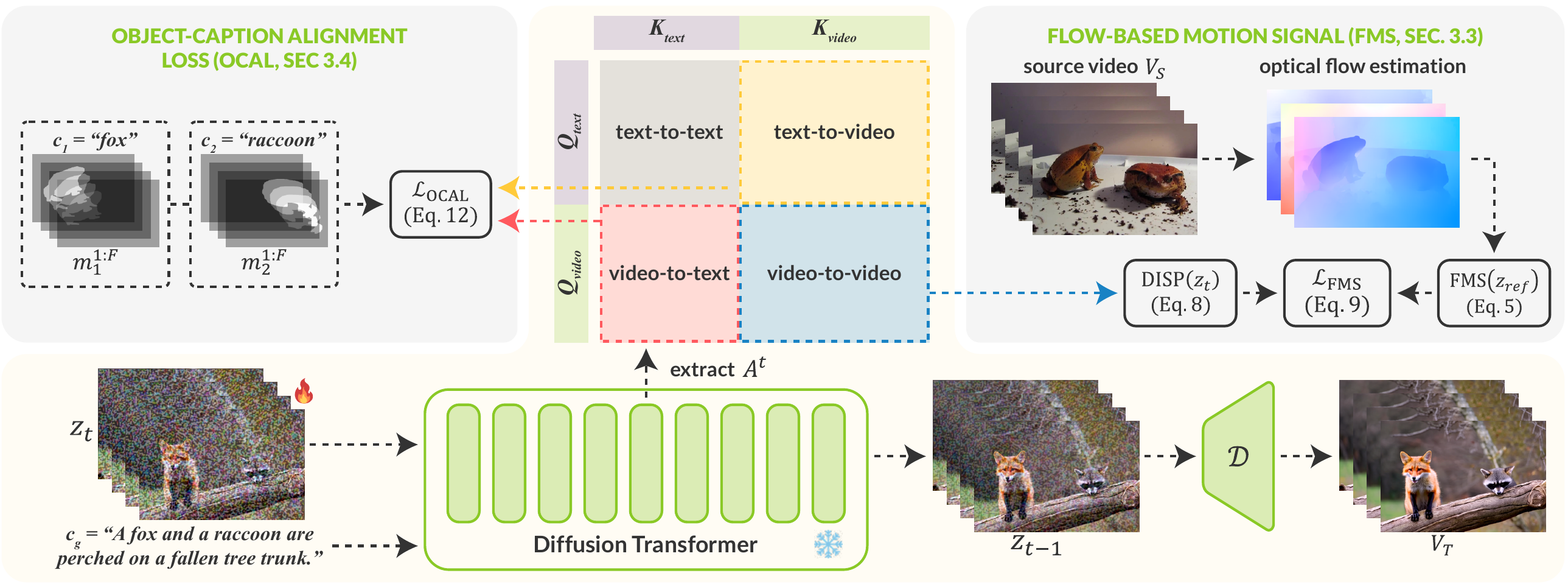}
    \vspace{-15pt}
    \caption{
        \textbf{Overall framework of MotionGrounder.}
        Given a source video $V_S$ with $N$ objects, global caption $c_g$, object captions $\{c_i\}_{i=1}^N$, and corresponding object masks $\{m_i^{1:F}\}_{i=1}^N$, MotionGrounder transfers motion dynamics to a text-defined target video $V_T$.
        A Flow-based Motion Signal (FMS, Sec.~\ref{sec:fms}) provides stable motion guidance, while the Object-Caption Alignment Loss (OCAL, Sec.~\ref{sec:ocal}) enforces spatial grounding between each object caption and its designated object region.
    }
    \label{fig:overallframework}
    \vspace{-12pt}
\end{figure}

\subsection{Overall Framework}

Fig.~\ref{fig:overallframework} illustrates the overall framework of our MotionGrounder.
Given (i) a source video $V_S$ with $F$ frames and $N$ objects, (ii) a global caption $c_g$ containing the object captions $\{c_i\}_{i=1}^N$ (i.e., $c_i \in c_g$), and (iii) their corresponding object masks $\{M_i^{1:F}\}_{i=1}^N$, where $M_i^{1:F} \in \mathbb{R}^{F \times H \times W}$ denotes the masks of the $i$-th object from the first to the $F$-th frame, MotionGrounder transfers motion dynamics from $V_S$ to a target video $V_T$ while grounding each object caption $c_i$ to its designated spatial region defined by $M_i^{1:F}$.
We linearly sample $J$ frames from $V_S$, where $J{=}F/4$ corresponds to the number of latent frames after VAE \citep{cogvideox2025yang} temporal compression.
We estimate their optical flows using GMFlow \citep{gmflow2022xu} and construct a Flow-based Motion Signal (FMS, Sec.~\ref{sec:fms}).
During denoising, we align the motion of $V_{T}$ with that of $V_{S}$ via the FMS, and enforce spatial alignment between each object caption $c_i$ and object region designated by $M_i^{1:F}$ using our Object-Caption Alignment Loss (OCAL, Sec.~\ref{sec:ocal}).
The pretrained DiT stays frozen; both losses are backpropagated only into the noisy latent $z_t$ (Algorithm~\mbox{%
\ref{alg:motiongrounder_inference}}%
), so neither acts as a conditioning signal.

\subsection{Flow-based Motion Signal (FMS)}
\label{sec:fms}

\begin{figure}[!h]
    \centering
    \includegraphics[width=1.0\linewidth]{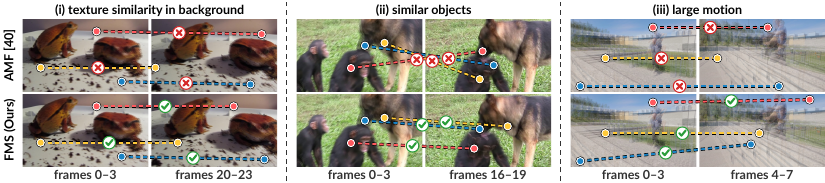}
\vspace{-18.4pt}
    \caption{
        \textbf{Attention Motion Flow (AMF) \citep{ditflow2025pondaven} vs.\ our Flow-based Motion Signal (FMS, Sec.~\ref{sec:fms}).} Each panel overlays the four input frames of one latent frame, the same in both rows. Dashed lines link three source patches to their estimates, crossed if wrong. AMF (i) drifts on texture, (ii) swaps similar objects, and (iii) breaks under large motion, while our FMS stays stable.
    }
    \label{fig:amf_vs_fms}
    \vspace{-4pt}
\end{figure}

\begin{wrapfigure}{r}{0.56\linewidth}
    \vspace{6.4pt}
    \centering
    \includegraphics[width=1.0\linewidth]{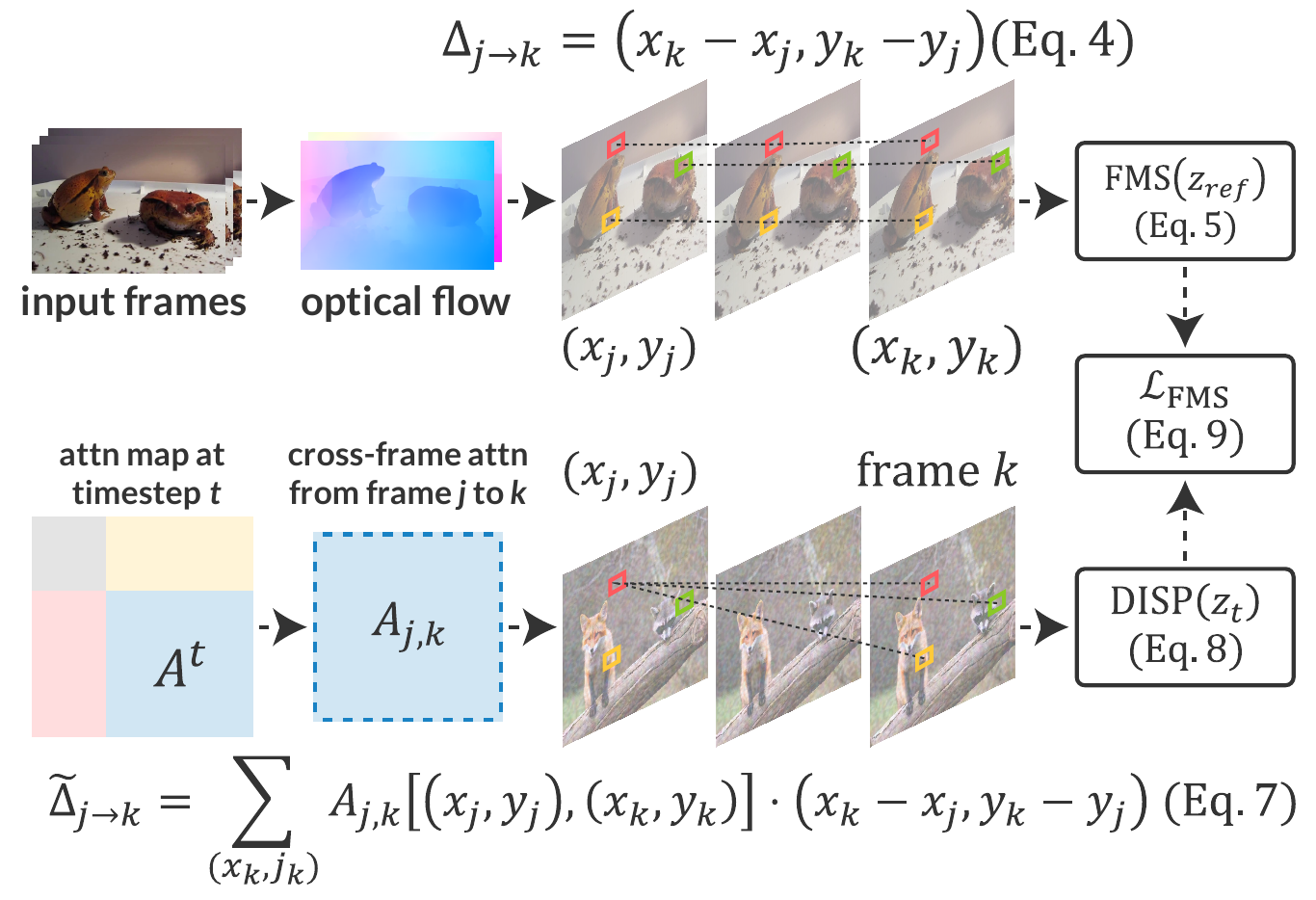}
    \vspace{-12pt}
    \caption{
        \textbf{Flow-based Motion Signal (FMS, Sec.~\ref{sec:fms}).}
        FMS constructs stable latent space patch trajectories from optical flows estimated on sampled source video frames, and uses the resulting displacements to supervise motion transfer.
    }
    \label{fig:fms}
    \vspace{-8.4pt}
\end{wrapfigure}

\textbf{Optical flow estimation.}
Attention Motion Flow (AMF) \citep{ditflow2025pondaven} reads displacement from an argmax over cross-frame attention, a signal never trained to match points, so it is unreliable under repeated texture, similar objects and large motion (Fig.~\mbox{%
\ref{fig:amf_vs_fms}}%
, Appx.~\mbox{%
\ref{app:fms_motivation}}\hskip0pt%
).
To obtain a more stable motion prior, we introduce the Flow-based Motion Signal (FMS), inspired by FLATTEN \citep{flatten2024cong}.
Fig. \ref{fig:fms} illustrates the concept of our FMS.
The FMS (i) uniformly samples $J$ frames from the input video, (ii) computes dense optical flows between consecutive frames using GMFlow \citep{gmflow2022xu}, and (iii) downsamples the resulting displacement fields $(f_x, f_y)$ to the latent resolution, resulting in $(\hat{f}_x, \hat{f}_y)$.
A patch at $(x_j, y_j)$ in frame $j$ is propagated to frame $j{+}1$ in the latent space via%
\begin{equation}
    (x_{j+1}, y_{j+1}) =
    \big(x_j + \hat{f}_x(x_j, y_j),\;
         y_j + \hat{f}_y(x_j, y_j)\big).
\end{equation}

\noindent
\textbf{Trajectory construction.}
Patch trajectories are formed by propagating latent space patch locations across frames using the downsampled displacement fields. For a patch at $(x_j, y_j)$, this yields a temporally linked trajectory

{\begin{equation}
    \text{traj}(x_j, y_j) = \{ (x_k, y_k) \}_{k=1}^{J},
\end{equation}
}

where $J{=}F/4$ is the number of latent frames.
We form new trajectories for newly observed patches and allow trajectory duplication, unlike FLATTEN, as latent-space compression may map multiple pixels to the same spatial location.

\textbf{Motion signal.}
Given a patch trajectory $\text{traj}(x_j, y_j)$, the displacement from anchor frame $j$ to frame $k$ is defined as%
\begin{equation}
    \Delta_{j \rightarrow k} = (x_k - x_j,\; y_k - y_j),
\end{equation}%
where $(x_k, y_k) \in \text{traj}(x_j, y_j)$.
The FMS aggregates these displacements across all patches:%
\begin{equation}
    \text{FMS}(z_{\text{ref}}) =
    \{ \Delta_{j \rightarrow k} \mid \forall (x_j, y_j),\; j,k \in \{1,\dots,J\} \}.
\end{equation}%
Unlike FLATTEN, we use FMS as a supervised motion signal for optimization rather than for attention guidance.
FMS is thus a geometric target: flow is estimated in pixel space on the clean source frames, outside the generator, and applied on the latent grid, so it inherits no attention errors and drives different DiT backbones (Appx.~\mbox{%
\ref{app:generalization}}%
); Appx.~\mbox{%
\ref{app:flow_estimator} }\hskip0pt%
analyzes the flow estimator and oracle flow.

\noindent
\textbf{Motion guidance.}
At denoising timestep $t$, cross-frame attention is computed as:%
\begin{equation}
    \label{eqn:cross_frame_attn}
    A_{j,k} = \sigma ( \tau \cdot Q_j K_k^{\top} / \sqrt{d} ),
\end{equation}%
where $Q_j$ and $K_k$ are the query and key matrices of frames $j$ and $k$, respectively, $\sigma$ is the softmax function, $\tau$ is the temperature, and $d$ is the feature dimension.
Following \citep{ditflow2025pondaven}, we compute displacements via attention-weighted coordinate differences to preserve gradients:%
\begin{equation}
    \label{eqn:amf_disp}
    \textstyle \tilde{\Delta}_{j \rightarrow k} = \sum_{(x_k, y_k)} A_{j,k}\!\left[(x_j, y_j), (x_k, y_k)\right] \cdot (x_k - x_j,\; y_k - y_j),
\end{equation}
where $A_{j,k}\!\left[(x_j, y_j), (x_k, y_k)\right]$ denotes the attention weight between patches at spatial locations $(x_j, y_j)$ and $(x_k, y_k)$.
The displacement field over all patches is defined as%
\begin{equation}
    \text{DISP}(z_t)
    =
    \{ \tilde{\Delta}_{j \rightarrow k}
    \mid
    \forall (x_j, y_j),\;
    j,k \in \{1,\dots,J\} \}.
\end{equation}%
Motion transfer is supervised by minimizing%
\begin{equation}
    \mathcal{L}_{\text{FMS}}
    =
    \lVert \text{FMS}(z_{\text{ref}}) - \text{DISP}(z_t) \rVert_2^2.
\end{equation}%
\subsection{Object-Caption Alignment Loss (OCAL)}
\label{sec:ocal}

\begin{figure}[!h]
    \centering
    \includegraphics[width=1.0\linewidth]{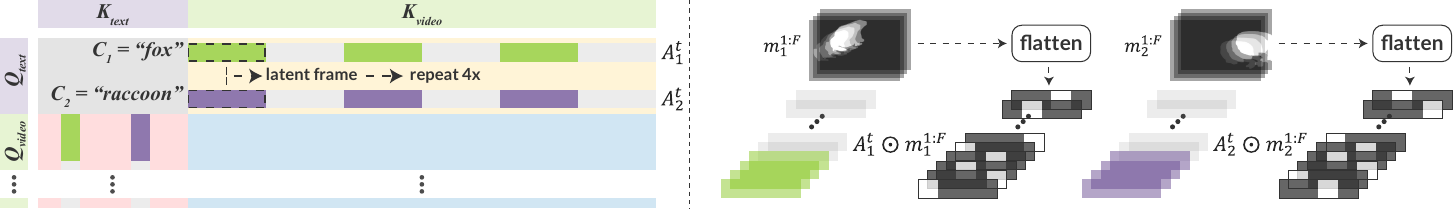}
    \vspace{-16.6pt}
    \caption{
        \textbf{Object-Caption Alignment Loss (OCAL, Sec.~\ref{sec:ocal}).}
        OCAL aggregates object-specific attention maps and aligns them with their corresponding object masks to enforce spatial grounding of each object caption during generation.
    }
    \label{fig:ocal}
    \vspace{-5.3pt}
\end{figure}

Since real-world videos often contain multiple objects, the model must be guided on where to generate each object in the video.
We achieve this using our Object-Caption Alignment Loss (OCAL) at inference.
Fig.~\mbox{%
\ref{fig:ocal} }\hskip0pt%
visualizes OCAL.
For each object $i$ with caption $c_i \in c_g$ and downsampled object masks $m_i^{1:F} \in \mathbb{R}^{F \times (H/16) \times (W/16)}$ from $M_{i}^{1:F}$, we extract the 3D attention map $A^t$ from transformer block $B$ at diffusion timestep $t$.
We aggregate the text-to-video and video-to-text attention maps corresponding to the tokens of $c_i$, denoted as $A_i^t \in \mathbb{R}^{(F/4) \times (H/16) \times (W/16)}$.
Due to the temporal compression in the VAE encoder, we repeat the attention values associated with each latent frame along the temporal dimension by a factor of four, which yields ${A}_{i}^{t} \rightarrow \tilde{A}_{i}^{t} \in \mathbb{R}^{F \times (H/16) \times (W/16)}$.
This repetition tightens each target from the union of an object's masks over four frames to its per-frame mask, aligning $\tilde{A}_i^t$ with $m_i^{1:F}$ for precise object placement.

\noindent
\textbf{OCAL objective.}
OCAL encourages attention values within each object region to be maximized via the foreground loss as:%
\begin{equation}
    \label{eqn:ocal_fg}
    \mathcal{L}_{\text{FG}}^{(i)} = \big( 1 - \text{sum}( \tilde{A}_i^t \odot m_i^{1:F} ) \big)^2,
\end{equation}
where $\odot$ is the element-wise multiplication.
The attention outside each object region is simultaneously suppressed by the background loss as:%
\begin{equation}
    \label{eqn:ocal_bg}
    \mathcal{L}_{\text{BG}}^{(i)} = \big( \text{sum}( \tilde{A}_i^t \odot (1 - m_i^{1:F}) ) \big)^2.
\end{equation}
The final OCAL objective is defined as:%
\begin{equation}
    \label{eqn:ocal}
    \textstyle \mathcal{L}_{\text{OCAL}} = \frac{1}{N} \sum_{i=1}^{N} \lambda_{size}^{(i)} \cdot \big( \mathcal{L}_{\text{FG}}^{(i)} + \mathcal{L}_{\text{BG}}^{(i)} \big),
\end{equation}
where the size regularizer $\lambda_{size}^{(i)}=1-\sum m_i^{1:F} / |m_{i}^{1:F}|$ serves as a size-aware weight, encouraging proper generation of smaller objects.
It is one scalar per object per clip, so it balances object sizes but does not track size changes over time, which the per-frame masks handle (Appx.~\mbox{%
\ref{app:size_reg_zoom}}%
).

\noindent
\textbf{Overall objective.} The total objective at timestep $t$ is:%
\begin{equation}
\mathcal{L}_{t}
=
\lambda_{\text{FMS}} \cdot \mathcal{L}_{\text{FMS}} +
\lambda_{\text{OCAL}} \cdot \mathcal{L}_{\text{OCAL}},
\end{equation}%
where $\lambda_{\text{FMS}}$ and $\lambda_{\text{OCAL}}$ control the strengths of motion supervision and object-caption alignment, respectively.
The terms are complementary: FMS carries motion geometry from optical flow, and OCAL, a flow-independent attention term, binds each motion to its captioned object. Neither constrains shape: FMS supervises only displacement, and OCAL is a coarse spatial prior that never compares pixels with the source (Appx.~\mbox{%
\ref{app:shape}}%
).
Algorithm~\ref{alg:motiongrounder_inference} (Appx.~\mbox{%
\ref{app:algorithm}}%
) summarizes the procedure.

\section{Experiments}
\label{sec:experiments}

\textbf{Dataset}.
Due to the multi-object nature of our task, we construct a dedicated evaluation dataset of 52 high-motion videos: 19 from DAVIS \citep{davis2016Perazzi}, 29 from YouTube-VOS \citep{youtubevos2018xu}, and 4 from ConMo \citep{conmo2025gao}.
We rank candidates by summed per-frame flow magnitude, keep those in which every captioned object has its own per-frame mask, and take the top of each object-count bucket, so object count is controlled (Appx.~\mbox{%
\ref{app:dataset_selection}}%
).
Near-static sources leave little motion to transfer, so favoring high motion also makes the benchmark harder (Appx.~\mbox{%
\ref{app:near_static}}%
).
Object counts are 21 videos with $N{=}1$, 16 with $N{=}2$, 10 with $N{=}3$, and 5 with $N{\geq}4$, up to eight.
For each video, we uniformly sample 24 frames and center-crop them to $480 \times 720$ pixels.

Since the original videos lack captions, we describe each with CogVLM2-Caption \citep{cogvlm22024hong}, summarize it with GPT-5 \citep{gpt52023openai} into \texttt{<subject>} \texttt{<verb>} \texttt{<scene>}, manually tag each object $i$ in the \texttt{<subject>} to bind it to its masks $M_{i}^{1:F}$, and form the target caption $c_g$ by editing subject, verb, and scene.
Following \citep{ditflow2025pondaven}, we evaluate under three prompt categories.
\textit{Caption} describes the source video, \textit{Subject} changes the objects but keeps the background, and \textit{Scene} specifies an entirely new scene (examples in Fig.~\mbox{%
\ref{fig:supp_sample_frames}}%
, Appx.~\mbox{%
\ref{app:dataset_samples}}%
).
We also test on a benchmark we did not design, DiTFlow's DAVIS subset with its own prompts and protocol (43 sequences, 32 outside ours; Table~\mbox{%
\ref{tab:davis}}%
, Appx.~\mbox{%
\ref{app:davis}}%
).
We will release the dataset (details in Appx.~\mbox{%
\ref{app:dataset}}%
).

\textbf{Object Grounding Score (OGS).}
Motion transfer is commonly evaluated with a CLIP \mbox{%
\citep{clip2021radford} }\hskip0pt%
similarity between the frames and the caption, which is global and never checks whether the right object appears in the right place.
We therefore propose the Object Grounding Score (OGS), which combines two per-object factors.
The IoU $s_{\text{IoU}}$ is the overlap between the generated object's region and the source object's region, so it measures object localization.
The local CLIP similarity $s_{\text{CLIP}}$, adopted from InstanceDiffusion \mbox{%
\citep{instancediffusion2024wang}}%
, is the similarity between the cropped generated object and its object caption $c_{i}$, so it measures object--caption association.
OGS requires both jointly for each object, which no global CLIP score, local CLIP score, or mask IoU alone captures.
Since a generated object may move and change shape, we localize it with Grounding DINO \mbox{%
\citep{groundingdino2024liu} }\hskip0pt%
and SAM2 \mbox{%
\citep{sam22024ravi} }\hskip0pt%
instead of its source mask.
With $\text{OGS}_{i,f} = s_{\text{IoU}}^{i,f} \cdot s_{\text{CLIP}}^{i,f}$ for object $i$ in frame $f$, the score of a video is
\begin{equation}
    \label{eqn:OGS}
    \textstyle \text{OGS} = \frac{1}{F} \sum_{f=1}^{F} \big( \frac{1}{N_f} \sum_{i=1}^{N_f} \text{OGS}_{i,f} \big),
\end{equation}
where $F$ is the number of frames and $N_f$ the number of objects in frame $f$.

\textbf{Motion Grounding Score (MGS).}
Motion Fidelity \mbox{%
\citep{dmt2023yatim} }\hskip0pt%
compares tracklets over the whole frame, so it misses motion misattribution, where one object moves with another's motion: the right trajectories sit on the wrong objects, and MF stays high.
We therefore propose the Motion Grounding Score (MGS), which measures whether each object moves like its source and does so in the right location.
For object $i$, we track points seeded inside its source mask and inside its detected region (as for OGS), and compare the two sets with the MF similarity, giving $s_{\text{MF}}^{i} \in [-1, 1]$.
On its own this says nothing about location, so we map $s_{\text{MF}}^{i}$ to $[0, 1]$ and weight it by the object's IoU in each frame, $\text{MGS}_{i,f} = s_{\text{IoU}}^{i,f} \cdot (1 + s_{\text{MF}}^{i})/2$, and aggregate it as OGS in Eq.~\mbox{%
\ref{eqn:OGS}}%
,
\begin{equation}
    \label{eqn:MGS}
    \textstyle \text{MGS} = \frac{1}{F} \sum_{f=1}^{F} \big( \frac{1}{N_f} \sum_{i=1}^{N_f} \text{MGS}_{i,f} \big).
\end{equation}
See Appx.~\mbox{%
\ref{app:ogs} }\hskip0pt%
for details of both OGS and MGS.

\begin{figure}[t]
    \centering
    \includegraphics[width=1.0\linewidth]{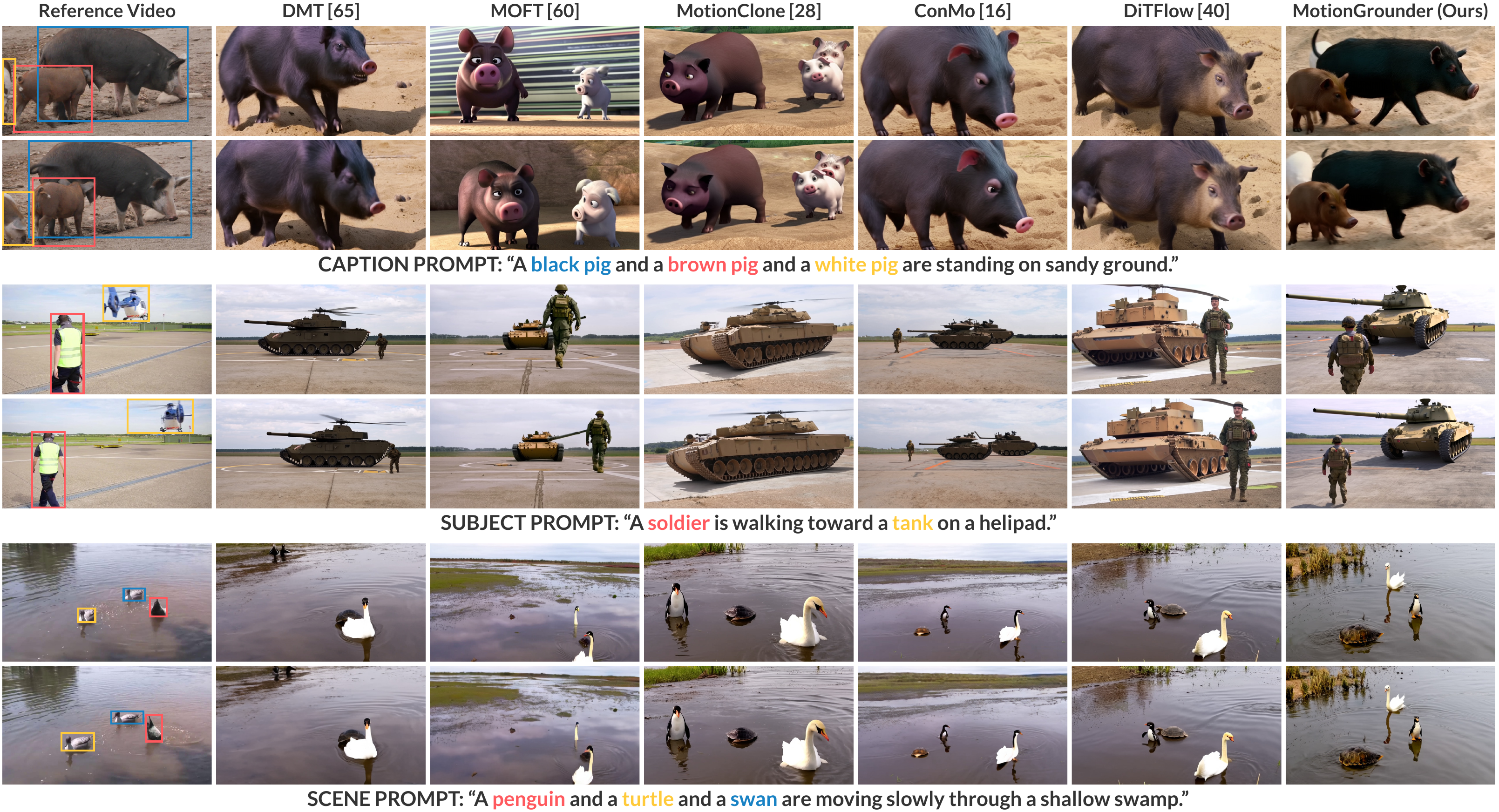}
    \vspace{-18pt}
    \caption{
        \textbf{Qualitative comparison.}
        For clarity, we show color-coded bounding boxes inferred from object masks, where each color corresponds to an object in the caption.
        While all other methods suffer from \textit{spatial misalignment} and \textit{motion misattribution}, our MotionGrounder generates correct objects, preserves spatial alignment, and maintains object-specific motion across all scenarios.
    }
    \label{fig:main_qualitative_comparison}
    \vspace{-12.7pt}
\end{figure}

\textbf{Evaluation metrics.}
For quantitative evaluation, we use eight metrics:
\textbf{(i) Motion Fidelity (MF)} \citep{dmt2023yatim}, which compares source and generated tracklets extracted with CoTracker \citep{cotracker2024karaev};
\textbf{(ii) Intersection-over-Union (IoU)} and \textbf{(iii) Local Textual Alignment (LTA)} \citep{instancediffusion2024wang}, the per-object IoU and local CLIP similarity used in OGS;
\textbf{(iv) our OGS}; \textbf{(v) our MGS};
\textbf{(vi) Global Textual Alignment (GTA)}, the CLIP score between generated frames and the global caption $c_g$;
and \textbf{(vii) CLIP} and \textbf{(viii) DINO Temporal Consistency (CTC, DTC)}, the cosine similarity of successive frame embeddings from CLIP \citep{clip2021radford} and DINO \citep{dinov22024oquab2024}, respectively.

\textbf{Baselines}.
We compare MotionGrounder against five zero-shot, latent optimization-based baselines:
\textbf{DMT} \citep{dmt2023yatim},
\textbf{MOFT} \citep{moft2024xiao},
\textbf{MotionClone} \citep{motionclone2024ling},
\textbf{ConMo} \citep{conmo2025gao}, and
\textbf{DiTFlow} \citep{ditflow2025pondaven}.
All methods use CogVideoX-5B \citep{cogvideox2025yang} as the backbone in the main paper.
Results on CogVideoX-2B and Wan2.1 \citep{wan2025wanopenadvancedlargescale} (Table~\mbox{%
\ref{tab:generalization}}%
, Appx.~\mbox{%
\ref{app:generalization}}%
) show that our method is model-agnostic.
Following DiTFlow \citep{ditflow2025pondaven}, we replace DDIM inversion with KV injection for DMT and ConMo, and fuse ConMo's object masks every four frames to match the VAE's temporal compression.

\textbf{Implementation details}.
For fairness, all methods use 50 denoising steps and the Adam \citep{adam2017kingma} optimizer, with linearly decayed learning rate (0.002 $\rightarrow$ 0.001) following \citep{dmt2023yatim}.
For MotionGrounder, FMS and OCAL act at transformer blocks $B{=}20$ and $B{=}10$, respectively, during the first 30\% of denoising, with 5 optimization steps per denoising step, $\lambda_{\text{FMS}}{=}1.25$, $\lambda_{\text{OCAL}}{=}1.00$, and $\tau{=}2$ in Eq.~\ref{eqn:cross_frame_attn}.
All experiments run on an NVIDIA A6000 GPU.
Optimal baseline hyperparameters are in Appx.~\mbox{%
\ref{app:hyperparameters}}%
, and runtime and memory are compared in Table~\mbox{%
\ref{tab:runtime_and_memory} }\hskip0pt%
(Appx.~\mbox{%
\ref{app:runtime}}%
).

\subsection{Qualitative evaluation}

Fig.~\ref{fig:main_qualitative_comparison} presents qualitative comparisons across different prompt settings, where each color-coded bounding box inferred from object masks corresponds to an object in the global caption.
In the \textit{\textbf{Caption}} and \textit{\textbf{Subject}} settings, prior methods do not consistently align generated objects with the objects in the reference video.
For example, under the \textit{\textbf{Subject}} prompt, the soldier is misplaced relative to the person in the reference video.
Prior methods also misattribute motion, assigning one object's motion to another (e.g., the soldier follows the motion of the helicopter, as in MOFT and DiTFlow).
This stems from lacking an explicit object--caption alignment, so their models associate objects with incorrect motion signals.
Under the \textit{\textbf{Scene}} prompt, DMT and MOFT fail to generate all specified objects (DMT fuses the turtle into the swan), while MotionClone, ConMo, and DiTFlow generate all three but misplace them (e.g., MotionClone and DiTFlow put the penguin on the left instead of the right).
In contrast, our MotionGrounder (i) generates all target objects, (ii) places them at their intended regions, and (iii) assigns correct object-specific motion, demonstrating robust object grounding across all scenarios.
FMS and OCAL do not strictly enforce the source shapes, as the large shape changes in Fig.~\ref{fig:main_qualitative_comparison} show: FMS supervises displacement only, OCAL is a coarse prior on the $30 \times 45$ attention grid, and guidance stops after the first 15 of 50 denoising steps, leaving the model free to form a suitable shape (Appx.~\mbox{%
\ref{app:shape}}%
).
Fig.~\mbox{%
\ref{fig:supp_more_qualitative} }\hskip0pt%
(Appx.~\mbox{%
\ref{app:more_qualitative}}%
), Fig.~\mbox{%
\ref{fig:supp_controllability_demo} }\hskip0pt%
(Appx.~\mbox{%
\ref{app:controllability}}%
), and Fig.~\mbox{%
\ref{fig:supp_limitations} }\hskip0pt%
(Appx.~\mbox{%
\ref{app:limitation}}%
) give more comparisons, a controllability demo, and the point where each method fails as objects are added, respectively.

\begin{table}[t]
  \caption{
        \textbf{Quantitative comparison} averaged over the \textit{Caption}, \textit{Subject}, and \textit{Scene} settings, split by the number of objects $N$ in the source video and, in the last block, over \textit{\textbf{All videos}} (number of videos in parentheses, on the left). Our MotionGrounder achieves the strongest performance overall, attaining the best and second best results on the majority of metrics. Best results in each block are in \textbf{bold} and second best are \underline{underlined}.
    }
  \label{tab:main_quantitative_comparison}
  \label{tab:per_count}
  \centering
  \fontsize{7pt}{8pt}\selectfont
  \setlength{\tabcolsep}{4.6pt}\renewcommand{\arraystretch}{1.0}
  \begin{tabular}{cl*{8}{c}}
    \toprule
    & \textbf{Methods} & MF $\uparrow$ & IoU $\uparrow$ & LTA $\uparrow$ & OGS $\uparrow$ & MGS $\uparrow$ & GTA $\uparrow$ & CTC $\uparrow$ & DTC $\uparrow$ \\
    \midrule
    \multirow{6}{*}{\rotatebox[origin=c]{90}{\shortstack{\textbf{$N{=}1$}\\(21)}}} & DMT \citep{dmt2023yatim} & \underline{0.6429} & \textbf{0.2553} & 0.2672 & \underline{0.0725} & \underline{0.1920} & 0.3126 & 0.9323 & 0.8817 \\
     & MOFT \citep{moft2024xiao} & 0.5195 & 0.2294 & 0.2669 & 0.0657 & 0.1591 & 0.3139 & 0.9381 & 0.8906 \\
     & MotionClone \citep{motionclone2024ling} & 0.4961 & 0.2475 & 0.2682 & 0.0714 & 0.1689 & 0.3064 & \underline{0.9384} & \underline{0.9287} \\
     & ConMo \citep{conmo2025gao} & 0.6255 & 0.2343 & 0.2690 & 0.0679 & 0.1708 & 0.3163 & 0.9343 & 0.8861 \\
     & DiTFlow \citep{ditflow2025pondaven} & 0.6163 & 0.2243 & \underline{0.2692} & 0.0644 & 0.1589 & \textbf{0.3188} & 0.9376 & 0.9132 \\
     & MotionGrounder (Ours) & \textbf{0.6607} & \underline{0.2487} & \textbf{0.2704} & \textbf{0.0729} & \textbf{0.1965} & \underline{0.3164} & \textbf{0.9412} & \textbf{0.9318} \\
    \midrule
    \multirow{6}{*}{\rotatebox[origin=c]{90}{\shortstack{\textbf{$N{=}2$}\\(16)}}} & DMT \citep{dmt2023yatim} & 0.7064 & 0.2139 & \underline{0.2693} & 0.0582 & 0.1655 & \underline{0.3191} & 0.9363 & 0.9174 \\
     & MOFT \citep{moft2024xiao} & 0.5543 & 0.2181 & 0.2669 & 0.0597 & 0.1549 & 0.3151 & 0.9421 & 0.9012 \\
     & MotionClone \citep{motionclone2024ling} & 0.5965 & \underline{0.2330} & 0.2648 & \underline{0.0633} & \underline{0.1697} & 0.3154 & 0.9405 & 0.9370 \\
     & ConMo \citep{conmo2025gao} & 0.6782 & 0.2148 & 0.2655 & 0.0584 & 0.1654 & 0.3182 & 0.9385 & 0.8970 \\
     & DiTFlow \citep{ditflow2025pondaven} & \textbf{0.7159} & 0.2012 & 0.2671 & 0.0542 & 0.1500 & 0.3167 & \underline{0.9430} & \underline{0.9393} \\
     & MotionGrounder (Ours) & \underline{0.7140} & \textbf{0.3103} & \textbf{0.2728} & \textbf{0.0857} & \textbf{0.2353} & \textbf{0.3241} & \textbf{0.9440} & \textbf{0.9436} \\
    \midrule
    \multirow{6}{*}{\rotatebox[origin=c]{90}{\shortstack{\textbf{$N{\geq}3$}\\(15)}}} & DMT \citep{dmt2023yatim} & \textbf{0.7335} & 0.1274 & \textbf{0.2595} & 0.0344 & \underline{0.1011} & 0.3158 & 0.9366 & 0.9268 \\
     & MOFT \citep{moft2024xiao} & 0.5708 & \underline{0.1354} & 0.2535 & \underline{0.0360} & 0.0926 & 0.3121 & 0.9357 & 0.8797 \\
     & MotionClone \citep{motionclone2024ling} & 0.5847 & 0.1262 & 0.2593 & 0.0341 & 0.0904 & \textbf{0.3184} & \underline{0.9391} & \underline{0.9401} \\
     & ConMo \citep{conmo2025gao} & 0.6552 & 0.1246 & 0.2562 & 0.0341 & 0.0986 & \underline{0.3159} & 0.9377 & 0.9166 \\
     & DiTFlow \citep{ditflow2025pondaven} & 0.6745 & 0.1189 & \underline{0.2594} & 0.0322 & 0.0893 & 0.3099 & \textbf{0.9399} & 0.9293 \\
     & MotionGrounder (Ours) & \underline{0.6753} & \textbf{0.1428} & 0.2584 & \textbf{0.0392} & \textbf{0.1095} & 0.3144 & \textbf{0.9399} & \textbf{0.9404} \\
    \midrule
    \multirow{6}{*}{\rotatebox[origin=c]{90}{\shortstack{\textit{\textbf{All videos}}\\(52)}}} & DMT \citep{dmt2023yatim} & \textbf{0.6886} & 0.2057 & 0.2656 & 0.0571 & \underline{0.1574} & 0.3155 & 0.9348 & 0.9057 \\
     & MOFT \citep{moft2024xiao} & 0.5450 & 0.1988 & 0.2630 & 0.0553 & 0.1385 & 0.3138 & 0.9387 & 0.8907 \\
     & MotionClone \citep{motionclone2024ling} & 0.5526 & \underline{0.2081} & 0.2646 & \underline{0.0581} & 0.1463 & 0.3127 & 0.9392 & \underline{0.9345} \\
     & ConMo \citep{conmo2025gao} & 0.6503 & 0.1967 & 0.2642 & 0.0553 & 0.1483 & \underline{0.3168} & 0.9366 & 0.8982 \\
     & DiTFlow \citep{ditflow2025pondaven} & 0.6637 & 0.1868 & \underline{0.2657} & 0.0520 & 0.1361 & 0.3156 & \underline{0.9399} & 0.9259 \\
     & MotionGrounder (Ours) & \underline{0.6813} & \textbf{0.2371} & \textbf{0.2677} & \textbf{0.0671} & \textbf{0.1832} & \textbf{0.3182} & \textbf{0.9417} & \textbf{0.9379} \\
    \bottomrule
  \end{tabular}
  \vspace{-10pt}
  \end{table}

\subsection{Quantitative evaluation}

Table~\ref{tab:main_quantitative_comparison} reports results averaged over the \textit{\textbf{Caption}}, \textit{\textbf{Subject}}, and \textit{\textbf{Scene}} settings with ground-truth masks, over all videos and by object count $N$ (per-prompt results in Appx.~\mbox{%
\ref{app:per_count}}%
, automatic masks in Appx.~\mbox{%
\ref{app:noisy_masks}}%
).
Over \textit{\textbf{All videos}} we lead every metric except MF, where DMT is narrowly ahead.
The higher MF of DMT stems from the entanglement of its guidance signal with the source video $V_S$ \citep{ditflow2025pondaven}, which relies on spatially averaged global features from $V_S$.
This encourages excessive source adherence, resulting in stronger source motion preservation and higher MF scores.
While beneficial in some cases, \textit{this behavior is undesirable when source motion should not be rigidly preserved}.
For example, in Fig.~\ref{fig:main_qualitative_comparison} under the \textit{\textbf{Subject}} prompt setting, DMT preserves the original helicopter rotor motion on top of the tank, \textit{which is unrealistic}, with an MF of 0.9419, whereas our MotionGrounder generates more realistic motion with an MF of 0.9183, which we attribute to the source appearance-independent nature of FMS.
At $N{=}1$ we lead MF and match the best baseline on grounding.
From $N{=}2$ onward we lead IoU, OGS, and MGS (OGS 0.0857 and 0.0392 vs.\ 0.0633 and 0.0360, MGS 0.2353 and 0.1095 vs.\ 0.1697 and 0.1011 at $N{=}2$ and $N{\geq}3$), while DiTFlow ($N{=}2$) and DMT ($N{\geq}3$) lead MF.
Our grounding advantage is thus a multi-object one, appearing where objects can be confused.

\suppressfloats[t]

\begin{table}[t]
  \caption{
        \textbf{User study.} Mean rank (1 = best, 6 = worst) in motion adherence (MA), global textual faithfulness (GTF) and object grounding (OG), by object count $N$ and overall. OGS ranks are reported for reference.
    }
  \label{tab:user_study}
  \label{tab:user_study_per_count}
  \centering
  \fontsize{7pt}{8pt}\selectfont
  \setlength{\tabcolsep}{4.92pt}
  \begin{tabular}{l*{9}{c}*{4}{c}}
    \toprule
    \multirow{2}{*}{\textbf{Method}} & \multicolumn{3}{c}{$N{=}1$} & \multicolumn{3}{c}{$N{=}2$} & \multicolumn{3}{c}{$N{\geq}3$} & \multicolumn{3}{c}{All} & OGS \\
    \cmidrule(lr){2-4} \cmidrule(lr){5-7} \cmidrule(lr){8-10} \cmidrule(lr){11-13}
    & MA & GTF & OG & MA & GTF & OG & MA & GTF & OG & MA & GTF & OG & Rank \\
    \midrule
    DMT \citep{dmt2023yatim} & 3.61 & 3.70 & 3.61 & 3.58 & 3.62 & 3.70 & 3.88 & 3.92 & 3.95 & 3.65 & 3.70 & 3.72 & 3.68 \\
    MOFT \citep{moft2024xiao} & 3.68 & 3.54 & 3.59 & 3.86 & 3.87 & 3.89 & 3.84 & 3.91 & 3.94 & 3.81 & 3.78 & 3.81 & 3.74 \\
    MotionClone \citep{motionclone2024ling} & 3.76 & 3.91 & 4.06 & \underline{3.46} & \underline{3.30} & \underline{3.28} & 3.66 & 3.69 & 3.67 & 3.59 & 3.56 & 3.59 & 3.61 \\
    ConMo \citep{conmo2025gao} & 3.46 & 3.41 & 3.49 & 3.64 & 3.76 & 3.73 & 3.50 & 3.43 & \underline{3.37} & 3.56 & 3.59 & 3.58 & \underline{3.52} \\
    DiTFlow \citep{ditflow2025pondaven} & \textbf{3.22} & \textbf{3.17} & \underline{3.15} & 3.52 & 3.55 & 3.65 & \underline{3.38} & \underline{3.40} & 3.45 & \underline{3.40} & \underline{3.40} & \underline{3.46} & 3.75 \\
    MotionGrounder (Ours) & \underline{3.27} & \underline{3.27} & \textbf{3.11} & \textbf{2.93} & \textbf{2.91} & \textbf{2.75} & \textbf{2.74} & \textbf{2.66} & \textbf{2.63} & \textbf{2.99} & \textbf{2.97} & \textbf{2.83} & \textbf{2.70} \\
    \bottomrule
  \end{tabular}
  \vspace{-11.3pt}
  \end{table}

\textbf{User study.}
We conduct a user study comparing MotionGrounder against baselines and report results in Table~\ref{tab:user_study}.
We evaluate 50 videos with varying numbers of objects, where 49 participants rank each method in motion adherence (MA), global textual faithfulness (GTF), and object grounding (OG), the last of which evaluates whether the correct target objects are generated and placed in the appropriate spatial regions.
MotionGrounder achieves the best overall average rank, while Object Grounding Score (OGS) ranks align with human OG rankings.
By object count, our mean rank improves steadily as objects are added, as in Table~\mbox{%
\ref{tab:per_count}}%
: only on single-object clips is DiTFlow ranked slightly better on MA and GTF, while we stay ahead on OG.
Additional details are in Appx.~\mbox{%
\ref{app:user_study}}%
.

\Needspace{12\baselineskip}
\subsection{Ablation studies}

\begin{wrapfigure}{r}{0.65\linewidth}
    \vspace{-15.4pt}
    \centering
    \includegraphics[width=1.0\linewidth]{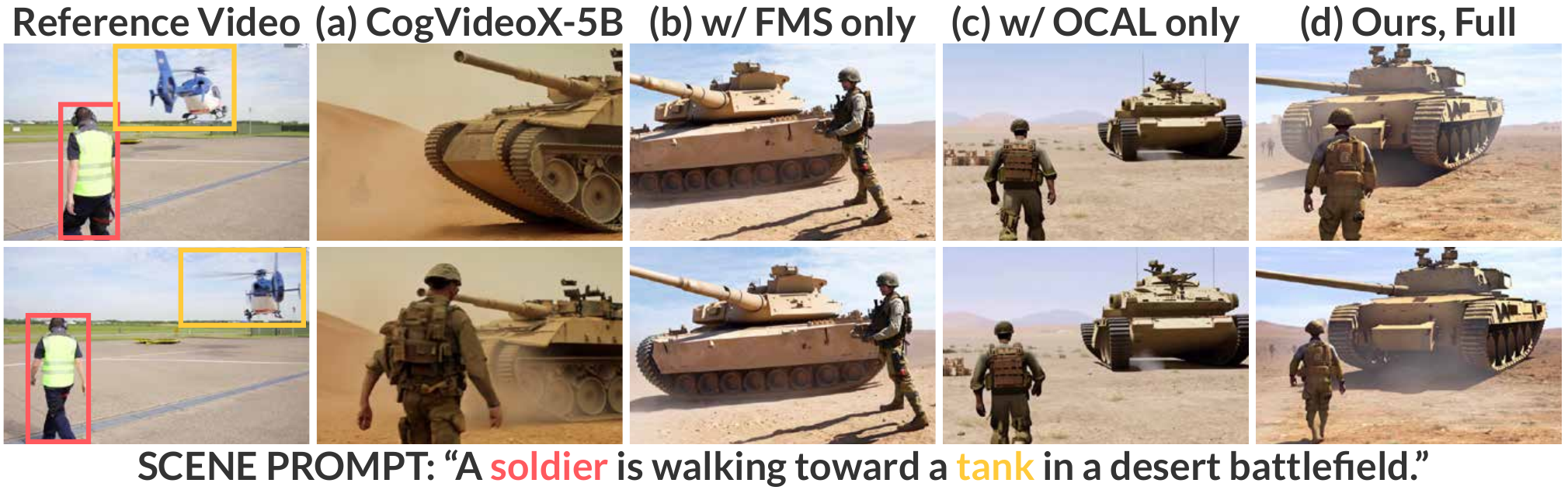}
    \vspace{-19pt}
    \caption{
        \textbf{Ablation of essential components.}
    }
    \vspace{-7.6pt}
    \label{fig:ablation_main}
\end{wrapfigure}

\textbf{Ablation of essential components.}
We study the individual and joint effects of FMS and OCAL in Table~\ref{tab:ablation_main} and Fig.~\ref{fig:ablation_main}, with per-prompt results in Table~\mbox{%
\ref{tab:ablation_main_full} }\hskip0pt%
(Appx.~\mbox{%
\ref{app:full_ablations}}%
) and results by object count in Table~\mbox{%
\ref{tab:per_count_full} }\hskip0pt%
(Appx.~\mbox{%
\ref{app:main_ablation}}%
).
Without FMS or OCAL, the backbone in (a) struggles with both motion transfer and object localization.
The ablations reveal  that:
(i) enabling only FMS in (b) substantially improves MF but \textit{yields relatively low IoU due to inaccurate object placement};
(ii) enabling only OCAL in (c) improves IoU, LTA, and OGS by enforcing object--caption association and spatial localization, but \textit{MF degrades due to the lack of explicit motion transfer}, as seen in Fig.~\ref{fig:ablation_main}-(c), where the soldier and tank are correctly placed but exhibit limited motion;
(iii) \textit{enabling both FMS and OCAL in (d) achieves faithful motion transfer and accurate object grounding}, yielding the highest MF and LTA and the second-best IoU, OGS, and MGS.
Adding OCAL on top of FMS leaves MF essentially unchanged, so grounding is gained without sacrificing motion.
Compared to (c), IoU and OGS in (d) slightly decrease as explicit motion constraints allow object positions to evolve temporally rather than enforcing rigid frame-wise localization, while LTA is maintained.
MGS also favors (c), for a different reason: (d) moves objects far more like the source (object-level MF 0.418 vs.\ 0.299, Appx.~\mbox{%
\ref{app:gobj}}%
), but MGS maps object-level MF to $[0, 1]$, so a nearly static object inside its region keeps most of its score and placement dominates. OCAL-only is thus not a usable operating point: it grounds objects that barely move.
Overall, our joint formulation avoids rigid placement and enables natural motion while preserving semantic grounding.

\begin{wrapfigure}{r}{0.62\linewidth}
    \vspace{-7pt}
    \centering
    \includegraphics[width=1.0\linewidth]{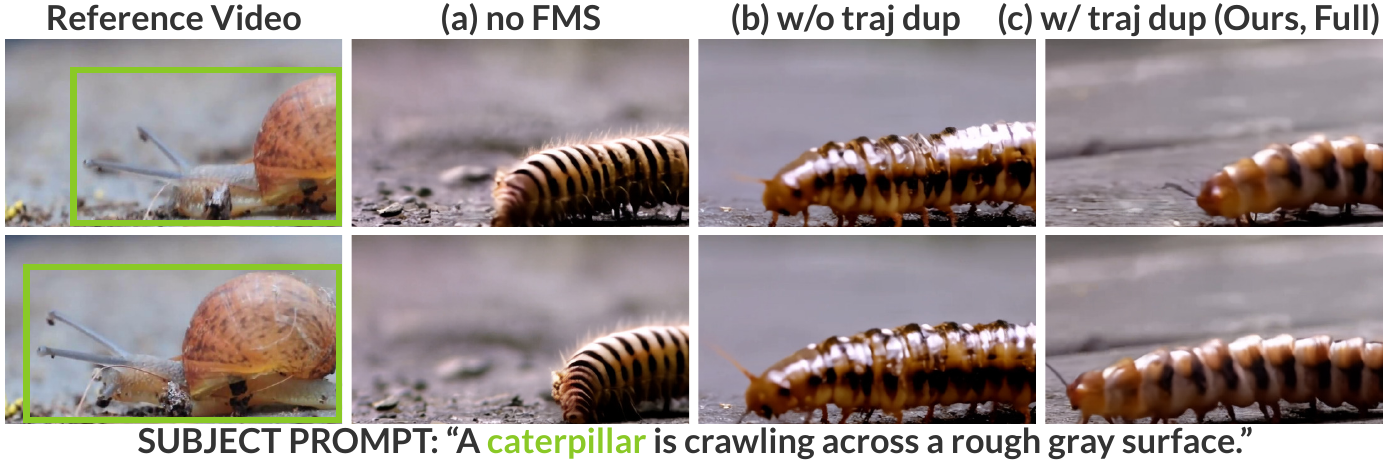}
    \vspace{-18pt}
    \caption{
        \textbf{FMS ablation.}
    }
    \label{fig:ablation_fms}
    \vspace{-4.1pt}
\end{wrapfigure}

\textbf{FMS ablation.}
We analyze FMS design choices in Fig.~\ref{fig:ablation_fms}, starting from an OCAL-only baseline in (a).
While OCAL correctly places the object in the target region, it fails to preserve the reference motion, as seen in the incorrect rightward motion of the caterpillar.
Introducing FMS substantially improves MF in both (b) and (c), confirming its effectiveness.
When trajectory duplication is disallowed, object motion is noticeably suppressed (Fig.~\ref{fig:ablation_fms}-(b)).
This is because enforcing one-to-one correspondences creates competition among patches, causing some patches to remain stationary and limiting overall object movement.
Allowing trajectory duplication in (c) lets patches share motion, so moving patches can propagate their motion together.
This enables larger object motion (Fig.~\ref{fig:ablation_fms}-(c)) and better spatial alignment.
The quantitative results are reported in Table~\mbox{%
\ref{tab:ablation_fms_full} }\hskip0pt%
(Appx.~\mbox{%
\ref{app:fms_ablation}}%
).

\begin{figure}[!h]
    \centering
    \includegraphics[width=1.0\linewidth]{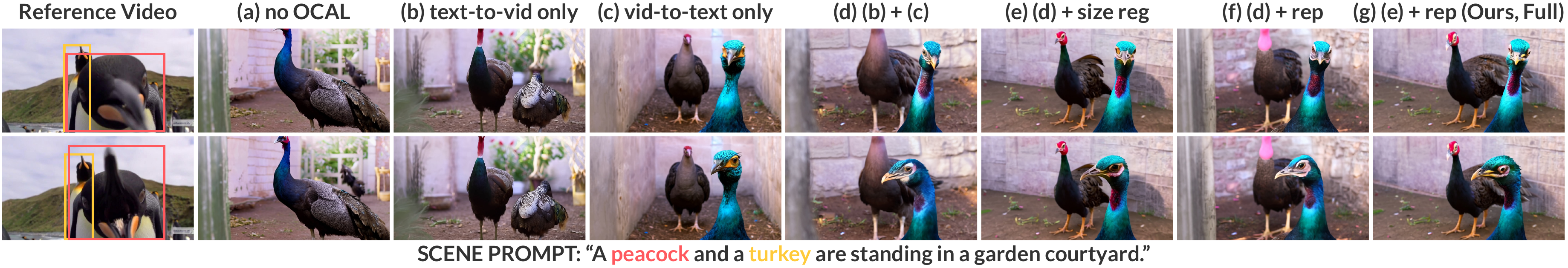}
    \vspace{-18pt}
    \caption{
        \textbf{OCAL ablation.}
    }
    \label{fig:ablation_ocal}
    \vspace{-9.9pt}
\end{figure}

\textbf{OCAL ablation.}
We further ablate the components of OCAL, with the results reported in Table~\ref{tab:ablation_ocal} and Fig.~\ref{fig:ablation_ocal}.
For (a)--(e), we disable repetition and fuse object masks every four frames.
The analysis results are as follows:
(a) removing OCAL leads to weak semantic alignment and object grounding, yielding the lowest IoU, LTA, OGS, and MGS;
using only (b) text-to-video or (c) video-to-text attention improves semantic and object alignment, reflected by higher LTA and IoU;
(d) combining (b) and (c) improves object grounding, increasing IoU;
(e) adding the size regularizer grounds best at a small cost in motion; (f) adding repetition alone has no measurable effect; and (g) adding repetition on top of (e) wins the motion back (MF 0.6775 to 0.6813).
The size regularizer prevents large objects from dominating optimization, as shown in Fig.~\ref{fig:ablation_ocal}-(e) and (g), where the turkey is fully generated despite occupying a small region, unlike in Fig.~\ref{fig:ablation_ocal}-(a) to (d) and (f).

\begin{table}[!htb]
  \fontsize{7pt}{8pt}\selectfont
  \setlength{\tabcolsep}{3pt}
  \begin{minipage}[t]{0.47\linewidth}
    \centering
    \caption{
        \textbf{Ablation of essential components.}
 }
    \label{tab:ablation_main}
    \begin{tabular}{l*{5}{c}}
      \toprule
      \textbf{Method} & MF $\uparrow$ & IoU $\uparrow$ & LTA $\uparrow$ & OGS $\uparrow$ & MGS $\uparrow$ \\
      \midrule
      (a) Backbone & 0.4257 & 0.1593 & 0.2662 & 0.0446 & 0.1008 \\
      (b) FMS only & \underline{0.6714} & 0.1652 & 0.2634 & 0.0462 & 0.1230 \\
      (c) OCAL only & 0.4860 & \textbf{0.3099} & \textbf{0.2684} & \textbf{0.0868} & \textbf{0.2072} \\
      (d) Ours, Full & \textbf{0.6813} & \underline{0.2371} & \underline{0.2677} & \underline{0.0671} & \underline{0.1832} \\
      \bottomrule
    \end{tabular}
  \end{minipage}
  \hfill
  \begin{minipage}[t]{0.52\linewidth}
    \centering
    \caption{
        \textbf{OCAL ablation.}
 }
    \label{tab:ablation_ocal}
    \begin{tabular}{l*{5}{c}}
      \toprule
      \textbf{Method} & MF $\uparrow$ & IoU $\uparrow$ & LTA $\uparrow$ & OGS $\uparrow$ & MGS $\uparrow$ \\
      \midrule
      (a) no OCAL & 0.6714 & 0.1652 & 0.2634 & 0.0462 & 0.1230 \\
      (b) text-to-vid only & 0.6749 & 0.2328 & 0.2668 & 0.0659 & 0.1736 \\
      (c) vid-to-text only & 0.6675 & 0.2157 & 0.2662 & 0.0609 & 0.1605 \\
      (d) (b) + (c) & 0.6791 & \underline{0.2375} & 0.2654 & 0.0667 & 0.1778 \\
      (e) (d) + size reg & 0.6775 & \textbf{0.2475} & \underline{0.2672} & \textbf{0.0698} & \textbf{0.1866} \\
      (f) (d) + rep & \underline{0.6800} & 0.2345 & 0.2644 & 0.0656 & 0.1782 \\
      (g) (e) + rep (Ours, Full) & \textbf{0.6813} & 0.2371 & \textbf{0.2677} & \underline{0.0671} & \underline{0.1832} \\
      \bottomrule
    \end{tabular}
  \end{minipage}
  \vspace{-12pt}
  \end{table}

\section{Conclusion}
\label{sec:conclusion}

In this work, we introduce MotionGrounder, a DiT-based framework that is the \textit{first} to jointly handle motion transfer and object grounding in \textit{multi-object} scenes.
Our FMS provides a stable motion prior, while our OCAL enables spatially grounded object captions.
We also propose OGS for joint evaluation of spatial grounding and semantic consistency, and MGS, which scores motion per object and weights it by object localization.
Our experiments demonstrate improved object grounding over recent state-of-the-art methods at competitive motion fidelity, an advantage that appears once several objects are present and is confirmed by human evaluation.

\clearpage
\begin{ack}
This work was supported by Institute of Information and communications Technology Planning and Evaluation (IITP) grant funded by the Korean government [Ministry of Science and ICT (Information and Communications Technology)] (Project Number: RS2022-00144444, Project Title: Deep Learning Based Visual Representational Learning and Rendering of Static and Dynamic Scenes). The authors declare no competing interests.
\end{ack}

\bibliographystyle{plainnat}
\bibliography{references}
\clearpage

\appendix
\raggedbottom
\setlength{\textfloatsep}{10.5pt plus 0.5pt minus 1pt}
\setlength{\floatsep}{10.5pt plus 0.5pt minus 1pt}
\setlength{\intextsep}{10.5pt plus 0.5pt minus 1pt}
\renewcommand{\topfraction}{0.9}
\renewcommand{\bottomfraction}{0.8}
\renewcommand{\textfraction}{0.07}
\renewcommand{\floatpagefraction}{0.85}
\setcounter{topnumber}{4}
\setcounter{bottomnumber}{3}
\setcounter{totalnumber}{6}
\makeatletter
\setlength{\@fptop}{0pt}
\setlength{\@fpsep}{10.5pt plus 1fil}
\setlength{\@fpbot}{0pt plus 1fil}
\makeatother
\newpage
\addtocontents{toc}{\protect\setcounter{tocdepth}{2}}
{\renewcommand{\contentsname}{Appendix Contents}\setlength{\parskip}{0pt}\makeatletter\renewcommand{\@pnumwidth}{2em}\renewcommand*\l@section[2]{\ifnum\c@tocdepth>\z@\addpenalty\@secpenalty\addvspace{0.45em \@plus\p@}\setlength\@tempdima{1.5em}\begingroup\parindent\z@\rightskip\@pnumwidth\parfillskip-\@pnumwidth\leavevmode\bfseries\advance\leftskip\@tempdima\hskip-\leftskip#1\nobreak\hfil\nobreak\hb@xt@\@pnumwidth{\hss#2}\par\penalty\@highpenalty\endgroup\fi}\makeatother\tableofcontents}
\newpage
\section*{Appendix}

\section{Broader Impact}
\label{app:broader_impact}
MotionGrounder advances controllable video generation by enabling multi-object motion transfer with explicit grounding between textual descriptions and object regions.
This capability can benefit creative industries, education, simulation, and assistive content creation, where precise control over video content is valuable.
However, like other video generation frameworks, MotionGrounder could be misused to create misleading, deceptive, or non-consensual media.
As it builds upon large-scale generative models, it also inherits broader societal risks related to bias, privacy, and the use of unsafe or uncurated training data.
Addressing these concerns calls for complementary efforts in detection, attribution, dataset curation, bias mitigation, and content authentication, alongside appropriate technical and institutional safeguards.

\section{Dataset}
\label{app:dataset}

\subsection{Video Selection}
\label{app:dataset_selection}
Due to the multi-object nature of our task, we construct a dedicated evaluation dataset.
We select 19 videos from DAVIS \citep{davis2016Perazzi}, 29 from YouTube-VOS \citep{youtubevos2018xu}, and 4 from ConMo \citep{conmo2025gao}, for a total of 52 videos, all of which are provided with object masks.
For each video, we uniformly sample 24 frames and center-crop them to a resolution of $480 \times 720$ pixels.

We rank candidate videos from DAVIS, YouTube-VOS, and ConMo by their summed per-frame optical flow magnitude over the clip.
We keep a candidate only if the source dataset provides a per-frame instance mask for every object our caption names across all 24 sampled frames, with instances that are separate rather than merged into one foreground region; videos fail this when an object is annotated for only part of the clip or when two objects share a single mask.
We then take the top-ranked surviving videos within each object-count bucket, since otherwise the list would collapse onto whichever count happens to move most.
This makes object count a controlled variable of the benchmark.

\subsection{Object-Count Statistics}
\label{app:dataset_stats}
\begin{figure}[!htb]
    \centering
    \includegraphics[width=0.60\linewidth]{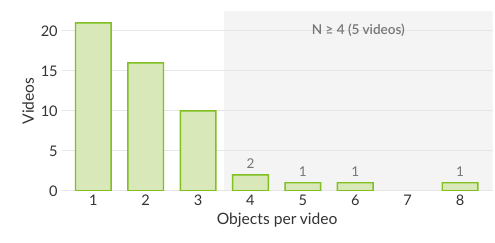}
\vspace{-14.6pt}
    \caption{
        \textbf{Object-count distribution of our 52-video benchmark.}
    }
    \label{fig:object_count_hist}
\end{figure}

Of the 52 videos, 21 contain one object, 16 two, 10 three, and 5 four or more (two videos with 4 objects and one each with 5, 6, and 8), i.e., 21 single-object and 31 multi-object videos.
Fig.~\mbox{%
\ref{fig:object_count_hist} }\hskip0pt%
shows the full distribution.

\textbf{Video list} (object count in parentheses).
\textit{DAVIS (19):} lindy-hop (8), dog-gooses (5), classic-car (3), dancing (3), dogs-jump (3), pigs (3), soapbox (3), judo (2), kid-football (2), car-turn (1), dog (1), drift-straight (1), elephant (1), koala (1), mallard-fly (1), parkour (1), planes-water (1), rallye (1), rollerblade (1).
\textit{YouTube-VOS (29):} b1a8a404ad (6), c6d9526e0d (4), ff5a1ec4f3 (4), 0f17fa6fcb (3), 193aa74f36 (3), a4065a7eda (3), a8999af004 (3), ab2e1b5577 (3), 1171141012 (2), 285b69e223 (2), 28e392de91 (2), 2efb07554a (2), 511e8ecd7f (2), 7b733d31d8 (2), 7ff84d66dd (2), bfab1ad8f9 (2), cad8a85930 (2), d1ac0d8b81 (2), 0d284ff4c2 (1), 0f6c2163de (1), 3401d58137 (1), 8aca214360 (1), 972c187c0d (1), b8fb75864e (1), db0968cdd3 (1), df365282c6 (1), f49e4866ac (1), f6d6f15ae7 (1), fad633fbe1 (1).
\textit{ConMo (4):} cart (2), couple (2), couple-runner (2), helicopter (2).
Masks, captions, and the construction pipeline will be released with the dataset.

\subsection{Caption Generation Pipeline}
\label{app:dataset_pipeline}
\begin{figure}[!htb]
    \centering
    \includegraphics[width=1.0\linewidth]{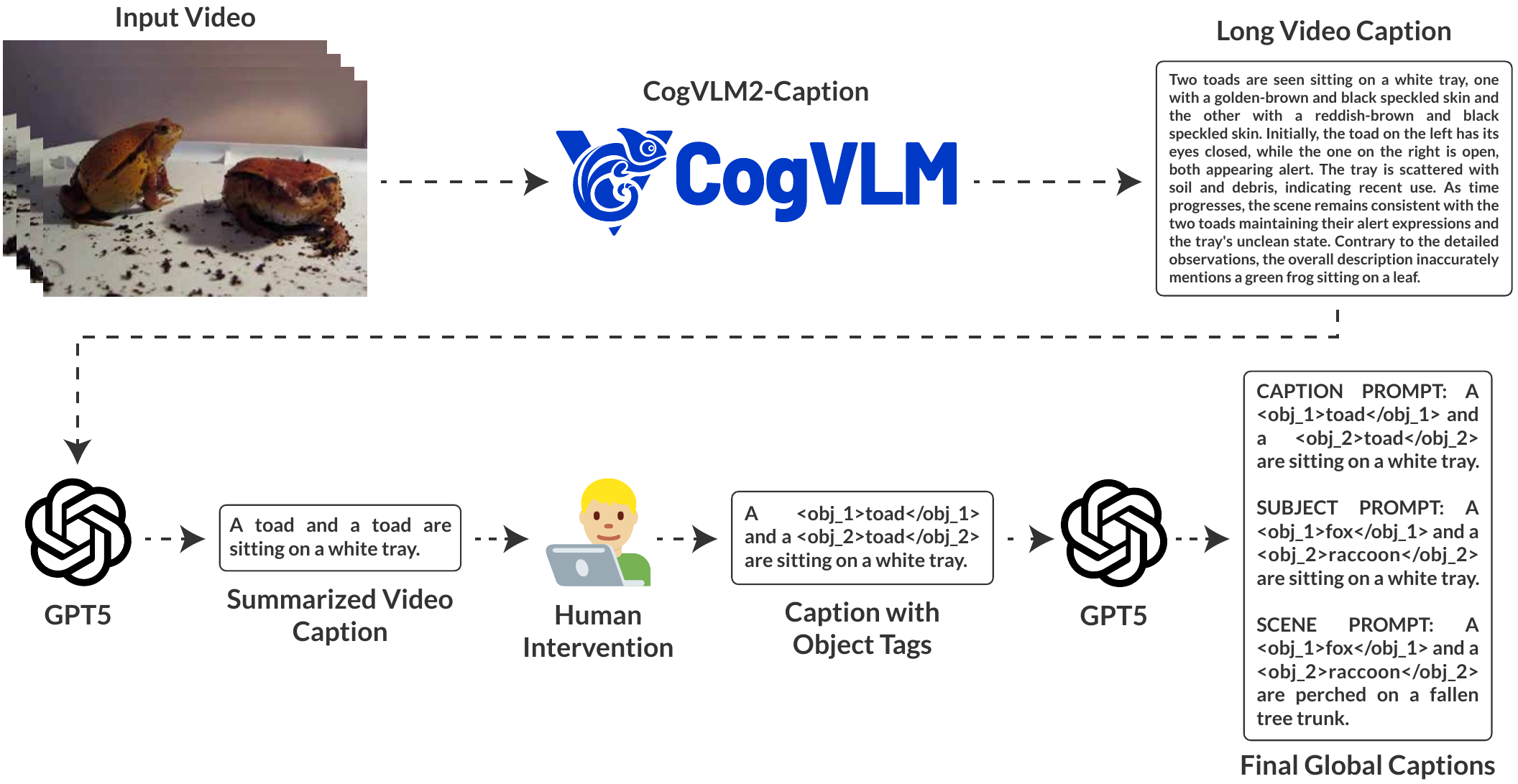}
\vspace{-19.1pt}
    \caption{
        \textbf{Dataset generation pipeline.}
        Long captions are first generated using CogVLM2-Caption \citep{cogvlm22024hong} and summarized into structured prompts with GPT-5 \citep{gpt52023openai}, followed by object--mask association, and target prompt construction.
    }
    \label{fig:supp_dataset_pipeline}
\end{figure}

Fig.~\mbox{%
\ref{fig:supp_dataset_pipeline} }\hskip0pt%
illustrates our caption generation pipeline.
As the original videos do not include captions, we first generate detailed descriptions using CogVLM2-Caption \citep{cogvlm22024hong}, which are then summarized with GPT-5 \citep{gpt52023openai} into the structured format \texttt{<subject>} \texttt{<verb>} \texttt{<scene>}.
Each object appearing in the \texttt{<subject>} is manually tagged to associate it with the corresponding object mask.
Target captions are subsequently created by modifying the subject, verb, and scene using GPT-5.
Following \citep{ditflow2025pondaven}, we construct three types of prompts:
(i) a \textit{\textbf{Caption}} prompt that directly describes the input video content;
(ii) a \textit{\textbf{Subject}} prompt that alters the main objects while preserving the original background; and
(iii) a \textit{\textbf{Scene}} prompt that specifies an entirely new scene distinct from the source video.

\clearpage{}
\subsection{Sample Frames and Captions}
\label{app:dataset_samples}
\begin{figure}[!htb]
    \centering
    \includegraphics[width=1.0\linewidth]{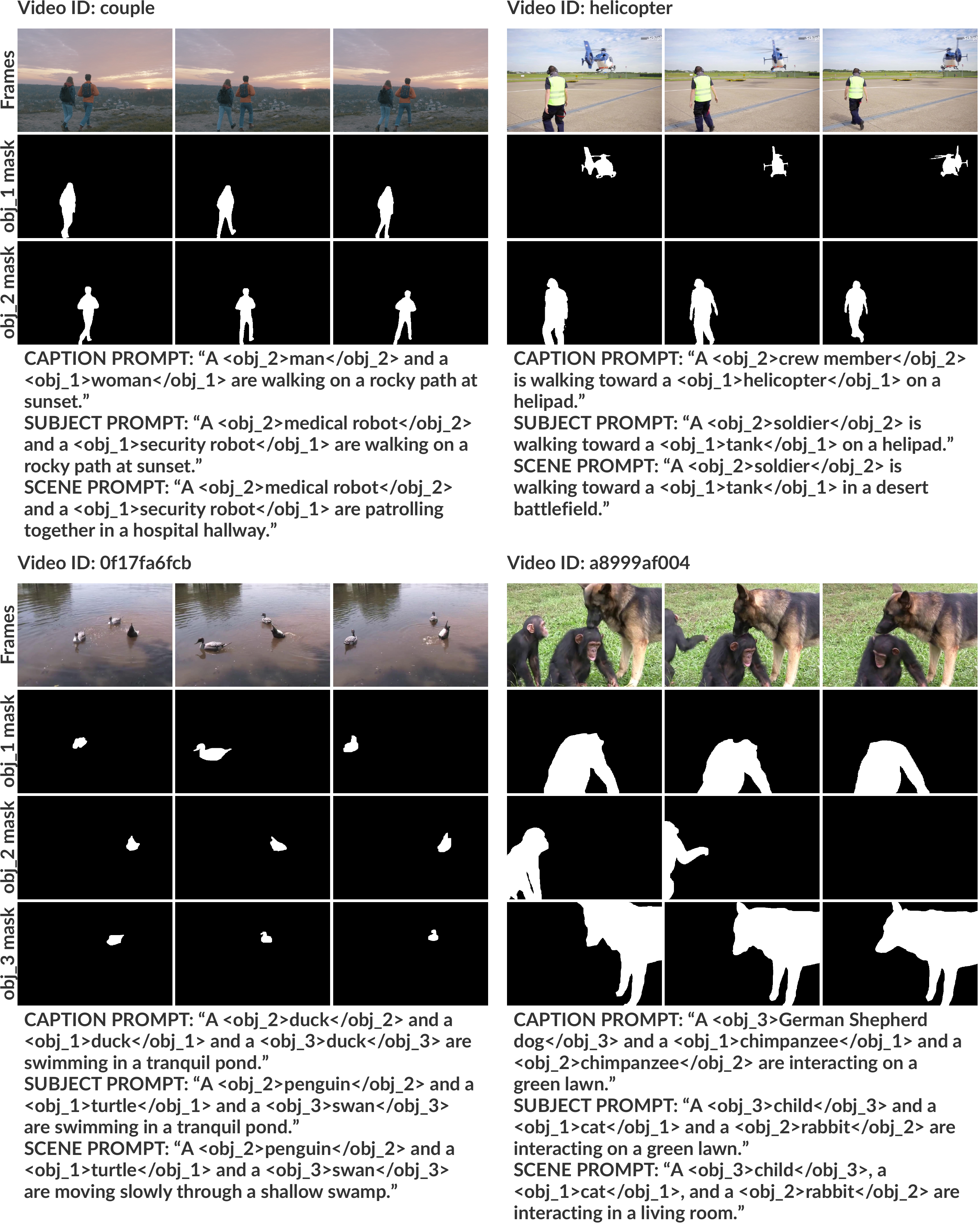}
\vspace{-10pt}
    \caption{
        Sample frames and masks from videos in our dataset, shown with their corresponding \textit{Caption}, \textit{Subject}, and \textit{Scene} prompts.
    }
    \label{fig:supp_sample_frames}
\end{figure}

Fig.~\ref{fig:supp_sample_frames} shows sample frames and masks from videos in our dataset, together with their corresponding \textit{\textbf{Caption}}, \textit{\textbf{Subject}}, and \textit{\textbf{Scene}} prompts.

\clearpage{}
\subsection{Effect of Source Motion}
\label{app:near_static}
We select high-motion sources on the assumption that near-static sources make transfer trivial, since reconstructing the source already scores well.
To verify this, we measure two quantities for each of the 52 source videos: object motion, as the median GMFlow displacement inside the object masks, and camera motion, as the median displacement of background feature tracks (Shi-Tomasi corners followed with Lucas-Kanade flow, keeping only tracks whose forward and backward estimates agree within 1 pixel).
We then compute the Spearman correlation of each quantity with every method's per-video scores, averaged over the three prompts.
The MF of every baseline falls as source object motion rises (MotionClone $\rho = -0.53$, DMT $-0.38$, ConMo $-0.33$, DiTFlow $-0.32$, MOFT $-0.29$, all $p < 0.05$), so the high-motion selection makes the benchmark harder rather than easier.
Our MF shows the weakest dependence on object motion ($\rho = -0.16$, $p = 0.27$, not significant) and none on camera motion ($\rho = +0.05$, $p = 0.74$), so our results are not explained by camera-driven motion.
Our OGS, however, falls as source object motion rises ($\rho = -0.43$, $p = 0.002$), so our grounding advantage is largest on sources with less object motion, whereas the baselines show no such correlation because their grounding is uniformly lower.
We discuss this limitation in Appx.~\mbox{%
\ref{app:limitation}}%
.
\section{Object Grounding Score (OGS)}
\label{app:ogs}
Prior works on motion transfer \citep{dmt2023yatim, ditflow2025pondaven, conmo2025gao, det2025shi} predominantly rely on a global CLIP score to evaluate textual alignment, which effectively captures overall text--video consistency but provides no notion of object localization or object--caption association.
As a result, fine-grained object-level failures, such as multiple target objects appearing in swapped spatial regions, may remain undetected when the relevant concepts are still present in the frame.

Intersection-over-Union (IoU) addresses spatial localization accuracy, but is agnostic to semantic alignment and therefore cannot determine whether a localized region corresponds to the correct object description.
Conversely, local CLIP similarity, which we adopt from InstanceDiffusion \mbox{%
\citep{instancediffusion2024wang}}%
, evaluates object--caption association and semantic consistency, but lacks explicit spatial grounding, making it insufficient to assess correct object placement.

Our proposed Object Grounding Score (OGS) deliberately focuses on object-level evaluation by combining IoU and local CLIP similarity, jointly capturing object localization and object--caption association without modeling global alignment.
This design enables reliable detection of fine-grained grounding errors that are overlooked by global caption-level metrics, while complementing existing global alignment evaluations rather than replacing them.
The contribution of OGS is thus not either component but their joint, per-object requirement that the right object also appears in the right region.
A comparison of these evaluation capabilities across different metrics is summarized in Table~\ref{tab:ogs_motivation}.

\begin{table}[!htb]
  \caption{
    \textbf{Motivation for Object Grounding Score (OGS).}
    Comparison of commonly used evaluation metrics and the aspects of alignment they capture.
  }
  \label{tab:ogs_motivation}
  \centering
  \fontsize{8pt}{9pt}\selectfont
  \setlength{\tabcolsep}{4pt}
    \begin{tabular}{l*{3}{c}}
      \toprule
      \textbf{Metric} & Global Alignment & Object Localization & Object--Caption Association \\
      \midrule

      Global CLIP Score
        & \cmark & \xmark & \xmark \\
      Intersection-over-Union
        & \xmark & \cmark & \xmark \\
      Local CLIP Score
        & \xmark & \xmark & \cmark \\
      \textbf{OGS (Ours)}
        & \xmark & \cmark & \cmark \\

      \bottomrule
    \end{tabular}
\end{table}

\subsection{Intersection over Union (IoU)}
\begin{figure}[!htb]
    \centering
    \includegraphics[width=1.0\linewidth]{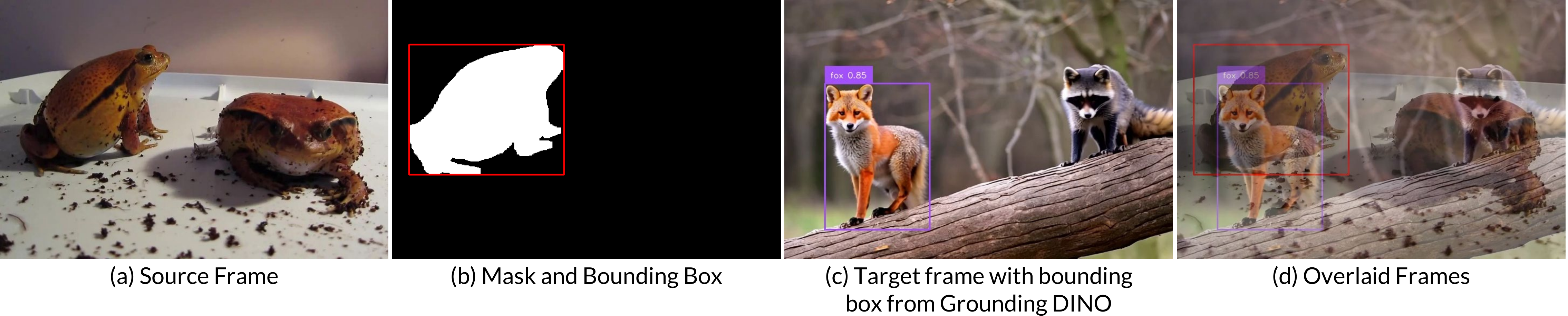}
\vspace{-18.0pt}
    \caption{
        \textbf{Comparison between bounding boxes inferred from ground-truth masks and Grounding DINO predictions.}
        We show the bounding box inferred from the ground-truth mask in (b), the Grounding DINO prediction in (c), and the overlaid frames and bounding boxes in (d), where it is evident that the ground-truth mask--based bounding box no longer accurately localizes the generated object.
    }
    \label{fig:supp_groundingdino}
\end{figure}

Given a source frame, we derive the bounding box directly from the object's ground-truth mask.
However, after generation, the newly generated object's location, shape, and scale may change relative to the original object in response to the target prompt.
As a result, the original ground-truth mask (and its bounding box) is no longer suitable for localizing the corresponding object in the output frame, as shown in Fig.~\ref{fig:supp_groundingdino}.
Therefore, we employ an open-vocabulary grounding model, Grounding DINO \citep{groundingdino2024liu}, to localize the target object in the generated frame and obtain its bounding box.
For each object $i$ in a frame, we compute the IoU between the source object region, defined by the bounding box inferred from the ground-truth mask, and the corresponding object region in the target video, defined by the predicted bounding box, and denote this score as $s_{\text{IoU}}$.
Higher values of $s_{\text{IoU}}$ indicate better spatial alignment between the source and generated objects.

\subsection{Local CLIP Similarity}
\begin{figure}[!htb]
    \centering
    \includegraphics[width=1.0\linewidth]{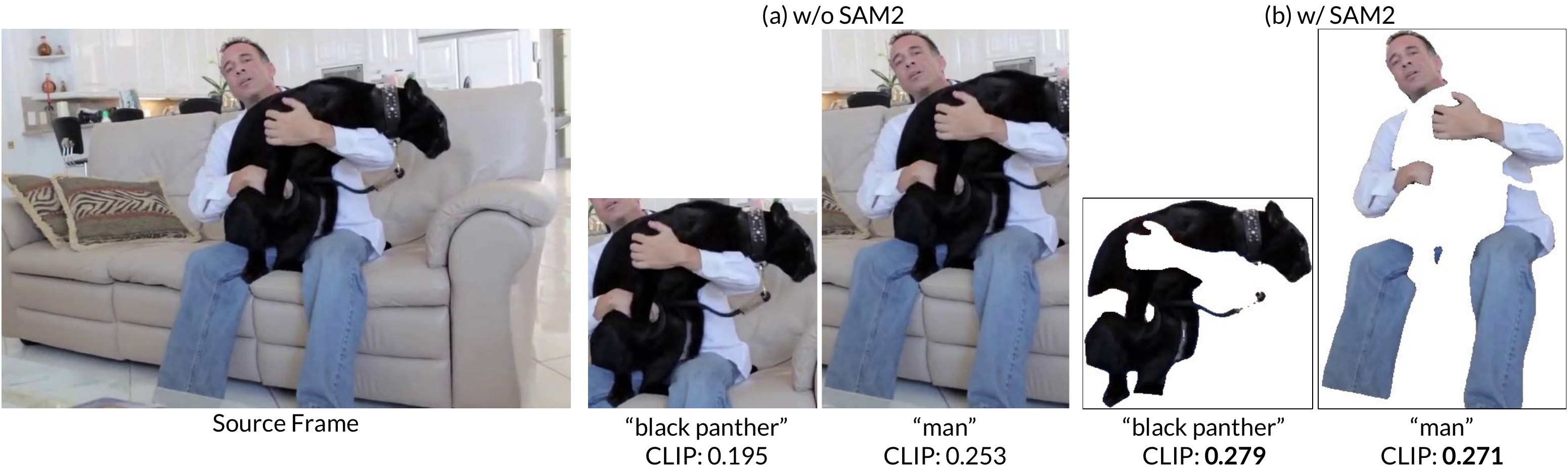}
\vspace{-18.7pt}
    \caption{
        \textbf{Effect of SAM2 \citep{sam22024ravi} segmentation masks on CLIP scores.}
        We report the local CLIP scores computed without applying the SAM2 segmentation mask in (a) and the local CLIP scores computed with the SAM2 segmentation mask in (b).
    }
    \label{fig:supp_sam2}
\end{figure}

CLIP similarity measures the alignment between textual and visual content.
To compute local CLIP similarity, reported as LTA in the tables \citep{instancediffusion2024wang}, we crop the object from the target video frame using the bounding box predicted by Grounding DINO \citep{groundingdino2024liu}.
To obtain a more reliable local score, we further refine the object region through segmentation.
Specifically, we use the Grounding DINO \citep{groundingdino2024liu} bounding box as a box prompt and apply SAM2 \citep{sam22024ravi} to segment the object and mask the remaining region of the cropped frame, yielding a more accurate crop.
CLIP similarity of a certain object $s_{\text{CLIP}}$ is then computed on the resulting masked crop.
As shown in Fig.~\ref{fig:supp_sam2}, without this refinement, CLIP similarity can be unreliable in certain cases (e.g., object overlap), since the crop may include background or other objects, diluting textual alignment.

\subsection{OGS Computation}
By combining $s_{\text{IoU}}$ and $s_{\text{CLIP}}$ into a single metric, which we term the Object Grounding Score (OGS), we measure how well a method generates the correct target objects in their correct spatial regions relative to the source video.
For each object $i$ in frame $f$, we define the object-level score as $\text{OGS}_{i,f} = s_{\text{IoU}}^{i,f} \cdot s_{\text{CLIP}}^{i,f}$.
However, in generation tasks, an object may be specified in the caption, but the model may fail to generate it in the target video.
To handle such cases, we explicitly define how these failures are treated in the computation of OGS.
When an object specified by the caption is not generated in a given frame, we assign its object-level score as zero, i.e., $\text{OGS}_{i,f} = 0$.
This value is included in the final OGS computation, thereby penalizing failures to generate the specified object.
Accordingly, we express $\text{OGS}_{i,f}$ as:

\begin{equation}
    \text{OGS}_{i,f} =
    \begin{cases}
    0, & \text{if object } i \text{ is specified by the caption but not generated},\\
    s_{\text{IoU}}^{i,f} \cdot s_{\text{CLIP}}^{i,f}, & \text{otherwise}.
    \end{cases}
\end{equation}

OGS is computed only over the objects named in the caption, so a generated object that the caption does not name is never scored. When a caption object is absent from a source frame, i.e.\ its mask is empty there, a detection in that frame counts as a false positive with $\text{OGS}_{i,f} = 0$, and no detection leaves the object unscored in that frame. Each frame is averaged over its scored objects, and a frame with none is omitted.
Overall, the video-level Object Grounding Score (OGS) is computed by averaging over all objects and frames:

\begin{equation}
    \text{OGS} = \frac{1}{F} \sum_{f=1}^{F} \left( \frac{1}{N_f} \sum_{i=1}^{N_f} \text{OGS}_{i,f} \right),
\end{equation}

where $F$ denotes the total number of frames and $N_f$ denotes the number of objects in frame $f$.

The proposed OGS provides a unified metric for evaluating object grounding performance over an entire video.
Higher OGS values indicate that target objects are generated at the correct locations with strong textual alignment to their corresponding captions, whereas lower values reflect failures in object generation, including errors in localization and semantic alignment.

\subsection{Object-level Motion Fidelity}
\label{app:gobj}
MF \mbox{%
\citep{dmt2023yatim} }\hskip0pt%
runs over tracklets on the whole frame, so nothing ties a tracklet to its object.
Following the design of OGS, we define an object-level MF: for each object $i$, we seed CoTracker \mbox{%
\citep{cotracker2024karaev} }\hskip0pt%
tracklets inside its source mask $M_i$ and inside the generated-object region, and apply the standard MF similarity between the two tracklet sets.
We denote this score as $s_{\text{MF}}^{i} \in [-1, 1]$.
As in OGS, the generated-object region is detected with Grounding DINO and SAM2 rather than reusing $M_i$, since the object is displaced and reshaped after generation.
On this ungrounded object-level MF, averaged per video as in OGS, DMT (0.482) and ConMo (0.452) rank above ours (0.418), followed by DiTFlow (0.402), MotionClone (0.357), and MOFT (0.353).
However, this score only asks whether the detected object moved like its source: tracklets are seeded wherever the object was found, so placement never enters the score, and a method that edits conservatively keeps its trajectories closest to the source.

\subsection{MGS Computation}
\label{app:mgs}
We therefore ground the object-level MF in the same way OGS grounds the local CLIP similarity: we map $s_{\text{MF}}^{i}$ to $[0, 1]$ and weight it by the object's IoU against the source region in each frame, $\text{MGS}_{i,f} = s_{\text{IoU}}^{i,f} \cdot (1 + s_{\text{MF}}^{i})/2$.
An object that is not generated in a frame has $s_{\text{IoU}}^{i,f} = 0$, as in OGS, and the video-level score averages over objects and frames in the same way,

{\begin{equation}
    \text{MGS} = \frac{1}{F} \sum_{f=1}^{F} \left( \frac{1}{N_f} \sum_{i=1}^{N_f} \text{MGS}_{i,f} \right).
\end{equation}
}

During the discussion phase, this score, then named GObj, was averaged over objects. We now average per frame and per video exactly as OGS, so the two scores are directly comparable. All tables were regenerated from the stored scoring outputs and agree with the numbers posted during the discussion. MGS differs from GObj by at most 0.004 on our benchmark and 0.009 on DAVIS, and all rankings are unchanged.
On MGS, ours ranks first (0.1832), ahead of DMT (0.1574), ConMo (0.1483), MotionClone (0.1463), MOFT (0.1385), and DiTFlow (0.1361). DMT reproduces object motion more closely, whereas ours places objects more accurately, and MGS rewards both.

\clearpage
\clearpage
\section{Additional Framework Details}
\label{app:framework}

\subsection{Motivation for Flow-based Motion Signal (FMS)}
\label{app:fms_motivation}
Attention Motion Flow (AMF) \citep{ditflow2025pondaven} reads displacement from an argmax over cross-frame attention, a signal never trained to match points, so its motion estimation is noisy. By noise we refer to the patch trajectories that AMF chains from these matches. The trajectory error compounds, since a wrong match shifts every later step of the chain.
To address this limitation, we introduce our Flow-based Motion Signal (FMS).
Fig.~\ref{fig:amf_vs_fms} compares patch correspondences from AMF and our FMS.
The visualization reflects the $4\times$ temporal compression of the VAE encoder \citep{cogvideox2025yang}, with frames fused every four input frames, so each panel overlays four input frames rather than showing one.

As illustrated in the figure, AMF suffers from several failure modes:
(i) it drifts on repeated background texture,
(ii) it confuses visually similar objects, which hinders accurate motion segregation and object-level motion preservation, and
(iii) it is sensitive to large or fast motions, leading to unstable trajectories.

\begin{figure}[!htb]
    \centering
    \includegraphics[width=1.0\linewidth]{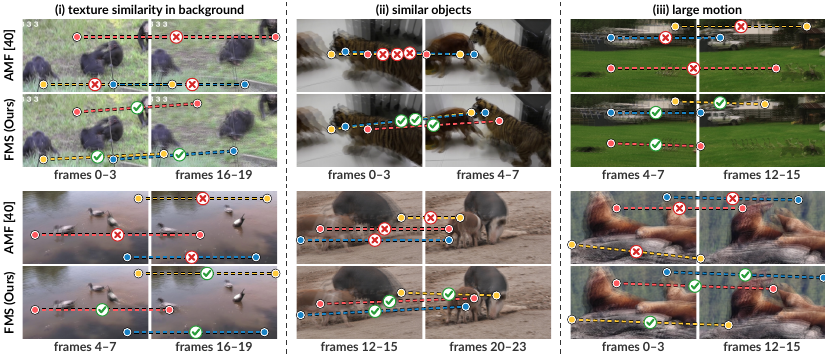}
\vspace{-18.3pt}
    \caption{
        \textbf{Extended comparison of AMF \citep{ditflow2025pondaven} and our FMS on six further videos.} Same construction as Fig.~\ref{fig:amf_vs_fms}: each panel overlays the four input frames of one latent frame, and dashed lines link three source patches to their estimated positions. A check marks an estimate within 1.25 patches of CoTracker3 \citep{cotracker2024karaev}, a cross one off by at least 3 patches (in (ii), on its own object vs.\ on another object under the ground-truth masks). AMF (i) moves nearly static background by up to 25 patches along repeated texture, (ii) jumps between similar animals, and (iii) misses camera pans of 10 to 14 patches, while our FMS follows all three.
    }
    \label{fig:amf_vs_fms_extended}
\end{figure}

In contrast, our FMS effectively mitigates these issues by providing a more stable and object-consistent motion signal, resulting in accurate and robust trajectory estimation even under large motion, as shown in Fig.~\ref{fig:amf_vs_fms}.
FMS builds its trajectories from a flow network trained to find correspondences, and because it is computed outside the generator, one flow field drives CogVideoX-2B, CogVideoX-5B, and Wan2.1 (Appx.~\mbox{%
\ref{app:generalization}}%
), whereas attention-derived signals must be re-derived per backbone and are only as good as that model's attention.
Fig.~\mbox{%
\ref{fig:amf_vs_fms_extended} }\hskip0pt%
extends the comparison of Fig.~\mbox{%
\ref{fig:amf_vs_fms} }\hskip0pt%
to six further videos, two per failure mode.

\subsection{Oracle Flow, Flow Estimator, and Motion Blur}
\label{app:flow_estimator}
We first test whether the accuracy of the optical flow limits FMS.
We run FMS alone, without OCAL, with CogVideoX-5B on the 23 sequences of MPI Sintel \mbox{%
\citep{sintel2012butler}}%
, which provides ground-truth flow, and vary only the flow source.
Since Sintel provides no captions, we caption each sequence once with a vision-language model and use the same caption for all three flow sources, so the captions do not affect the comparison.
MF is 0.503 with ground-truth flow, 0.495 with RAFT \mbox{%
\citep{raft2020teed}}%
, and 0.492 with GMFlow \mbox{%
\citep{gmflow2022xu}}%
.
Ground-truth flow improves MF by only 0.011, so flow quality is not a bottleneck.

We next compare the two flow estimators.
On our benchmark, GMFlow achieves an MF of 0.681 and RAFT 0.667 with unchanged grounding scores, while on Sintel, RAFT leads GMFlow by 0.003.
Neither estimator is consistently better, and we adopt GMFlow for all experiments.

We also examine the effect of motion blur.
Sintel renders every sequence both with motion blur (final pass) and without it (clean pass) under the same ground-truth flow, so the effect of blur can be isolated.
We measure the endpoint error (EPE, in units of the $30 \times 45$ patch grid) of the AMF and FMS patch trajectories against the ground-truth flow over 22 paired sequences.
AMF reaches an EPE of 1.542 on clean frames and 1.485 on blurred frames, whereas FMS reaches 0.092 and 0.251.
FMS is therefore 16.7$\times$ closer to the true flow than AMF on clean frames and 5.9$\times$ on blurred frames, and on blurred frames it remains more accurate than AMF on clean frames for all 22 sequences.
Motion blur reduces the accuracy of FMS, whereas AMF is already too inaccurate for blur to have a measurable effect.

\subsection{Tokenization and Grounding}
\label{app:tokenization}
\begin{figure}[!htb]
    \centering
    \includegraphics[width=1.0\linewidth]{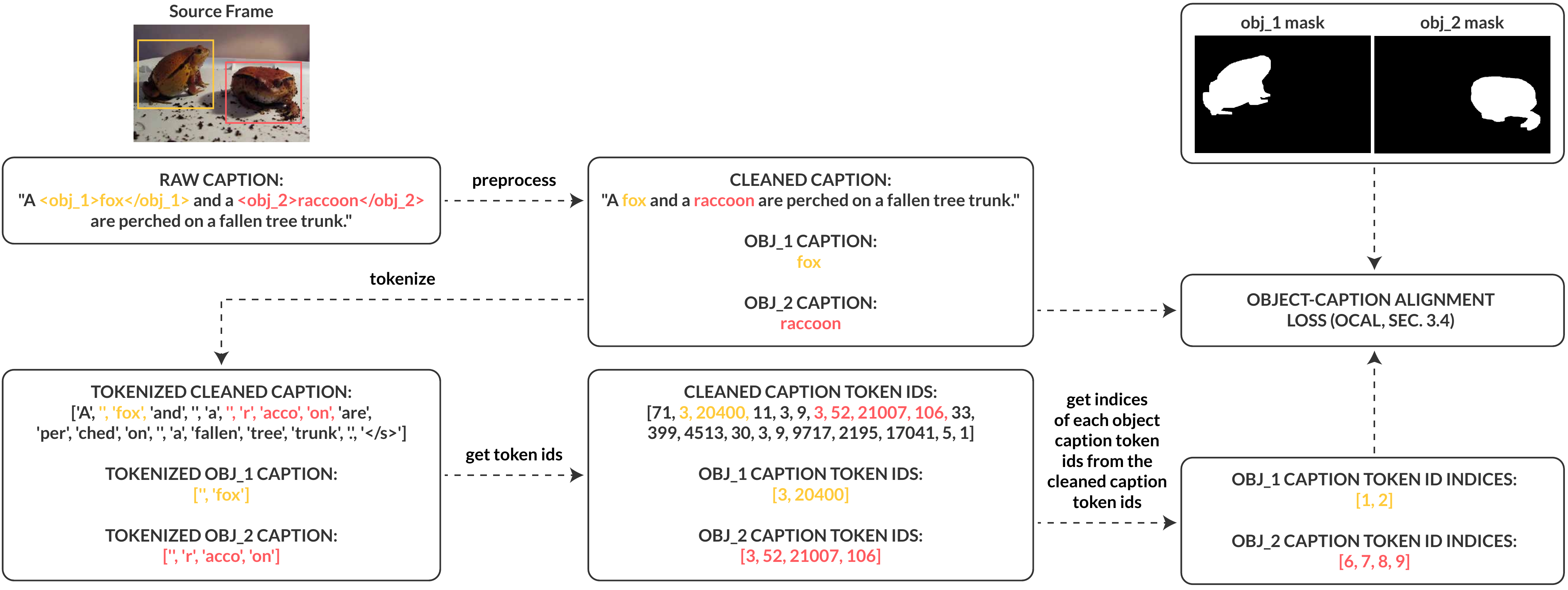}
\vspace{-18.7pt}
    \caption{
        \textbf{Tokenization and grounding.}
        The pipeline tokenizes a tagged raw caption into a cleaned global prompt and individual object captions.
        These tokenized captions are mapped to specific token IDs to identify their positional indices within the global sequence.
        Finally, the indices are paired with their corresponding binary object masks to compute the Object-Caption Alignment Loss (OCAL).
    }
    \label{fig:token_selection}
\end{figure}

Our framework requires a global caption and user-specified object captions which are subsets of the global caption.
We show details of how we tokenize and ground the captions in Fig.~\ref{fig:token_selection}.
Our tokenization pipeline begins with a raw caption containing object tags (e.g., <obj\_n>) that map specific objects to their corresponding binary masks.
The raw caption is preprocessed into a ``cleaned'' natural language caption, from which we obtain the individual object-specific captions.
We then tokenize both the cleaned global caption and the object captions.
Following tokenization, our pipeline maps these captions to their respective token IDs and identifies the specific positional indices of the object caption tokens within the global cleaned caption sequence.
Finally, these localized token indices are paired with their corresponding binary object masks to compute the Object-Caption Alignment Loss (OCAL).
Our pipeline is robust against repeated object captions and subword tokenization, as duplicates are removed and matching is performed on token IDs.
For ambiguous or synonymous terms, the user-defined object captions ensure that grounding is clear.

\clearpage{}
\subsection{MotionGrounder Algorithm}
\label{app:algorithm}
The motion transfer algorithm of our MotionGrounder is shown in Algorithm \mbox{%
\ref{alg:motiongrounder_inference} }\hskip0pt%
below.

\begin{algorithm}[h]
  \caption{MotionGrounder inference pipeline}
  \label{alg:motiongrounder_inference}
  \begin{algorithmic}
    \STATE {\bfseries Input:} Source video $V_s$, optical flow model $G$, trained DiT model $\epsilon_{\theta}$, decoder $\mathcal{D}$, global caption $c_{g}$, object captions $\{c_i\}_{i=1}^N$, object masks $\{m_i^{1:F}\}_{i=1}^N$, positional embedding $\rho$
    \STATE {\bfseries Output:} Generated video $V_T$ with transferred motion

    \STATE Extract and downsample optical flows: $(\hat{f}_{x}, \hat{f}_{y}) \leftarrow \text{Downsample}\!\left(G(V_s)\right)$
    \FOR{each $(x_j, y_j)$ where $j \in [1, J]$}
        \STATE Compute $\text{traj}(x_j, y_j)$ using $(\hat{f}_{x}, \hat{f}_{y})$
        \FOR{each $(x_k, y_k)$ where $(x_k, y_k) \in \text{traj}(x_j, y_j)$ and $k \in [1, J]$}
            \STATE Compute displacement $\Delta_{j \rightarrow k}$
        \ENDFOR
    \ENDFOR
    \STATE Construct $\text{FMS}(z_{ref}) \leftarrow \Delta_{j,k}$

    \STATE Initialize $z_{T} \sim \mathcal{N}(0,I)$
    \STATE Initialize $\rho_{T} = \rho$

    \FOR{denoising step $t=T$ {\bfseries to} $0$}
        \IF{${t > T_{opt}}$}
            \FOR{optimization step $i=0$ {\bfseries to} $I_{opt}$}

                \STATE Extract $Q$ and $K$: $\{ Q, K \} \leftarrow \epsilon_{\theta}(z_{t}, c_{g}, t, \rho_{t})$
                \FOR{each $j, k$ where $j, k \in [1, J]$}
                    \STATE Calculate cross-frame attention $A_{j,k}$
                    \STATE Compute displacement $\tilde{\Delta}_{j \rightarrow k}$
                \ENDFOR
                \STATE Construct $\text{DISP}(z_t) \leftarrow \tilde{\Delta}_{j \rightarrow k}$
                \STATE Get $\mathcal{L}_{\text{FMS}} \leftarrow ||\text{FMS}(z_{ref}) - \text{DISP}(z_{t})||_{2}^{2}$

                \FOR{each $m_i^{1:F}$}
                    \STATE Extract $\tilde{A}_{i}^{t}$
                    \STATE Compute $\mathcal{L}_{\text{FG}}^i \leftarrow (1-\text{sum}(\tilde{A}_{i}^{t} \odot m_i^{1:F}))^2$
                    \STATE Compute $\mathcal{L}_{\text{BG}}^i \leftarrow (\text{sum}(\tilde{A}_{i}^{t} \odot (1-m_i^{1:F})))^2$
                \ENDFOR
                \STATE Compute $\mathcal{L}_{\text{OCAL}} \leftarrow \lambda_{size} \cdot (\mathcal{L}_{\text{FG}} + \mathcal{L}_{\text{BG}})$

                \STATE Get $\mathcal{L}_{t} \leftarrow \lambda_{\text{FMS}} \cdot \mathcal{L}_{\text{FMS}} + \lambda_{\text{OCAL}} \cdot \mathcal{L}_{\text{OCAL}}$

                \STATE Update $z_t$ by minimizing $\mathcal{L}_{t}$
            \ENDFOR
        \ENDIF
        \STATE $z_{t-1} = S(z_t, \epsilon_{\theta}(z_t, c_g, t, \rho))$
    \ENDFOR

    {\bfseries return} $V_{T} = \mathcal{D}(z_{0})$
  \end{algorithmic}
\end{algorithm}

\clearpage
\section{Hyperparameters}
\label{app:hyperparameters}
For our experiments, we compare MotionGrounder against five zero-shot latent optimization-based baselines: DMT \citep{dmt2023yatim}, MOFT \citep{moft2024xiao}, MotionClone \citep{motionclone2024ling}, ConMo \citep{conmo2025gao}, and DiTFlow \citep{ditflow2025pondaven}.
We additionally perform extensive hyperparameter tuning for each baseline using the CogVideoX-5B backbone to ensure a fair comparison.

The evaluated hyperparameters include the learning rate ($l$), transformer block index for motion module insertion ($B$), number of optimization steps ($o$), number of early denoising steps used for optimization ($d$), temperature in Eq.~\ref{eqn:cross_frame_attn} ($\tau$), and mask fusion weight for ConMo ($w$).
The best-performing configurations are summarized below:
\begin{enumerate}
    \item DMT \citep{dmt2023yatim}:
        $l : 0.002 \rightarrow 0.001$,
        $B = 20$,
        $o = 5$,
        $d = 15$,
    \item MOFT \citep{moft2024xiao}:
        $l : 0.002 \rightarrow 0.001$,
        $B = 20$,
        $o = 5$,
        $d = 15$,
    \item MotionClone \citep{motionclone2024ling}:
        $l : 0.002 \rightarrow 0.001$,
        $B = 20$,
        $o = 10$,
        $d = 15$,
    \item ConMo \citep{conmo2025gao}:
        $l : 0.002 \rightarrow 0.001$,
        $B = 20$,
        $o = 5$,
        $d = 15$,
        $w = 0.75$
    \item DiTFlow \citep{ditflow2025pondaven}:
        $l : 0.002 \rightarrow 0.001$,
        $B = 20$,
        $o = 5$,
        $d = 15$,
        $\tau = 2$
    \item MotionGrounder (Ours):
        $l : 0.002 \rightarrow 0.001$,
        $B = 20$,
        $o = 5$,
        $d = 15$,
        $\tau = 2$
\end{enumerate}

\section{Additional Qualitative Results}
\label{app:qualitative}
\subsection{Controllability Demo}
\label{app:controllability}
\begin{figure}[!htb]
    \centering
    \includegraphics[width=1.0\linewidth]{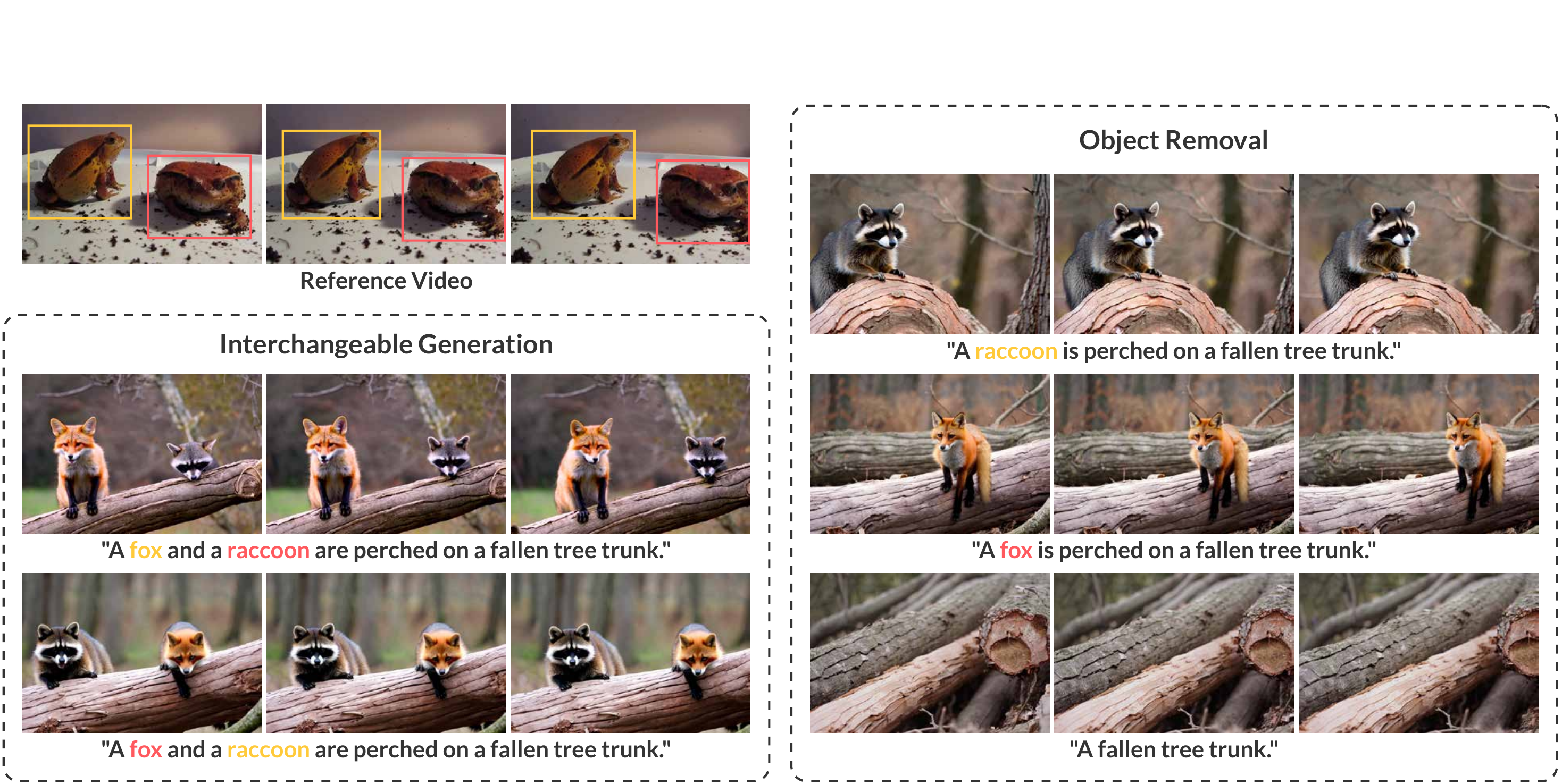}
\vspace{-18.0pt}
    \caption{
        \textbf{Controllability demo.}
        MotionGrounder demonstrates flexible, mask-driven object-level control, enabling seamless object swapping by interchanging masks and object removal by removing masks and corresponding object captions during inference.
    }
    \label{fig:supp_controllability_demo}
\end{figure}

We demonstrate the controllability of MotionGrounder in Fig.~\ref{fig:supp_controllability_demo}.
By interchanging object masks, MotionGrounder enables seamless object swapping while preserving the associated motion patterns.
Object removal is achieved by simply removing the masks and corresponding object captions from the global caption during inference.
These results highlight MotionGrounder's flexible, mask-driven object-level control.

\clearpage{}
\subsection{More Qualitative Comparisons}
\label{app:more_qualitative}
\begin{figure}[!h]
    \centering
    \includegraphics[width=1.0\linewidth]{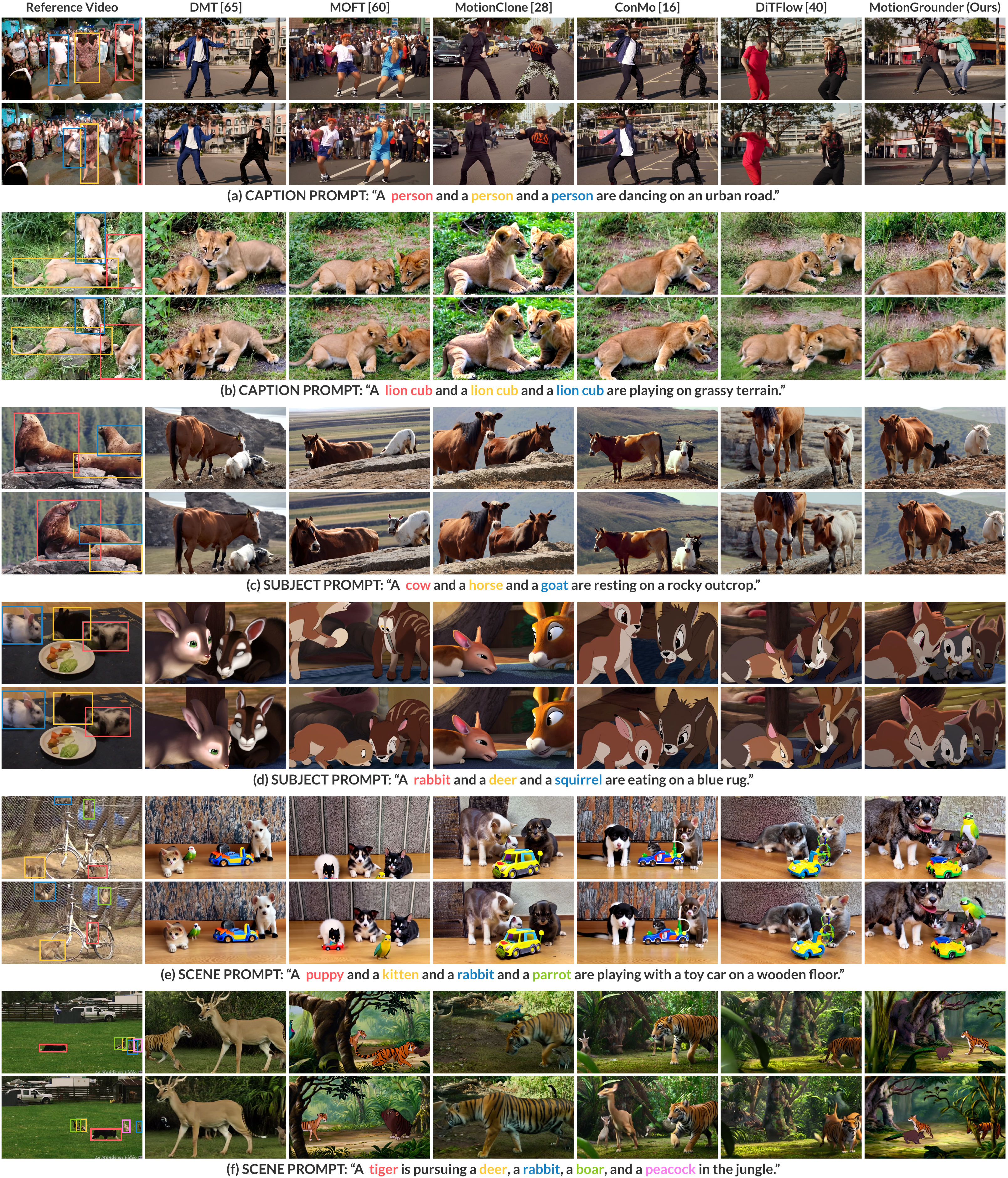}
\vspace{-18.4pt}
    \caption{
        \textbf{Additional qualitative comparisons} on sources with three or more objects ((a) to (d) $N{=}3$, (e) $N{=}4$, (f) $N{=}5$). The reference boxes are the bounding boxes of the ground-truth masks, in the colors of the object words in the prompts.
    }
    \label{fig:supp_more_qualitative}
\end{figure}

Fig.~\mbox{%
\ref{fig:supp_more_qualitative} }\hskip0pt%
extends the comparison of Fig.~\mbox{%
\ref{fig:main_qualitative_comparison} }\hskip0pt%
to six sources with three or more objects: three in (a) to (d), four in (e), and five in (f).
At $N{=}3$ the baselines tend to drop or merge objects: in (c) and (d) several of them render only two of the three animals, whereas our MotionGrounder renders all three.
In (a) and (b), where the caption restates the source, no method, ours included, keeps all three dancers or cubs.
At $N{=}4$ and $N{=}5$, in (e) and (f), every method, including ours, loses objects, as analyzed in Appx.~\mbox{%
\ref{app:limitation}}%
, but our MotionGrounder keeps the objects it renders closest to the reference layout, with the highest IoU and OGS of the six methods on both sources.

\clearpage{}
\subsection{Attention Weight Visualization}
\label{app:attn_viz}
\begin{figure}[!htb]
    \centering
    \includegraphics[width=1.0\linewidth]{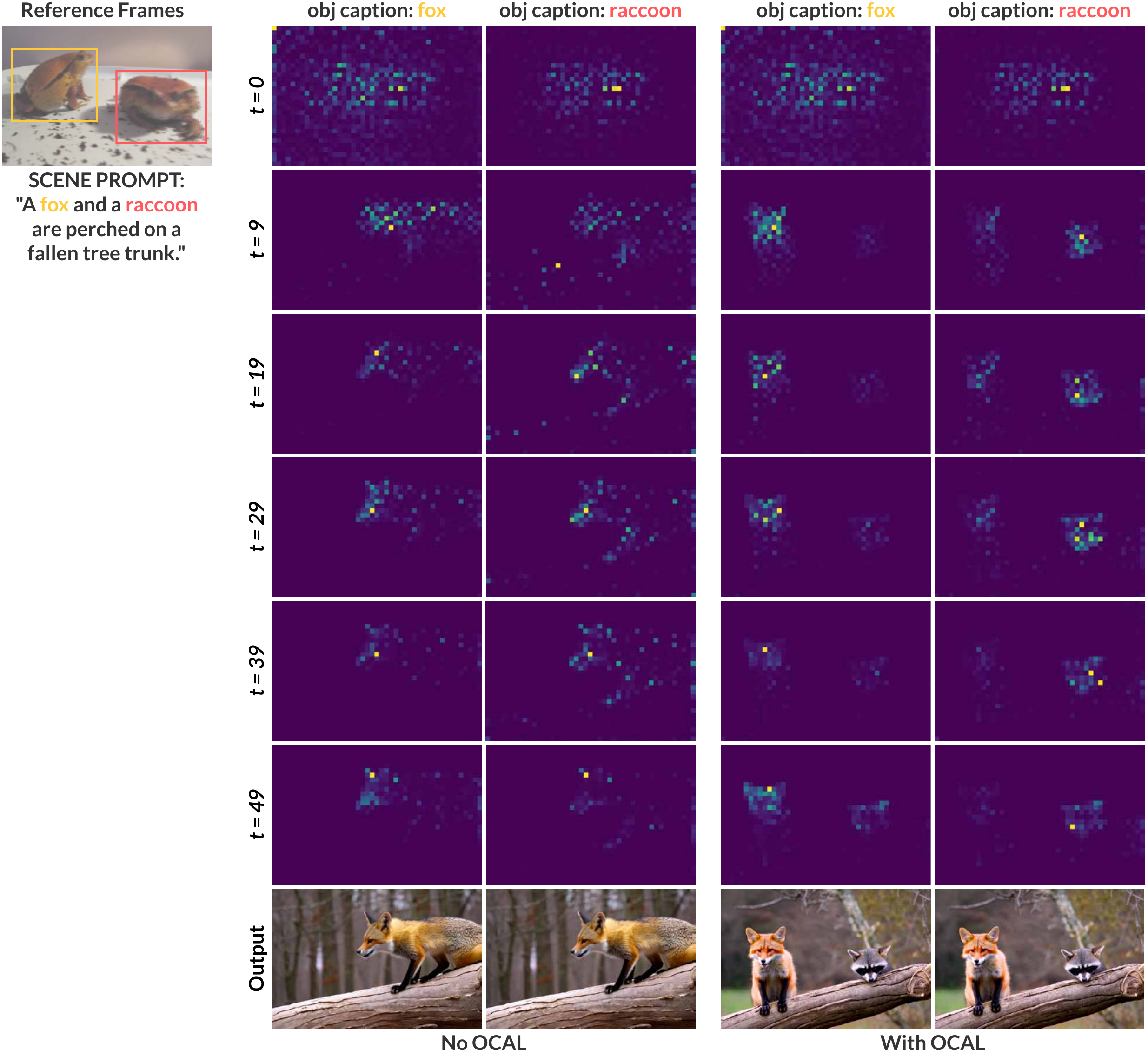}
\vspace{-19.8pt}
    \caption{
        \textbf{Attention weight visualization.}
        We compare the text-to-video attention maps for ``fox'' and ``raccoon'' tokens across diffusion timesteps $t$.
        Without OCAL, attention regions overlap, resulting in a single-object output with fused appearance.
        With OCAL, the model disentangles the attention maps into distinct spatial regions, accurately generating both objects in the prompt.
    }
    \label{fig:attn_viz}
\end{figure}

To evaluate the impact of our Object-Caption Alignment Loss (OCAL) on generation localization, we analyze text-to-video attention maps across multiple timesteps.
Fig.~\ref{fig:attn_viz} illustrates the attention activations for the tokens ``fox'' and ``raccoon'' both without (middle) and with (right) our OCAL.

Without OCAL, the model exhibits a significant failure in semantic grounding, as it is unable to differentiate between the two distinct objects within the scene.
The attention maps for both the ``fox'' and ``raccoon'' tokens overlap substantially, concentrating on a single spatial area rather than segregating into two individual identities.
Consequently, the ``raccoon'' token incorrectly activates the same region as the ``fox,'' leading to object fusion where the model collapses the two entities into a single object in the output frame.

In contrast, the integration of OCAL enforces a sharp spatial divergence between the two tokens.
As early as $t=9$, the attention maps begin to isolate into two distinct regions.
By the final timesteps ($t=39$ to $t=49$), the attention for ``fox'' is localized to the left while ``raccoon'' is localized to the right, aligning perfectly with our intended composition.

\clearpage{}
\subsection{Shape Flexibility in Multi-Object Scenes}
\label{app:shape}
Our MotionGrounder takes masks and optical flow from the source video, yet neither loss constrains shape.
(i) FMS supervises where latent content moves and has no appearance term.
(ii) OCAL only concentrates the caption tokens' attention inside each mask, downsampled to a coarse $30 \times 45$ grid ($16 \times 16$ pixels per cell); generated pixels are never compared with the source.
(iii) Guidance runs only in the first 15 of 50 denoising steps, so placement and motion are set early and the model then realizes a suitable shape.
Our OGS shares this assumption (Appx.~\mbox{%
\ref{app:ogs}}%
): it localizes the generated object instead of reusing the source mask, so it does not reward a copied shape.

DiTFlow \mbox{%
\citep{ditflow2025pondaven} }\hskip0pt%
already discusses the single-object case, so we focus on multi-object scenes.
In Fig.~\mbox{%
\ref{fig:main_figure}}%
, a crew member and a helicopter become a soldier and a tank, a dog with two chimpanzees becomes a wolf, a bear cub, and a rabbit, and four chimpanzees become a red fox, a raccoon, a badger, and a hedgehog, each keeping its own motion.
On the helicopter source (Figs.~\mbox{%
\ref{fig:main_qualitative_comparison} }\hskip0pt%
and~\mbox{%
\ref{fig:ablation_main}}%
), several baselines produce a tank that still carries rotors while ours does not, and our OGS (0.126) is more than double the next best (0.061).
In Fig.~\mbox{%
\ref{fig:ablation_ocal}}%
, the size regularizer keeps the smaller bird from being overpowered by the larger one while both change shape.

\begin{figure}[!h]
    \centering
    \includegraphics[width=1.0\linewidth]{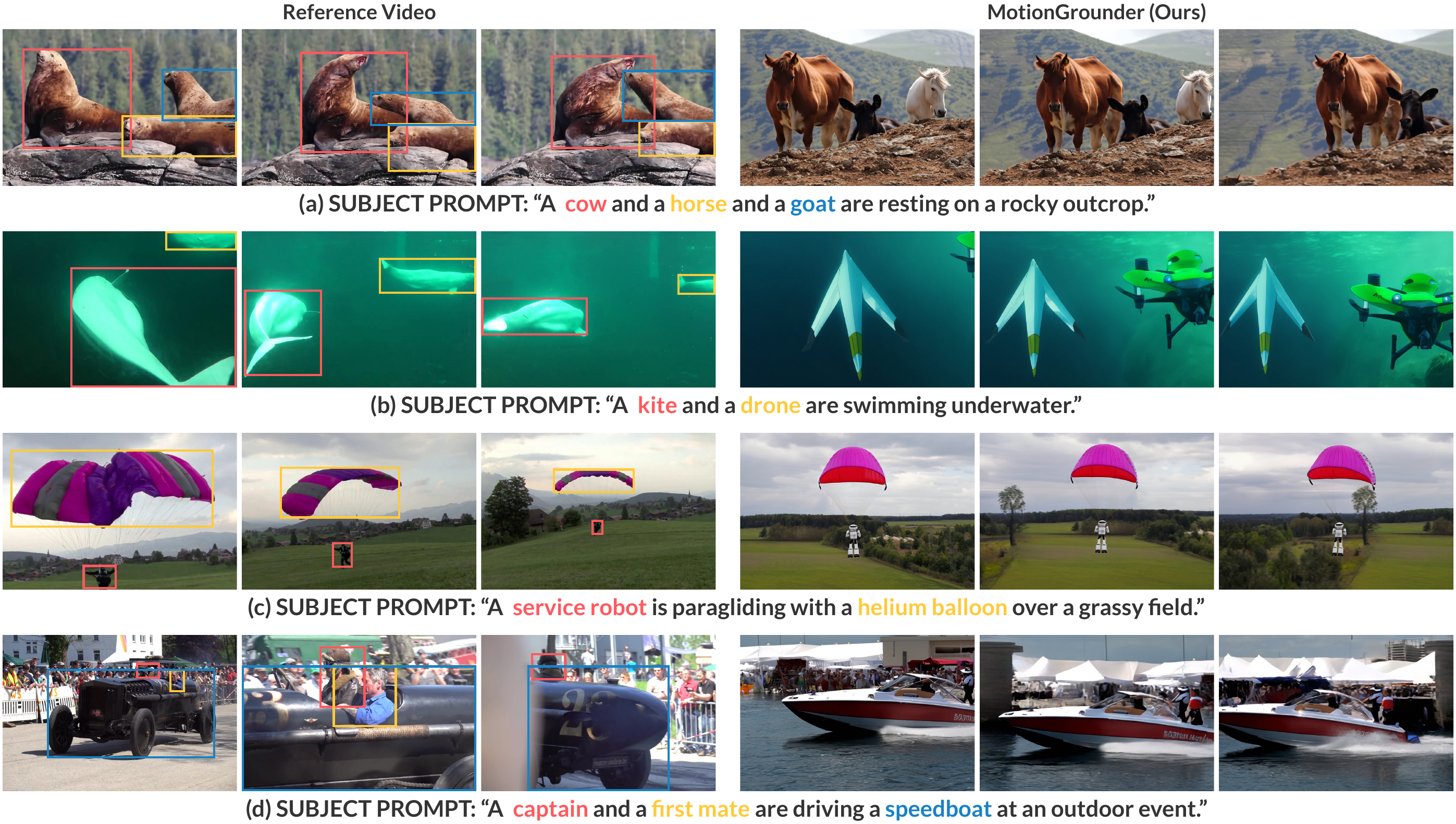}
    \vspace{-6pt}
    \caption{\textbf{Shape flexibility in multi-object scenes.} Left: reference frames 0, 12, and 23 with ground-truth mask boxes; right: our outputs at the same frames under the Subject prompts.}
    \label{fig:shape_flexibility}
\end{figure}

Fig.~\mbox{%
\ref{fig:shape_flexibility} }\hskip0pt%
adds four sources under their Subject prompts: three sea lions become a cow, a horse, and a goat; two beluga whales a kite and a drone; a paraglider a service robot under a helium balloon; and two men in a race car a captain and a first mate in a speedboat.
Each generated object appears in the region of its source object and moves with it, which is all that the mask and the flow constrain.

\clearpage{}

\section{Full Quantitative Results}
\label{app:per_count}
Table~\mbox{%
\ref{tab:main_quantitative_comparison} }\hskip0pt%
of the main paper reports results averaged over the three prompt settings.
Tables~\mbox{%
\ref{tab:per_prompt_caption}}%
, \mbox{%
\ref{tab:per_prompt_subject}}%
, and \mbox{%
\ref{tab:per_prompt_scene} }\hskip0pt%
present the same comparison for the \textit{\textbf{Caption}}, \textit{\textbf{Subject}}, and \textit{\textbf{Scene}} settings separately, over \textit{\textbf{All videos}} and split by the number of objects $N$.

Under every prompt setting, our MotionGrounder achieves the best OGS and MGS on \textit{\textbf{All videos}}.
Our OGS is 0.0763, 0.0667, and 0.0584 under the \textit{\textbf{Caption}}, \textit{\textbf{Subject}}, and \textit{\textbf{Scene}} settings, against 0.0734, 0.0588, and 0.0545 for the best baseline, and our MGS is 0.1970, 0.1896, and 0.1628, against 0.1787, 0.1661, and 0.1399.
Our method also achieves the best OGS and MGS at $N{=}2$ and $N{\geq}4$ in every setting, with the largest margins at $N{=}2$ under the \textit{\textbf{Caption}} and \textit{\textbf{Subject}} settings.
The $N{\geq}4$ bucket contains only 5 videos, and grounding drops sharply there for every method.

Our method does not achieve the best result in every block.
At $N{=}1$ under the \textit{\textbf{Caption}} setting, MOFT and MotionClone obtain a higher OGS (0.0918 and 0.0898 vs.\ 0.0864), and DMT and MOFT a higher MGS (0.2203 and 0.2118 vs.\ 0.2085).
At $N{=}1$ under the \textit{\textbf{Subject}} setting, DMT obtains a higher OGS and MGS (0.0778 vs.\ 0.0711 and 0.2091 vs.\ 0.2065).
At $N{=}3$, MOFT obtains a higher OGS under the \textit{\textbf{Caption}} setting (0.0560 vs.\ 0.0521), and ConMo a higher MGS under the \textit{\textbf{Subject}} setting (0.1326 vs.\ 0.1324).
Finally, every method obtains a lower MF, OGS, and MGS on \textit{\textbf{All videos}} under the \textit{\textbf{Scene}} setting than under the \textit{\textbf{Caption}} setting, as the target prompt departs further from the source video.

\begin{table}[!htb]
  \caption{
    \textbf{Quantitative comparison under the \textit{Caption} setting}, over all videos and split by the number of objects $N$ (number of videos in parentheses).
    Best results in each block are in \textbf{bold} and second best are \underline{underlined}.
  }
  \label{tab:per_prompt_caption}
  \centering
  \fontsize{8pt}{9pt}\selectfont
  \setlength{\tabcolsep}{2.5pt}
  \begin{tabular}{cl*{8}{c}}
    \toprule
    & \textbf{Methods} & MF $\uparrow$ & IoU $\uparrow$ & LTA $\uparrow$ & OGS $\uparrow$ & MGS $\uparrow$ & GTA $\uparrow$ & CTC $\uparrow$ & DTC $\uparrow$ \\
    \midrule
    \multirow{6}{*}{\rotatebox[origin=c]{90}{\shortstack{\textit{\textbf{All videos}}\\(52)}}} & DMT \citep{dmt2023yatim} & \textbf{0.7069} & 0.2172 & 0.2754 & 0.0627 & 0.1703 & \underline{0.3186} & 0.9358 & 0.9172 \\
     & MOFT \citep{moft2024xiao} & 0.5682 & \underline{0.2548} & 0.2767 & \underline{0.0734} & \underline{0.1787} & 0.3165 & 0.9390 & 0.9119 \\
     & MotionClone \citep{motionclone2024ling} & 0.5794 & 0.2302 & \textbf{0.2822} & 0.0676 & 0.1661 & 0.3158 & 0.9395 & \underline{0.9430} \\
     & ConMo \citep{conmo2025gao} & 0.6576 & 0.2292 & 0.2763 & 0.0670 & 0.1747 & 0.3165 & 0.9384 & 0.9104 \\
     & DiTFlow \citep{ditflow2025pondaven} & 0.6762 & 0.2134 & \underline{0.2781} & 0.0625 & 0.1586 & 0.3151 & \underline{0.9414} & 0.9342 \\
     & MotionGrounder (Ours) & \underline{0.6875} & \textbf{0.2638} & 0.2776 & \textbf{0.0763} & \textbf{0.1970} & \textbf{0.3224} & \textbf{0.9426} & \textbf{0.9439} \\
    \midrule
    \multirow{6}{*}{\rotatebox[origin=c]{90}{\shortstack{\textbf{$N{=}1$}\\(21)}}} & DMT \citep{dmt2023yatim} & \textbf{0.6568} & 0.2786 & 0.2758 & 0.0823 & \textbf{0.2203} & 0.3160 & 0.9329 & 0.8894 \\
     & MOFT \citep{moft2024xiao} & 0.5258 & \textbf{0.3089} & 0.2791 & \textbf{0.0918} & \underline{0.2118} & 0.3176 & 0.9375 & 0.9006 \\
     & MotionClone \citep{motionclone2024ling} & 0.5077 & \underline{0.2937} & \textbf{0.2897} & \underline{0.0898} & 0.2027 & 0.3090 & 0.9382 & \underline{0.9340} \\
     & ConMo \citep{conmo2025gao} & 0.5984 & 0.2797 & 0.2809 & 0.0846 & 0.2059 & 0.3171 & 0.9348 & 0.8801 \\
     & DiTFlow \citep{ditflow2025pondaven} & 0.6302 & 0.2737 & 0.2795 & 0.0823 & 0.1984 & \textbf{0.3216} & \underline{0.9390} & 0.9169 \\
     & MotionGrounder (Ours) & \underline{0.6524} & 0.2873 & \underline{0.2811} & 0.0864 & 0.2085 & \underline{0.3207} & \textbf{0.9422} & \textbf{0.9381} \\
    \midrule
    \multirow{6}{*}{\rotatebox[origin=c]{90}{\shortstack{\textbf{$N{=}2$}\\(16)}}} & DMT \citep{dmt2023yatim} & \textbf{0.7269} & 0.2225 & 0.2759 & 0.0626 & 0.1726 & \underline{0.3207} & 0.9397 & 0.9359 \\
     & MOFT \citep{moft2024xiao} & 0.5659 & \underline{0.2675} & 0.2786 & \underline{0.0762} & 0.1923 & 0.3133 & 0.9438 & 0.9297 \\
     & MotionClone \citep{motionclone2024ling} & 0.6374 & 0.2229 & \underline{0.2797} & 0.0632 & 0.1710 & 0.3206 & 0.9426 & 0.9524 \\
     & ConMo \citep{conmo2025gao} & 0.6835 & 0.2493 & 0.2743 & 0.0706 & \underline{0.1951} & 0.3137 & 0.9420 & 0.9254 \\
     & DiTFlow \citep{ditflow2025pondaven} & 0.7043 & 0.2140 & 0.2771 & 0.0613 & 0.1597 & 0.3112 & \underline{0.9450} & \textbf{0.9531} \\
     & MotionGrounder (Ours) & \underline{0.7117} & \textbf{0.3369} & \textbf{0.2812} & \textbf{0.0945} & \textbf{0.2544} & \textbf{0.3263} & \textbf{0.9455} & \underline{0.9527} \\
    \midrule
    \multirow{6}{*}{\rotatebox[origin=c]{90}{\shortstack{\textbf{$N{=}3$}\\(10)}}} & DMT \citep{dmt2023yatim} & \textbf{0.7238} & 0.1658 & \textbf{0.2781} & 0.0469 & 0.1331 & 0.3121 & 0.9343 & 0.9295 \\
     & MOFT \citep{moft2024xiao} & 0.5854 & \textbf{0.2082} & 0.2698 & \textbf{0.0560} & \underline{0.1463} & \underline{0.3150} & 0.9362 & 0.9023 \\
     & MotionClone \citep{motionclone2024ling} & 0.5974 & \underline{0.1896} & \underline{0.2761} & \underline{0.0521} & 0.1393 & 0.3129 & \textbf{0.9390} & \textbf{0.9444} \\
     & ConMo \citep{conmo2025gao} & 0.6883 & 0.1741 & 0.2760 & 0.0490 & 0.1397 & 0.3134 & 0.9374 & 0.9262 \\
     & DiTFlow \citep{ditflow2025pondaven} & 0.6989 & 0.1594 & 0.2730 & 0.0444 & 0.1265 & 0.2995 & \underline{0.9387} & 0.9315 \\
     & MotionGrounder (Ours) & \underline{0.7032} & 0.1880 & 0.2692 & \underline{0.0521} & \textbf{0.1478} & \textbf{0.3158} & \underline{0.9387} & \underline{0.9371} \\
    \midrule
    \multirow{6}{*}{\rotatebox[origin=c]{90}{\shortstack{\textbf{$N{\geq}4$}\\(5)}}} & DMT \citep{dmt2023yatim} & \textbf{0.8199} & 0.0446 & 0.2666 & 0.0124 & 0.0375 & \textbf{0.3359} & 0.9381 & 0.9494 \\
     & MOFT \citep{moft2024xiao} & 0.7193 & \underline{0.0797} & \underline{0.2739} & \underline{0.0225} & \underline{0.0608} & 0.3249 & 0.9359 & 0.9215 \\
     & MotionClone \citep{motionclone2024ling} & 0.6585 & 0.0677 & 0.2704 & 0.0192 & 0.0500 & \underline{0.3351} & 0.9365 & 0.9486 \\
     & ConMo \citep{conmo2025gao} & \underline{0.7619} & 0.0630 & 0.2637 & 0.0172 & 0.0484 & 0.3296 & \underline{0.9442} & \textbf{0.9580} \\
     & DiTFlow \citep{ditflow2025pondaven} & 0.7336 & 0.0668 & \textbf{0.2855} & 0.0199 & 0.0528 & 0.3319 & \textbf{0.9450} & 0.9521 \\
     & MotionGrounder (Ours) & 0.7262 & \textbf{0.0829} & 0.2687 & \textbf{0.0235} & \textbf{0.0629} & 0.3301 & 0.9426 & \underline{0.9535} \\
    \bottomrule
  \end{tabular}
\end{table}

\begin{table}[!htb]
  \caption{
    \textbf{Quantitative comparison under the \textit{Subject} setting}, over all videos and split by the number of objects $N$ (number of videos in parentheses).
    Best results in each block are in \textbf{bold} and second best are \underline{underlined}.
  }
  \label{tab:per_prompt_subject}
  \centering
  \fontsize{8pt}{9pt}\selectfont
  \setlength{\tabcolsep}{2.5pt}
  \begin{tabular}{cl*{8}{c}}
    \toprule
    & \textbf{Methods} & MF $\uparrow$ & IoU $\uparrow$ & LTA $\uparrow$ & OGS $\uparrow$ & MGS $\uparrow$ & GTA $\uparrow$ & CTC $\uparrow$ & DTC $\uparrow$ \\
    \midrule
    \multirow{6}{*}{\rotatebox[origin=c]{90}{\shortstack{\textit{\textbf{All videos}}\\(52)}}} & DMT \citep{dmt2023yatim} & \textbf{0.6840} & \underline{0.2158} & \underline{0.2610} & \underline{0.0588} & \underline{0.1661} & \textbf{0.3183} & 0.9337 & 0.9016 \\
     & MOFT \citep{moft2024xiao} & 0.5331 & 0.1899 & 0.2553 & 0.0514 & 0.1318 & 0.3116 & \underline{0.9379} & 0.8749 \\
     & MotionClone \citep{motionclone2024ling} & 0.5416 & 0.1952 & 0.2538 & 0.0523 & 0.1329 & 0.3137 & 0.9376 & \underline{0.9315} \\
     & ConMo \citep{conmo2025gao} & 0.6436 & 0.1997 & 0.2587 & 0.0545 & 0.1500 & \underline{0.3161} & 0.9347 & 0.8936 \\
     & DiTFlow \citep{ditflow2025pondaven} & 0.6483 & 0.1917 & 0.2587 & 0.0514 & 0.1394 & 0.3155 & 0.9375 & 0.9146 \\
     & MotionGrounder (Ours) & \underline{0.6818} & \textbf{0.2403} & \textbf{0.2617} & \textbf{0.0667} & \textbf{0.1896} & 0.3150 & \textbf{0.9410} & \textbf{0.9335} \\
    \midrule
    \multirow{6}{*}{\rotatebox[origin=c]{90}{\shortstack{\textbf{$N{=}1$}\\(21)}}} & DMT \citep{dmt2023yatim} & \underline{0.6362} & \textbf{0.2789} & \underline{0.2642} & \textbf{0.0778} & \textbf{0.2091} & 0.3191 & 0.9328 & 0.8812 \\
     & MOFT \citep{moft2024xiao} & 0.4981 & 0.2131 & 0.2622 & 0.0599 & 0.1476 & 0.3147 & \underline{0.9378} & 0.8836 \\
     & MotionClone \citep{motionclone2024ling} & 0.4892 & 0.2417 & 0.2608 & 0.0655 & 0.1573 & 0.3095 & 0.9367 & \underline{0.9243} \\
     & ConMo \citep{conmo2025gao} & 0.6197 & 0.2398 & \textbf{0.2646} & 0.0671 & 0.1763 & \textbf{0.3199} & 0.9337 & 0.8840 \\
     & DiTFlow \citep{ditflow2025pondaven} & 0.6042 & 0.2197 & 0.2601 & 0.0603 & 0.1535 & \underline{0.3193} & 0.9345 & 0.9010 \\
     & MotionGrounder (Ours) & \textbf{0.6678} & \underline{0.2478} & 0.2639 & \underline{0.0711} & \underline{0.2065} & 0.3139 & \textbf{0.9392} & \textbf{0.9262} \\
    \midrule
    \multirow{6}{*}{\rotatebox[origin=c]{90}{\shortstack{\textbf{$N{=}2$}\\(16)}}} & DMT \citep{dmt2023yatim} & 0.6956 & 0.2092 & \underline{0.2642} & \underline{0.0559} & \underline{0.1650} & \textbf{0.3217} & 0.9319 & 0.8973 \\
     & MOFT \citep{moft2024xiao} & 0.5565 & 0.2076 & 0.2561 & 0.0548 & 0.1450 & 0.3146 & \underline{0.9410} & 0.8768 \\
     & MotionClone \citep{motionclone2024ling} & 0.5832 & 0.2069 & 0.2486 & 0.0542 & 0.1468 & 0.3178 & 0.9388 & \underline{0.9371} \\
     & ConMo \citep{conmo2025gao} & 0.6656 & 0.2082 & 0.2620 & 0.0552 & 0.1582 & 0.3189 & 0.9347 & 0.8872 \\
     & DiTFlow \citep{ditflow2025pondaven} & \underline{0.7027} & \underline{0.2142} & 0.2619 & 0.0558 & 0.1619 & \underline{0.3204} & 0.9406 & 0.9262 \\
     & MotionGrounder (Ours) & \textbf{0.7212} & \textbf{0.3242} & \textbf{0.2672} & \textbf{0.0875} & \textbf{0.2445} & 0.3202 & \textbf{0.9442} & \textbf{0.9394} \\
    \midrule
    \multirow{6}{*}{\rotatebox[origin=c]{90}{\shortstack{\textbf{$N{=}3$}\\(10)}}} & DMT \citep{dmt2023yatim} & \textbf{0.7086} & 0.1662 & \textbf{0.2613} & 0.0442 & \underline{0.1324} & \underline{0.3109} & 0.9366 & 0.9328 \\
     & MOFT \citep{moft2024xiao} & 0.5612 & \textbf{0.1738} & 0.2522 & \underline{0.0458} & 0.1200 & 0.3030 & 0.9339 & 0.8635 \\
     & MotionClone \citep{motionclone2024ling} & 0.5659 & 0.1461 & 0.2554 & 0.0403 & 0.1061 & \textbf{0.3146} & 0.9390 & \underline{0.9422} \\
     & ConMo \citep{conmo2025gao} & 0.6329 & 0.1683 & 0.2521 & 0.0454 & \textbf{0.1326} & 0.3036 & 0.9344 & 0.9204 \\
     & DiTFlow \citep{ditflow2025pondaven} & 0.6440 & 0.1637 & \underline{0.2595} & 0.0441 & 0.1229 & 0.3018 & \underline{0.9397} & 0.9256 \\
     & MotionGrounder (Ours) & \underline{0.6600} & \underline{0.1723} & 0.2551 & \textbf{0.0480} & \underline{0.1324} & 0.3051 & \textbf{0.9407} & \textbf{0.9436} \\
    \midrule
    \multirow{6}{*}{\rotatebox[origin=c]{90}{\shortstack{\textbf{$N{\geq}4$}\\(5)}}} & DMT \citep{dmt2023yatim} & \textbf{0.7982} & \underline{0.0707} & 0.2362 & \underline{0.0179} & \underline{0.0562} & \underline{0.3189} & 0.9375 & \textbf{0.9386} \\
     & MOFT \citep{moft2024xiao} & 0.5495 & 0.0678 & 0.2300 & 0.0164 & 0.0493 & 0.3059 & 0.9359 & 0.8551 \\
     & MotionClone \citep{motionclone2024ling} & 0.5800 & 0.0605 & 0.2380 & 0.0152 & 0.0395 & 0.3157 & 0.9350 & 0.9222 \\
     & ConMo \citep{conmo2025gao} & \underline{0.6955} & 0.0670 & 0.2365 & 0.0173 & 0.0482 & 0.3156 & \textbf{0.9402} & 0.9007 \\
     & DiTFlow \citep{ditflow2025pondaven} & 0.6686 & 0.0582 & \underline{0.2407} & 0.0144 & 0.0412 & 0.3113 & 0.9355 & 0.9134 \\
     & MotionGrounder (Ours) & 0.6576 & \textbf{0.0767} & \textbf{0.2477} & \textbf{0.0196} & \textbf{0.0571} & \textbf{0.3226} & \underline{0.9393} & \underline{0.9249} \\
    \bottomrule
  \end{tabular}
\end{table}

\clearpage
\begin{table}[!htb]
  \caption{
    \textbf{Quantitative comparison under the \textit{Scene} setting}, over all videos and split by the number of objects $N$ (number of videos in parentheses).
    Best results in each block are in \textbf{bold} and second best are \underline{underlined}.
  }
  \label{tab:per_prompt_scene}
  \centering
  \fontsize{8pt}{9pt}\selectfont
  \setlength{\tabcolsep}{2.5pt}
  \begin{tabular}{cl*{8}{c}}
    \toprule
    & \textbf{Methods} & MF $\uparrow$ & IoU $\uparrow$ & LTA $\uparrow$ & OGS $\uparrow$ & MGS $\uparrow$ & GTA $\uparrow$ & CTC $\uparrow$ & DTC $\uparrow$ \\
    \midrule
    \multirow{6}{*}{\rotatebox[origin=c]{90}{\shortstack{\textit{\textbf{All videos}}\\(52)}}} & DMT \citep{dmt2023yatim} & \textbf{0.6749} & 0.1841 & \underline{0.2606} & 0.0499 & 0.1361 & 0.3095 & 0.9348 & 0.8982 \\
     & MOFT \citep{moft2024xiao} & 0.5337 & 0.1518 & 0.2572 & 0.0411 & 0.1050 & 0.3132 & 0.9391 & 0.8853 \\
     & MotionClone \citep{motionclone2024ling} & 0.5367 & \underline{0.1988} & 0.2579 & \underline{0.0545} & \underline{0.1399} & 0.3085 & 0.9406 & \underline{0.9291} \\
     & ConMo \citep{conmo2025gao} & 0.6496 & 0.1610 & 0.2577 & 0.0444 & 0.1202 & \textbf{0.3178} & 0.9366 & 0.8907 \\
     & DiTFlow \citep{ditflow2025pondaven} & 0.6667 & 0.1552 & 0.2604 & 0.0420 & 0.1102 & 0.3161 & \underline{0.9409} & 0.9288 \\
     & MotionGrounder (Ours) & \underline{0.6746} & \textbf{0.2073} & \textbf{0.2637} & \textbf{0.0584} & \textbf{0.1628} & \underline{0.3173} & \textbf{0.9415} & \textbf{0.9362} \\
    \midrule
    \multirow{6}{*}{\rotatebox[origin=c]{90}{\shortstack{\textbf{$N{=}1$}\\(21)}}} & DMT \citep{dmt2023yatim} & 0.6358 & \underline{0.2083} & 0.2617 & 0.0575 & \underline{0.1480} & 0.3025 & 0.9312 & 0.8744 \\
     & MOFT \citep{moft2024xiao} & 0.5346 & 0.1663 & 0.2595 & 0.0456 & 0.1179 & 0.3092 & 0.9391 & 0.8875 \\
     & MotionClone \citep{motionclone2024ling} & 0.4914 & 0.2072 & 0.2542 & \underline{0.0588} & 0.1455 & 0.3008 & \underline{0.9402} & \underline{0.9280} \\
     & ConMo \citep{conmo2025gao} & \underline{0.6584} & 0.1834 & 0.2615 & 0.0520 & 0.1302 & 0.3121 & 0.9345 & 0.8941 \\
     & DiTFlow \citep{ditflow2025pondaven} & 0.6144 & 0.1794 & \textbf{0.2680} & 0.0506 & 0.1248 & \textbf{0.3154} & 0.9394 & 0.9218 \\
     & MotionGrounder (Ours) & \textbf{0.6618} & \textbf{0.2111} & \underline{0.2661} & \textbf{0.0613} & \textbf{0.1735} & \underline{0.3146} & \textbf{0.9423} & \textbf{0.9309} \\
    \midrule
    \multirow{6}{*}{\rotatebox[origin=c]{90}{\shortstack{\textbf{$N{=}2$}\\(16)}}} & DMT \citep{dmt2023yatim} & 0.6967 & 0.2102 & \underline{0.2680} & 0.0562 & 0.1590 & 0.3147 & 0.9373 & 0.9190 \\
     & MOFT \citep{moft2024xiao} & 0.5404 & 0.1792 & 0.2660 & 0.0481 & 0.1269 & 0.3174 & 0.9417 & 0.8970 \\
     & MotionClone \citep{motionclone2024ling} & 0.5689 & \underline{0.2691} & 0.2662 & \underline{0.0724} & \underline{0.1914} & 0.3078 & 0.9401 & 0.9215 \\
     & ConMo \citep{conmo2025gao} & 0.6855 & 0.1868 & 0.2601 & 0.0496 & 0.1430 & \underline{0.3221} & 0.9387 & 0.8784 \\
     & DiTFlow \citep{ditflow2025pondaven} & \textbf{0.7406} & 0.1756 & 0.2624 & 0.0455 & 0.1284 & 0.3186 & \textbf{0.9433} & \textbf{0.9387} \\
     & MotionGrounder (Ours) & \underline{0.7091} & \textbf{0.2699} & \textbf{0.2701} & \textbf{0.0750} & \textbf{0.2069} & \textbf{0.3259} & \underline{0.9424} & \underline{0.9386} \\
    \midrule
    \multirow{6}{*}{\rotatebox[origin=c]{90}{\shortstack{\textbf{$N{=}3$}\\(10)}}} & DMT \citep{dmt2023yatim} & \textbf{0.6680} & \underline{0.1539} & 0.2572 & \underline{0.0408} & \underline{0.1202} & 0.3138 & 0.9356 & 0.8987 \\
     & MOFT \citep{moft2024xiao} & 0.4848 & 0.1239 & 0.2459 & 0.0335 & 0.0766 & 0.3061 & 0.9365 & 0.8627 \\
     & MotionClone \citep{motionclone2024ling} & 0.5651 & 0.1451 & \textbf{0.2599} & 0.0380 & 0.1016 & \textbf{0.3189} & \textbf{0.9422} & \underline{0.9386} \\
     & ConMo \citep{conmo2025gao} & 0.6229 & 0.1300 & 0.2548 & 0.0361 & 0.1079 & \underline{0.3177} & 0.9384 & 0.9028 \\
     & DiTFlow \citep{ditflow2025pondaven} & \underline{0.6528} & 0.1266 & 0.2566 & 0.0335 & 0.0893 & 0.3123 & \underline{0.9416} & \underline{0.9386} \\
     & MotionGrounder (Ours) & 0.6297 & \textbf{0.1621} & \underline{0.2577} & \textbf{0.0442} & \textbf{0.1227} & 0.3001 & 0.9393 & \textbf{0.9445} \\
    \midrule
    \multirow{6}{*}{\rotatebox[origin=c]{90}{\shortstack{\textbf{$N{\geq}4$}\\(5)}}} & DMT \citep{dmt2023yatim} & \textbf{0.7831} & \underline{0.0598} & 0.2395 & \underline{0.0154} & \underline{0.0446} & 0.3138 & \textbf{0.9405} & 0.9312 \\
     & MOFT \citep{moft2024xiao} & 0.6057 & 0.0592 & 0.2419 & 0.0146 & 0.0376 & \underline{0.3302} & 0.9366 & 0.8841 \\
     & MotionClone \citep{motionclone2024ling} & 0.5670 & 0.0462 & \underline{0.2429} & 0.0119 & 0.0297 & 0.3220 & \underline{0.9404} & \textbf{0.9395} \\
     & ConMo \citep{conmo2025gao} & 0.5516 & 0.0469 & 0.2396 & 0.0117 & 0.0301 & 0.3287 & 0.9348 & 0.8915 \\
     & DiTFlow \citep{ditflow2025pondaven} & 0.6773 & 0.0458 & 0.2299 & 0.0119 & 0.0323 & 0.3187 & 0.9385 & 0.9066 \\
     & MotionGrounder (Ours) & \underline{0.7076} & \textbf{0.0812} & \textbf{0.2453} & \textbf{0.0214} & \textbf{0.0592} & \textbf{0.3354} & 0.9397 & \underline{0.9343} \\
    \bottomrule
  \end{tabular}
\end{table}

\section{Full Ablation Study Results}
\label{app:full_ablations}

The main paper reports our ablations over \textit{\textbf{All videos}} only.
Tables~\ref{tab:ablation_main_full}, \ref{tab:ablation_fms_full}, and \mbox{%
\ref{tab:ablation_ocal_full} }\hskip0pt%
give the full breakdown per prompt setting (\textit{\textbf{Caption}}, \textit{\textbf{Subject}}, \textit{\textbf{Scene}}, and \textit{\textbf{All}}), Table~\mbox{%
\ref{tab:per_count_full} }\hskip0pt%
splits the essential components by object count, and Table~\mbox{%
\ref{tab:ablation_ocal_no_fms} }\hskip0pt%
repeats the OCAL ablation without FMS.

\subsection{Ablation of Essential Components}
\label{app:main_ablation}
Table~\ref{tab:ablation_main_full} shows the complementary roles of our two components in every prompt setting: FMS raises MF, OCAL raises IoU, LTA, OGS, and MGS, and our full model has the best MF and the best or second-best IoU, LTA, OGS, and MGS in every setting.
The per-setting results follow the \textit{\textbf{All}} averages of Table~\mbox{%
\ref{tab:ablation_main}}%
.

Paired Wilcoxon signed-rank tests over all 156 video-prompt cells confirm that each component is significant at what it is designed for.
From (a) to (d), MF improves by 60.0\% ($p = 3 \times 10^{-25}$, better on 144 of 156 cells), IoU by 48.8\% ($p = 6 \times 10^{-10}$), and OGS by 50.4\% ($p = 9 \times 10^{-11}$).
FMS supplies motion: from (a) to (b), MF improves by 57.7\% ($p = 5 \times 10^{-23}$) with grounding unchanged (IoU $+3.7\%$, $p = 0.55$).
OCAL supplies grounding: from (b) to (d), IoU improves by 43.5\% and OGS by 45.2\% (both $p < 10^{-15}$).
Removing FMS from our full model, (d) to (c), lowers MF by 40.2\% ($p = 4 \times 10^{-20}$).
The one column where a component is not significant, FMS on grounding, is exactly where it is not meant to act.

\begin{table}[!htb]
\caption{
        \textbf{Ablation of essential components.} Full breakdown of Table~\mbox{%
\ref{tab:ablation_main} }\hskip0pt%
per prompt setting. FMS improves MF, OCAL improves grounding, and our full model has the best MF and the second-best IoU, OGS, and MGS in every setting.
    }
  \label{tab:ablation_main_full}
  \centering
  \fontsize{8pt}{9pt}\selectfont
  \setlength{\tabcolsep}{4pt}
  \begin{tabular}{cl*{8}{c}}
    \toprule
    & \textbf{Method} & MF $\uparrow$ & IoU $\uparrow$ & LTA $\uparrow$ & OGS $\uparrow$ & MGS $\uparrow$ & GTA $\uparrow$ & CTC $\uparrow$ & DTC $\uparrow$ \\
    \midrule
    \multirow{4}{*}{\rotatebox[origin=c]{90}{\textbf{\textit{Caption}}}} & (a) Backbone & 0.4288 & 0.1749 & 0.2770 & 0.0504 & 0.1094 & 0.3158 & \textbf{0.9434} & \textbf{0.9463} \\
     & (b) FMS only & \underline{0.6846} & 0.1863 & 0.2763 & 0.0541 & 0.1409 & 0.3140 & 0.9426 & 0.9408 \\
     & (c) OCAL only & 0.5011 & \textbf{0.3278} & \textbf{0.2789} & \textbf{0.0945} & \textbf{0.2230} & \underline{0.3202} & \underline{0.9430} & \underline{0.9444} \\
     & (d) Ours, Full & \textbf{0.6875} & \underline{0.2638} & \underline{0.2776} & \underline{0.0763} & \underline{0.1970} & \textbf{0.3224} & 0.9426 & 0.9439 \\
    \midrule
    \multirow{4}{*}{\rotatebox[origin=c]{90}{\textbf{\textit{Subject}}}} & (a) Backbone & 0.4306 & 0.1527 & 0.2608 & 0.0415 & 0.0975 & \underline{0.3204} & \textbf{0.9431} & \textbf{0.9400} \\
     & (b) FMS only & \underline{0.6730} & 0.1652 & 0.2532 & 0.0450 & 0.1232 & 0.3082 & 0.9414 & 0.9312 \\
     & (c) OCAL only & 0.4964 & \textbf{0.2993} & \textbf{0.2633} & \textbf{0.0827} & \textbf{0.1976} & \textbf{0.3262} & \underline{0.9419} & \underline{0.9374} \\
     & (d) Ours, Full & \textbf{0.6818} & \underline{0.2403} & \underline{0.2617} & \underline{0.0667} & \underline{0.1896} & 0.3150 & 0.9410 & 0.9335 \\
    \midrule
    \multirow{4}{*}{\rotatebox[origin=c]{90}{\textbf{\textit{Scene}}}} & (a) Backbone & 0.4178 & 0.1502 & 0.2609 & 0.0420 & 0.0956 & 0.3170 & \underline{0.9425} & 0.9316 \\
     & (b) FMS only & \underline{0.6566} & 0.1442 & 0.2605 & 0.0397 & 0.1049 & 0.3153 & 0.9416 & 0.9296 \\
     & (c) OCAL only & 0.4603 & \textbf{0.3027} & \underline{0.2629} & \textbf{0.0833} & \textbf{0.2009} & \textbf{0.3206} & \textbf{0.9440} & \textbf{0.9450} \\
     & (d) Ours, Full & \textbf{0.6746} & \underline{0.2073} & \textbf{0.2637} & \underline{0.0584} & \underline{0.1628} & \underline{0.3173} & 0.9415 & \underline{0.9362} \\
    \midrule
    \multirow{4}{*}{\rotatebox[origin=c]{90}{\textbf{\textit{All}}}} & (a) Backbone & 0.4257 & 0.1593 & 0.2662 & 0.0446 & 0.1008 & 0.3177 & \textbf{0.9430} & \underline{0.9393} \\
     & (b) FMS only & \underline{0.6714} & 0.1652 & 0.2634 & 0.0462 & 0.1230 & 0.3125 & \underline{0.9419} & 0.9339 \\
     & (c) OCAL only & 0.4860 & \textbf{0.3099} & \textbf{0.2684} & \textbf{0.0868} & \textbf{0.2072} & \textbf{0.3223} & \textbf{0.9430} & \textbf{0.9422} \\
     & (d) Ours, Full & \textbf{0.6813} & \underline{0.2371} & \underline{0.2677} & \underline{0.0671} & \underline{0.1832} & \underline{0.3182} & 0.9417 & 0.9379 \\
    \bottomrule
    \end{tabular}
\end{table}

Table~\mbox{%
\ref{tab:per_count_full} }\hskip0pt%
splits the essential components of Table~\mbox{%
\ref{tab:ablation_main} }\hskip0pt%
by object count.
LTA, GTA, CTC, and DTC are nearly constant across methods and counts (LTA 0.2466--0.2728, GTA 0.3045--0.3357, CTC 0.9323--0.9466, DTC 0.8762--0.9502), so IoU drives OGS.
Our losses do not degrade the video quality of the backbone: GTA, CTC, and DTC of our full model stay within 0.02 of the backbone at every count.
OCAL alone tops IoU and OGS at every count because those metrics measure placement, not motion: with OCAL alone almost nothing moves (MF 0.4513--0.5285, close to the backbone's 0.3932--0.4505), and a nearly static object inside its mask region is easy to detect and ground, whereas forcing a reshaped object along the source trajectory makes grounding harder.
On a motion transfer benchmark, OCAL alone is therefore not a usable operating point.

\begin{table}[!htb]
  \caption{
    \textbf{Essential components by object count.} The four configurations of Table~\ref{tab:ablation_main}, averaged over the three prompt settings and split by the number of objects $N$ in the source video (number of videos in parentheses). FMS improves MF at every count while leaving grounding almost unchanged; OCAL improves IoU and OGS at every count, and adding it to FMS raises MF at $N{\leq}3$ and lowers it by 0.019 at $N{\geq}4$. Best results in each block are in \textbf{bold}.
  }
  \label{tab:per_count_full}
  \centering
  \fontsize{8pt}{9pt}\selectfont
  \setlength{\tabcolsep}{2pt}

  \begin{tabular}{cl*{8}{c}}
    \toprule
    & \textbf{Method} & MF $\uparrow$ & IoU $\uparrow$ & LTA $\uparrow$ & OGS $\uparrow$ & MGS $\uparrow$ & GTA $\uparrow$ & CTC $\uparrow$ & DTC $\uparrow$ \\
    \midrule
    \multirow{4}{*}{\rotatebox[origin=c]{90}{\shortstack{\textbf{$N{=}1$}\\(21)}}} & (a) Backbone & 0.4505 & 0.1801 & 0.2704 & 0.0527 & 0.1155 & \textbf{0.3189} & \textbf{0.9466} & \textbf{0.9502} \\
     & (b) FMS only & 0.6382 & 0.1892 & 0.2652 & 0.0553 & 0.1381 & 0.3093 & 0.9402 & 0.9239 \\
     & (c) OCAL only & 0.4618 & \textbf{0.3669} & \textbf{0.2720} & \textbf{0.1040} & \textbf{0.2475} & 0.3150 & 0.9446 & 0.9439 \\
     & (d) Ours, Full & \textbf{0.6607} & 0.2487 & 0.2704 & 0.0729 & 0.1965 & 0.3164 & 0.9412 & 0.9318 \\
    \midrule
    \multirow{4}{*}{\rotatebox[origin=c]{90}{\shortstack{\textbf{$N{=}2$}\\(16)}}} & (a) Backbone & 0.4203 & 0.1764 & 0.2654 & 0.0474 & 0.1126 & 0.3194 & 0.9392 & 0.9311 \\
     & (b) FMS only & 0.7077 & 0.1866 & 0.2657 & 0.0499 & 0.1400 & 0.3146 & \textbf{0.9455} & \textbf{0.9444} \\
     & (c) OCAL only & 0.5285 & \textbf{0.3649} & 0.2727 & \textbf{0.1019} & \textbf{0.2438} & \textbf{0.3335} & 0.9434 & 0.9442 \\
     & (d) Ours, Full & \textbf{0.7140} & 0.3103 & \textbf{0.2728} & 0.0857 & 0.2353 & 0.3241 & 0.9440 & 0.9436 \\
    \midrule
    \multirow{4}{*}{\rotatebox[origin=c]{90}{\shortstack{\textbf{$N{=}3$}\\(10)}}} & (a) Backbone & 0.3986 & 0.1391 & \textbf{0.2651} & 0.0381 & 0.0840 & \textbf{0.3143} & \textbf{0.9433} & \textbf{0.9428} \\
     & (b) FMS only & 0.6609 & 0.1308 & 0.2606 & 0.0358 & 0.1013 & 0.3111 & 0.9403 & 0.9385 \\
     & (c) OCAL only & 0.4859 & \textbf{0.2076} & 0.2610 & \textbf{0.0566} & \textbf{0.1385} & 0.3130 & 0.9402 & 0.9409 \\
     & (d) Ours, Full & \textbf{0.6643} & 0.1741 & 0.2607 & 0.0481 & 0.1343 & 0.3070 & 0.9395 & 0.9417 \\
    \midrule
    \multirow{4}{*}{\rotatebox[origin=c]{90}{\shortstack{\textbf{$N{\geq}4$}\\(5)}}} & (a) Backbone & 0.3932 & 0.0573 & 0.2536 & 0.0151 & 0.0353 & 0.3144 & 0.9394 & 0.9127 \\
     & (b) FMS only & \textbf{0.7159} & 0.0650 & 0.2539 & 0.0172 & 0.0488 & 0.3221 & 0.9404 & 0.9329 \\
     & (c) OCAL only & 0.4513 & \textbf{0.0998} & \textbf{0.2542} & \textbf{0.0269} & 0.0577 & \textbf{0.3357} & \textbf{0.9405} & 0.9315 \\
     & (d) Ours, Full & 0.6972 & 0.0803 & 0.2539 & 0.0215 & \textbf{0.0597} & 0.3293 & \textbf{0.9405} & \textbf{0.9376} \\
    \bottomrule
    \end{tabular}

\end{table}

\subsection{FMS Ablation}
\label{app:fms_ablation}
Table~\ref{tab:ablation_fms_full} analyzes our FMS and its trajectory duplication across the \textit{\textbf{Caption}}, \textit{\textbf{Subject}}, \textit{\textbf{Scene}}, and \textit{\textbf{All}} settings.
Removing FMS in (a) gives the lowest MF in every setting: OCAL alone grounds the objects but does not preserve the reference motion.
Adding FMS in (b) and (c) raises MF by 0.18 to 0.22 in every setting.
Trajectory duplication, (b) to (c), raises the scores: over \textit{\textbf{All videos}} MF rises by 0.004, IoU, OGS, and MGS by 0.006, 0.002, and 0.010, and DTC falls by 0.003; under the \textit{\textbf{Scene}} setting (b) is ahead in MF, IoU, and OGS.
Its effect is clearest visually: without duplication, one-to-one correspondences make patches compete and some stay still, so object motion is suppressed (Fig.~\mbox{%
\ref{fig:ablation_fms}}\hskip0pt%
-(b)), whereas sharing trajectories lets moving patches carry their neighbours along (Fig.~\mbox{%
\ref{fig:ablation_fms}}\hskip0pt%
-(c)).
The grounding scores of (b) and (c) stay below the OCAL-only row (a) because FMS moves the objects along the source trajectory (Appx.~\mbox{%
\ref{app:main_ablation}}%
).

\begin{table}[!htb]
\caption{
        \textbf{FMS ablation.} Full breakdown of Fig.~\mbox{%
\ref{fig:ablation_fms} }\hskip0pt%
per prompt setting. FMS raises MF in every setting, and trajectory duplication, (b) to (c), further raises MF, IoU, LTA, OGS, and MGS over \textit{\textbf{All videos}} and keeps the objects moving (Fig.~\mbox{%
\ref{fig:ablation_fms}}\hskip0pt%
).
    }
  \label{tab:ablation_fms_full}
  \centering
  \fontsize{8pt}{9pt}\selectfont
  \setlength{\tabcolsep}{4pt}
  \begin{tabular}{cl*{8}{c}}
    \toprule
    & \textbf{Method} & MF $\uparrow$ & IoU $\uparrow$ & LTA $\uparrow$ & OGS $\uparrow$ & MGS $\uparrow$ & GTA $\uparrow$ & CTC $\uparrow$ & DTC $\uparrow$ \\
    \midrule
    \multirow{3}{*}{\rotatebox[origin=c]{90}{\textbf{\textit{Caption}}}} & (a) no FMS & 0.5011 & \textbf{0.3278} & \textbf{0.2789} & \textbf{0.0945} & \textbf{0.2230} & \underline{0.3202} & \textbf{0.9430} & \underline{0.9444} \\
     & (b) w/o traj dup & \underline{0.6814} & 0.2574 & \underline{0.2781} & 0.0740 & 0.1941 & 0.3194 & 0.9417 & \textbf{0.9484} \\
     & (c) w/ traj dup (Ours, Full) & \textbf{0.6875} & \underline{0.2638} & 0.2776 & \underline{0.0763} & \underline{0.1970} & \textbf{0.3224} & \underline{0.9426} & 0.9439 \\
    \midrule
    \multirow{3}{*}{\rotatebox[origin=c]{90}{\textbf{\textit{Subject}}}} & (a) no FMS & 0.4964 & \textbf{0.2993} & \textbf{0.2633} & \textbf{0.0827} & \textbf{0.1976} & \textbf{0.3262} & \textbf{0.9419} & \textbf{0.9374} \\
     & (b) w/o traj dup & \underline{0.6737} & 0.2244 & 0.2604 & 0.0621 & 0.1657 & \underline{0.3187} & 0.9404 & \underline{0.9369} \\
     & (c) w/ traj dup (Ours, Full) & \textbf{0.6818} & \underline{0.2403} & \underline{0.2617} & \underline{0.0667} & \underline{0.1896} & 0.3150 & \underline{0.9410} & 0.9335 \\
    \midrule
    \multirow{3}{*}{\rotatebox[origin=c]{90}{\textbf{\textit{Scene}}}} & (a) no FMS & 0.4603 & \textbf{0.3027} & \underline{0.2629} & \textbf{0.0833} & \textbf{0.2009} & \textbf{0.3206} & \textbf{0.9440} & \textbf{0.9450} \\
     & (b) w/o traj dup & \textbf{0.6767} & \underline{0.2130} & 0.2612 & \underline{0.0597} & 0.1611 & 0.3154 & \underline{0.9434} & \underline{0.9376} \\
     & (c) w/ traj dup (Ours, Full) & \underline{0.6746} & 0.2073 & \textbf{0.2637} & 0.0584 & \underline{0.1628} & \underline{0.3173} & 0.9415 & 0.9362 \\
    \midrule
    \multirow{3}{*}{\rotatebox[origin=c]{90}{\textbf{\textit{All}}}} & (a) no FMS & 0.4860 & \textbf{0.3099} & \textbf{0.2684} & \textbf{0.0868} & \textbf{0.2072} & \textbf{0.3223} & \textbf{0.9430} & \textbf{0.9422} \\
     & (b) w/o traj dup & \underline{0.6772} & 0.2316 & 0.2665 & 0.0653 & 0.1736 & 0.3179 & \underline{0.9418} & \underline{0.9410} \\
     & (c) w/ traj dup (Ours, Full) & \textbf{0.6813} & \underline{0.2371} & \underline{0.2677} & \underline{0.0671} & \underline{0.1832} & \underline{0.3182} & 0.9417 & 0.9379 \\
    \bottomrule
    \end{tabular}
\end{table}

\subsection{OCAL Ablation}
\label{app:ocal_ablation}
Table~\ref{tab:ablation_ocal_full} presents the full ablation of OCAL.
In every prompt setting, every OCAL variant (b) to (g) improves IoU, OGS, and MGS over (a), while LTA moves by at most 0.009 and not always upward.
The size regularizer alone, (d) to (e), buys grounding (IoU 0.2375 to 0.2475) at a small cost in motion (MF 0.6791 to 0.6775). Repetition alone has no measurable effect ((d) to (f): $p = 0.84$ / $0.58$ / $0.34$ for MF / IoU / OGS in the \textit{\textbf{All}} setting), while adding it on top of the size regularizer, (e) to (g), wins the motion back (MF 0.6813) and gives the grounding gain back (IoU 0.2371). Neither step is significant in MF ($p = 0.70$ and $0.19$), IoU, or OGS, so (g) over (e) is a trade rather than a gain, and we report (e) as an equally valid operating point.

\begin{table}[!htb]
\caption{
        \textbf{OCAL ablation.} OCAL improves IoU, OGS, and MGS in every prompt setting; repetition and the size regularizer trade motion against grounding within noise, so (e) and (g) are equally valid operating points.
    }
  \label{tab:ablation_ocal_full}
  \centering
  \fontsize{8pt}{9pt}\selectfont
  \setlength{\tabcolsep}{4pt}
  \begin{tabular}{cl*{8}{c}}
    \toprule
    & \textbf{Method} & MF $\uparrow$ & IoU $\uparrow$ & LTA $\uparrow$ & OGS $\uparrow$ & MGS $\uparrow$ & GTA $\uparrow$ & CTC $\uparrow$ & DTC $\uparrow$ \\
    \midrule
    \multirow{7}{*}{\rotatebox[origin=c]{90}{\textbf{\textit{Caption}}}} & (a) no OCAL & 0.6846 & 0.1863 & 0.2763 & 0.0541 & 0.1409 & 0.3140 & 0.9426 & 0.9408 \\
     & (b) text-to-vid only & 0.6815 & 0.2601 & \textbf{0.2796} & 0.0752 & 0.1932 & 0.3206 & 0.9426 & 0.9432 \\
     & (c) vid-to-text only & 0.6744 & 0.2441 & 0.2752 & 0.0704 & 0.1817 & 0.3160 & \underline{0.9430} & \underline{0.9437} \\
     & (d) (b) + (c) & 0.6800 & \underline{0.2672} & \underline{0.2779} & \underline{0.0766} & \underline{0.2018} & 0.3209 & \textbf{0.9431} & 0.9424 \\
     & (e) (d) + size reg & \underline{0.6860} & \textbf{0.2808} & 0.2777 & \textbf{0.0805} & \textbf{0.2119} & 0.3192 & 0.9424 & 0.9406 \\
     & (f) (d) + rep & 0.6832 & 0.2628 & 0.2770 & 0.0752 & 0.1976 & \underline{0.3219} & 0.9425 & 0.9399 \\
     & (g) (e) + rep (Ours, Full) & \textbf{0.6875} & 0.2638 & 0.2776 & 0.0763 & 0.1970 & \textbf{0.3224} & 0.9426 & \textbf{0.9439} \\
    \midrule
    \multirow{7}{*}{\rotatebox[origin=c]{90}{\textbf{\textit{Subject}}}} & (a) no OCAL & 0.6730 & 0.1652 & 0.2532 & 0.0450 & 0.1232 & 0.3082 & \textbf{0.9414} & 0.9312 \\
     & (b) text-to-vid only & 0.6718 & 0.2332 & 0.2586 & 0.0643 & 0.1755 & 0.3134 & 0.9401 & 0.9315 \\
     & (c) vid-to-text only & 0.6680 & 0.2239 & \underline{0.2613} & 0.0623 & 0.1662 & 0.3122 & 0.9409 & 0.9322 \\
     & (d) (b) + (c) & \underline{0.6812} & 0.2367 & 0.2586 & 0.0652 & 0.1786 & 0.3144 & 0.9408 & 0.9311 \\
     & (e) (d) + size reg & 0.6811 & \textbf{0.2439} & 0.2586 & \textbf{0.0674} & \underline{0.1860} & \underline{0.3150} & 0.9407 & \underline{0.9326} \\
     & (f) (d) + rep & 0.6784 & 0.2367 & 0.2584 & 0.0651 & 0.1815 & \textbf{0.3151} & 0.9401 & 0.9284 \\
     & (g) (e) + rep (Ours, Full) & \textbf{0.6818} & \underline{0.2403} & \textbf{0.2617} & \underline{0.0667} & \textbf{0.1896} & \underline{0.3150} & \underline{0.9410} & \textbf{0.9335} \\
    \midrule
    \multirow{7}{*}{\rotatebox[origin=c]{90}{\textbf{\textit{Scene}}}} & (a) no OCAL & 0.6566 & 0.1442 & 0.2605 & 0.0397 & 0.1049 & 0.3153 & 0.9416 & 0.9296 \\
     & (b) text-to-vid only & 0.6716 & 0.2052 & 0.2623 & 0.0580 & 0.1520 & \underline{0.3175} & 0.9420 & 0.9343 \\
     & (c) vid-to-text only & 0.6602 & 0.1793 & 0.2621 & 0.0499 & 0.1331 & 0.3170 & 0.9421 & 0.9344 \\
     & (d) (b) + (c) & \underline{0.6762} & \underline{0.2087} & 0.2597 & 0.0582 & 0.1529 & 0.3164 & 0.9421 & 0.9348 \\
     & (e) (d) + size reg & 0.6655 & \textbf{0.2177} & \textbf{0.2654} & \textbf{0.0614} & \underline{0.1618} & \textbf{0.3189} & \textbf{0.9424} & \textbf{0.9374} \\
     & (f) (d) + rep & \textbf{0.6784} & 0.2041 & 0.2577 & 0.0566 & 0.1550 & 0.3160 & \underline{0.9422} & \underline{0.9365} \\
     & (g) (e) + rep (Ours, Full) & 0.6746 & 0.2073 & \underline{0.2637} & \underline{0.0584} & \textbf{0.1628} & 0.3173 & 0.9415 & 0.9362 \\
    \midrule
    \multirow{7}{*}{\rotatebox[origin=c]{90}{\textbf{\textit{All}}}} & (a) no OCAL & 0.6714 & 0.1652 & 0.2634 & 0.0462 & 0.1230 & 0.3125 & \underline{0.9419} & 0.9339 \\
     & (b) text-to-vid only & 0.6749 & 0.2328 & 0.2668 & 0.0659 & 0.1736 & 0.3172 & 0.9416 & 0.9363 \\
     & (c) vid-to-text only & 0.6675 & 0.2157 & 0.2662 & 0.0609 & 0.1605 & 0.3151 & \textbf{0.9420} & 0.9368 \\
     & (d) (b) + (c) & 0.6791 & \underline{0.2375} & 0.2654 & 0.0667 & 0.1778 & 0.3172 & \textbf{0.9420} & 0.9361 \\
     & (e) (d) + size reg & 0.6775 & \textbf{0.2475} & \underline{0.2672} & \textbf{0.0698} & \textbf{0.1866} & \underline{0.3177} & 0.9418 & \underline{0.9369} \\
     & (f) (d) + rep & \underline{0.6800} & 0.2345 & 0.2644 & 0.0656 & 0.1782 & 0.3176 & 0.9416 & 0.9349 \\
     & (g) (e) + rep (Ours, Full) & \textbf{0.6813} & 0.2371 & \textbf{0.2677} & \underline{0.0671} & \underline{0.1832} & \textbf{0.3182} & 0.9417 & \textbf{0.9379} \\
    \bottomrule
    \end{tabular}
\end{table}

\subsection{OCAL Ablation without FMS}
\label{app:ocal_no_fms}
Table~\mbox{%
\ref{tab:ablation_ocal_full} }\hskip0pt%
ablates OCAL on top of FMS, which compresses the differences between its components.
Table~\mbox{%
\ref{tab:ablation_ocal_no_fms} }\hskip0pt%
repeats the ablation without FMS, starting from the unguided backbone.
Either alignment direction alone recovers most of the grounding, and both together add a small margin: IoU 0.3108 against 0.2959 for text-to-video only ($p < 10^{-3}$ on IoU and OGS) and against 0.3027 for video-to-text only ($p = 0.21$ and $0.33$).
MF stays between 0.479 and 0.499 in every OCAL row: OCAL grounds, it does not move, which is why we never use it alone despite its higher IoU.
The effect of repetition depends on FMS.
Repetition tightens each target from the union of an object's masks over four frames to its per-frame mask, so there is motion to lock onto only when FMS supplies it: on top of the size regularizer, repetition wins motion back with FMS (MF 0.6775 to 0.6813, (e) to (g) in Table~\mbox{%
\ref{tab:ablation_ocal}}%
) and costs it without (MF 0.4989 to 0.4860, (e$'$) to (g$'$)).
The size regularizer without repetition grounds best in both settings.
None of these last steps is statistically significant, so the choice between (e) and (g) is a trade rather than a gain; we keep (g) as our default (Appx.~\mbox{%
\ref{app:ocal_ablation}}%
).

\begin{table}[!htb]
  \caption{
    \textbf{OCAL ablation without FMS.} The same components as Table~\ref{tab:ablation_ocal}, starting from the unguided backbone of Table~\ref{tab:ablation_main}-(a), so that the two alignment directions can be read on their own. One direction recovers most of the grounding gain and both together add a small margin, while MF stays between 0.479 and 0.499 in every OCAL row: OCAL grounds, it does not move. Best results are in \textbf{bold} and second best are \underline{underlined}.
  }
  \label{tab:ablation_ocal_no_fms}
  \centering
  \fontsize{8pt}{9pt}\selectfont
  \setlength{\tabcolsep}{4pt}

  \begin{tabular}{l*{8}{c}}
    \toprule
    \textbf{Method} & MF $\uparrow$ & IoU $\uparrow$ & LTA $\uparrow$ & OGS $\uparrow$ & MGS $\uparrow$ & GTA $\uparrow$ & CTC $\uparrow$ & DTC $\uparrow$ \\
    \midrule
    (a) Backbone & 0.4257 & 0.1593 & 0.2662 & 0.0446 & 0.1008 & 0.3177 & \underline{0.9430} & 0.9393 \\
    (b$'$) text-to-vid only & \underline{0.4966} & 0.2959 & 0.2682 & 0.0826 & 0.1958 & 0.3216 & 0.9425 & 0.9405 \\
    (c$'$) vid-to-text only & 0.4794 & 0.3027 & \textbf{0.2695} & 0.0854 & 0.2014 & 0.3201 & 0.9427 & \textbf{0.9435} \\
    (d$'$) text-to-vid + vid-to-text & 0.4911 & 0.3108 & \underline{0.2692} & \underline{0.0874} & 0.2076 & \textbf{0.3231} & 0.9425 & 0.9415 \\
    (e$'$) (d$'$) + size reg only & \textbf{0.4989} & \textbf{0.3138} & 0.2687 & \textbf{0.0882} & \textbf{0.2105} & 0.3223 & 0.9428 & 0.9407 \\
    (f$'$) (d$'$) + rep only & 0.4908 & \underline{0.3116} & 0.2686 & 0.0874 & \underline{0.2089} & 0.3209 & \textbf{0.9431} & \underline{0.9432} \\
    (g$'$) (e$'$) + rep (= OCAL only) & 0.4860 & 0.3099 & 0.2684 & 0.0868 & 0.2072 & \underline{0.3223} & 0.9430 & 0.9422 \\
    \bottomrule
  \end{tabular}

\end{table}

\clearpage
\section{Additional Ablation Studies}
\label{app:additional_ablations}

\subsection{Ablation study on $\lambda_{\text{FMS}}$}
We perform an ablation study on the weighting factor $\lambda_{\text{FMS}}$ to analyze its impact on motion fidelity and overall generation quality.
We report the quantitative results in Table \ref{tab:ablation_lambda_fms} below.
Increasing $\lambda_{\text{FMS}}$ improves motion fidelity (MF) up to $\lambda_{\text{FMS}}=1.25$, after which MF falls back to 0.6648 over \textit{\textbf{All}}.
At $\lambda_{\text{FMS}}=1.25$, our method achieves the best overall results across \textit{\textbf{Caption}}, \textit{\textbf{Subject}}, \textit{\textbf{Scene}}, and \textit{\textbf{All}} settings.
Both neighbouring settings are significantly worse in MF, IoU, and OGS.
IoU, LTA, and OGS also peak at $\lambda_{\text{FMS}}=1.25$ and differ by at most 0.010 over \textit{\textbf{All}} across the three settings, so FMS does not harm spatial or semantic alignment.
Temporal consistency metrics (CTC and DTC) vary minimally across settings, indicating that FMS preserves coherent motion.

\begin{table}[!htb]
\caption{
    \textbf{Ablation study on $\lambda_{\text{FMS}}$.}
    Setting $\lambda_{\text{FMS}}=1.25$ achieves the best overall performance, improving motion fidelity (MF) while preserving object grounding (highest IoU, LTA, OGS) and temporal consistency across different prompt settings.
  }
  \label{tab:ablation_lambda_fms}
  \centering
  \fontsize{8pt}{9pt}\selectfont
  \setlength{\tabcolsep}{4pt}
  \begin{tabular}{cl*{8}{c}}
    \toprule
    & \textbf{Setting} & MF $\uparrow$ & IoU $\uparrow$ & LTA $\uparrow$ & OGS $\uparrow$ & MGS $\uparrow$ & GTA $\uparrow$ & CTC $\uparrow$ & DTC $\uparrow$ \\
    \midrule
    \multirow{3}{*}{\rotatebox[origin=c]{90}{\textbf{\textit{Caption}}}} & (a) $\lambda_{\text{FMS}} = 1.00$ & 0.6681 & 0.2533 & \underline{0.2772} & \underline{0.0732} & \underline{0.1891} & \underline{0.3157} & 0.9419 & 0.9433 \\
     & (b) $\lambda_{\text{FMS}} = 1.25$ (Ours, Full) & \textbf{0.6875} & \textbf{0.2638} & \textbf{0.2776} & \textbf{0.0763} & \textbf{0.1970} & \textbf{0.3224} & \underline{0.9426} & \underline{0.9439} \\
     & (c) $\lambda_{\text{FMS}} = 1.50$ & \underline{0.6708} & \underline{0.2535} & 0.2747 & 0.0728 & 0.1887 & 0.3145 & \textbf{0.9427} & \textbf{0.9466} \\
    \midrule
    \multirow{3}{*}{\rotatebox[origin=c]{90}{\textbf{\textit{Subject}}}} & (a) $\lambda_{\text{FMS}} = 1.00$ & 0.6498 & 0.2232 & 0.2582 & \underline{0.0620} & 0.1638 & 0.3123 & \textbf{0.9410} & \textbf{0.9358} \\
     & (b) $\lambda_{\text{FMS}} = 1.25$ (Ours, Full) & \textbf{0.6818} & \textbf{0.2403} & \textbf{0.2617} & \textbf{0.0667} & \textbf{0.1896} & \textbf{0.3150} & \textbf{0.9410} & 0.9335 \\
     & (c) $\lambda_{\text{FMS}} = 1.50$ & \underline{0.6538} & \underline{0.2239} & \underline{0.2592} & 0.0616 & \underline{0.1643} & \underline{0.3126} & \underline{0.9406} & \underline{0.9345} \\
    \midrule
    \multirow{3}{*}{\rotatebox[origin=c]{90}{\textbf{\textit{Scene}}}} & (a) $\lambda_{\text{FMS}} = 1.00$ & \underline{0.6715} & \underline{0.2054} & \underline{0.2623} & \underline{0.0571} & \underline{0.1521} & \textbf{0.3186} & \underline{0.9427} & \underline{0.9359} \\
     & (b) $\lambda_{\text{FMS}} = 1.25$ (Ours, Full) & \textbf{0.6746} & \textbf{0.2073} & \textbf{0.2637} & \textbf{0.0584} & \textbf{0.1628} & 0.3173 & 0.9415 & \textbf{0.9362} \\
     & (c) $\lambda_{\text{FMS}} = 1.50$ & 0.6697 & 0.2048 & 0.2608 & 0.0565 & 0.1504 & \underline{0.3178} & \textbf{0.9428} & 0.9352 \\
    \midrule
    \multirow{3}{*}{\rotatebox[origin=c]{90}{\textbf{\textit{All}}}} & (a) $\lambda_{\text{FMS}} = 1.00$ & 0.6631 & 0.2273 & \underline{0.2659} & \underline{0.0641} & \underline{0.1683} & \underline{0.3155} & \underline{0.9419} & \underline{0.9383} \\
     & (b) $\lambda_{\text{FMS}} = 1.25$ (Ours, Full) & \textbf{0.6813} & \textbf{0.2371} & \textbf{0.2677} & \textbf{0.0671} & \textbf{0.1832} & \textbf{0.3182} & 0.9417 & 0.9379 \\
     & (c) $\lambda_{\text{FMS}} = 1.50$ & \underline{0.6648} & \underline{0.2274} & 0.2649 & 0.0636 & 0.1678 & 0.3150 & \textbf{0.9420} & \textbf{0.9388} \\
    \bottomrule
    \end{tabular}
\end{table}

\subsection{Ablation study on $\lambda_{\text{OCAL}}$}
We perform an ablation study on the weighting factor $\lambda_{\text{OCAL}}$ to analyze its impact on generation quality.
We report the quantitative results in Table \ref{tab:ablation_lambda_ocal} below.
Raising $\lambda_{\text{OCAL}}$ from 0.75 to 1.00 improves IoU, LTA, and OGS in every prompt setting (over \textit{\textbf{All}}, IoU from 0.2261 to 0.2371 and OGS from 0.0637 to 0.0671) and MF over \textit{\textbf{All}}.
Raising it further to 1.25 adds at most 0.004 IoU and 0.001 OGS in any setting, while LTA (0.2677 to 0.2661) and MF (0.6813 to 0.6758) fall over \textit{\textbf{All}}.
We therefore select $\lambda_{\text{OCAL}}=1.0$: it gives the highest MF across \textit{\textbf{Caption}}, \textit{\textbf{Subject}}, and \textit{\textbf{All}} settings and the highest LTA in every setting, and the IoU and OGS it gives up to 1.25 are within 0.004.

\begin{table}[!htb]
\caption{
        \textbf{Ablation study on $\lambda_{\text{OCAL}}$.} Raising $\lambda_{\text{OCAL}}$ from 0.75 to 1.00 improves IoU, LTA, and OGS in every prompt setting; 1.25 adds at most 0.004 IoU at lower MF and LTA. We choose 1.00, which gives the highest MF across \textit{\textbf{Caption}}, \textit{\textbf{Subject}}, and \textit{\textbf{All}} settings and the highest LTA in every setting.
    }
  \label{tab:ablation_lambda_ocal}
  \centering
  \fontsize{8pt}{9pt}\selectfont
  \setlength{\tabcolsep}{4pt}
  \begin{tabular}{cl*{8}{c}}
    \toprule
    & \textbf{Setting} & MF $\uparrow$ & IoU $\uparrow$ & LTA $\uparrow$ & OGS $\uparrow$ & MGS $\uparrow$ & GTA $\uparrow$ & CTC $\uparrow$ & DTC $\uparrow$ \\
    \midrule
    \multirow{3}{*}{\rotatebox[origin=c]{90}{\textbf{\textit{Caption}}}} & (a) $\lambda_{\text{OCAL}} = 0.75$ & \underline{0.6757} & 0.2576 & \underline{0.2775} & 0.0744 & 0.1928 & 0.3194 & 0.9422 & \underline{0.9425} \\
     & (b) $\lambda_{\text{OCAL}} = 1.00$ (Ours, Full) & \textbf{0.6875} & \underline{0.2638} & \textbf{0.2776} & \textbf{0.0763} & \underline{0.1970} & \textbf{0.3224} & \underline{0.9426} & \textbf{0.9439} \\
     & (c) $\lambda_{\text{OCAL}} = 1.25$ & 0.6748 & \textbf{0.2655} & 0.2754 & \underline{0.0759} & \textbf{0.2007} & \underline{0.3211} & \textbf{0.9427} & 0.9423 \\
    \midrule
    \multirow{3}{*}{\rotatebox[origin=c]{90}{\textbf{\textit{Subject}}}} & (a) $\lambda_{\text{OCAL}} = 0.75$ & 0.6712 & 0.2210 & 0.2581 & 0.0608 & 0.1665 & 0.3124 & \underline{0.9409} & 0.9321\\
     & (b) $\lambda_{\text{OCAL}} = 1.00$ (Ours, Full) & \textbf{0.6818} & \underline{0.2403} & \textbf{0.2617} & \underline{0.0667} & \textbf{0.1896} & \underline{0.3150} & \textbf{0.9410} & \textbf{0.9335} \\
     & (c) $\lambda_{\text{OCAL}} = 1.25$ & \underline{0.6769} & \textbf{0.2437} & \underline{0.2613} & \textbf{0.0673} & \underline{0.1822} & \textbf{0.3169} & 0.9402 & \underline{0.9331} \\
    \midrule
    \multirow{3}{*}{\rotatebox[origin=c]{90}{\textbf{\textit{Scene}}}} & (a) $\lambda_{\text{OCAL}} = 0.75$ & \underline{0.6751} & 0.1998 & 0.2604 & 0.0557 & 0.1490 & \underline{0.3167} & \underline{0.9419} & \textbf{0.9376} \\
     & (b) $\lambda_{\text{OCAL}} = 1.00$ (Ours, Full) & 0.6746 & \underline{0.2073} & \textbf{0.2637} & \underline{0.0584} & \textbf{0.1628} & \textbf{0.3173} & 0.9415 & \underline{0.9362} \\
     & (c) $\lambda_{\text{OCAL}} = 1.25$ & \textbf{0.6759} & \textbf{0.2093} & \underline{0.2617} & \textbf{0.0586} & \underline{0.1566} & 0.3152 & \textbf{0.9423} & 0.9356 \\
    \midrule
    \multirow{3}{*}{\rotatebox[origin=c]{90}{\textbf{\textit{All}}}} & (a) $\lambda_{\text{OCAL}} = 0.75$ & 0.6740 & 0.2261 & 0.2653 & 0.0637 & 0.1694 & 0.3162 & \textbf{0.9417} & \underline{0.9374} \\
     & (b) $\lambda_{\text{OCAL}} = 1.00$ (Ours, Full) & \textbf{0.6813} & \underline{0.2371} & \textbf{0.2677} & \underline{0.0671} & \textbf{0.1832} & \textbf{0.3182} & \textbf{0.9417} & \textbf{0.9379} \\
     & (c) $\lambda_{\text{OCAL}} = 1.25$ & \underline{0.6758} & \textbf{0.2395} & \underline{0.2661} & \textbf{0.0673} & \underline{0.1800} & \underline{0.3177} & \textbf{0.9417} & 0.9370\\
    \bottomrule
    \end{tabular}
\end{table}

\subsection{Ablation study on the number of FMS denoising steps}
We perform an ablation study on the number of early denoising steps $T_{opt,\text{FMS}}$ during which FMS is applied ($d$ in Appx.~\mbox{%
\ref{app:hyperparameters}}%
; 15 of 50 in our setting, with 5 optimization steps per denoising step throughout) to analyze its impact on generation quality.
We report the quantitative results in Table \ref{tab:ablation_fms_optim_steps} below.
MF rises steeply from 5 to 15 steps (0.5545 to 0.6813 over \textit{\textbf{All}}) and does not rise further at 20 steps (0.6783).
IoU and OGS move the other way: from 5 to 15 steps they fall from 0.2722 to 0.2371 and from 0.0761 to 0.0671 over \textit{\textbf{All}}, while MGS stays within 0.001 (0.1839 to 0.1832) because the higher object-level MF offsets the lower IoU, and LTA stays within 0.003 and GTA within 0.006.
With few FMS steps the objects stay where OCAL places them but barely move (MF 0.5545), the same competition between the two losses as in Table~\mbox{%
\ref{tab:ablation_main} }\hskip0pt%
and Appx.~\mbox{%
\ref{app:ocal_no_fms}}%
.
Temporal consistency metrics (CTC, DTC) remain largely stable across all settings.
We choose 15 steps, the highest MF over \textit{\textbf{All}}; its IoU and OGS are within 0.010 and 0.002 of the 10- and 20-step settings, and only the 5-step setting, whose motion is far weaker, grounds better.

\begin{table}[!htb]
\caption{
        \textbf{Ablation study on the number of FMS denoising steps.} Increasing $T_{opt,\text{FMS}}$ improves MF up to 15 steps, after which it plateaus. IoU and OGS fall from 5 to 15 steps (0.2722 to 0.2371 and 0.0761 to 0.0671 over \textit{\textbf{All}}), a trade-off between motion fidelity and grounding, while LTA, GTA, CTC, and DTC stay within 0.006.
    }
  \label{tab:ablation_fms_optim_steps}
  \centering
  \fontsize{8pt}{9pt}\selectfont
  \setlength{\tabcolsep}{4pt}
  \begin{tabular}{cl*{8}{c}}
    \toprule
    & \textbf{Setting} & MF $\uparrow$ & IoU $\uparrow$ & LTA $\uparrow$ & OGS $\uparrow$ & MGS $\uparrow$ & GTA $\uparrow$ & CTC $\uparrow$ & DTC $\uparrow$ \\
    \midrule
    \multirow{4}{*}{\rotatebox[origin=c]{90}{\textbf{\textit{Caption}}}} & (a) $T_{opt,\text{FMS}}=5$ & 0.5751 & \textbf{0.2823} & \textbf{0.2827} & \textbf{0.0820} & 0.1959 & \textbf{0.3231} & \underline{0.9425} & 0.9422 \\
     & (b) $T_{opt,\text{FMS}}=10$ & 0.6690 & \underline{0.2757} & \underline{0.2798} & \underline{0.0794} & \textbf{0.2015} & 0.3212 & 0.9419 & 0.9418 \\
     & (c) $T_{opt,\text{FMS}}=15$ (Ours, Full) & \textbf{0.6875} & 0.2638 & 0.2776 & 0.0763 & 0.1970 & \underline{0.3224} & \textbf{0.9426} & \textbf{0.9439} \\
     & (d) $T_{opt,\text{FMS}}=20$ & \underline{0.6805} & 0.2652 & 0.2781 & 0.0763 & \underline{0.2000} & 0.3196 & 0.9415 & \underline{0.9435} \\
    \midrule
    \multirow{4}{*}{\rotatebox[origin=c]{90}{\textbf{\textit{Subject}}}} & (a) $T_{opt,\text{FMS}}=5$ & 0.5473 & \textbf{0.2712} & \underline{0.2604} & \textbf{0.0744} & 0.1779 & \textbf{0.3222} & \underline{0.9405} & \textbf{0.9397} \\
     & (b) $T_{opt,\text{FMS}}=10$ & 0.6604 & 0.2362 & 0.2603 & 0.0651 & 0.1731 & \underline{0.3160} & 0.9400 & 0.9310 \\
     & (c) $T_{opt,\text{FMS}}=15$ (Ours, Full) & \textbf{0.6818} & \underline{0.2403} & \textbf{0.2617} & \underline{0.0667} & \textbf{0.1896} & 0.3150 & \textbf{0.9410} & \underline{0.9335} \\
     & (d) $T_{opt,\text{FMS}}=20$ & \underline{0.6779} & 0.2348 & 0.2592 & 0.0643 & \underline{0.1792} & 0.3149 & 0.9404 & 0.9316 \\
    \midrule
    \multirow{4}{*}{\rotatebox[origin=c]{90}{\textbf{\textit{Scene}}}} & (a) $T_{opt,\text{FMS}}=5$ & 0.5411 & \textbf{0.2630} & \underline{0.2634} & \textbf{0.0719} & \textbf{0.1778} & \textbf{0.3218} & \underline{0.9423} & \textbf{0.9368} \\
     & (b) $T_{opt,\text{FMS}}=10$ & 0.6563 & \underline{0.2262} & 0.2613 & \underline{0.0625} & \underline{0.1650} & \underline{0.3206} & \textbf{0.9426} & 0.9354 \\
     & (c) $T_{opt,\text{FMS}}=15$ (Ours, Full) & \underline{0.6746} & 0.2073 & \textbf{0.2637} & 0.0584 & 0.1628 & 0.3173 & 0.9415 & \underline{0.9362} \\
     & (d) $T_{opt,\text{FMS}}=20$ & \textbf{0.6766} & 0.2103 & 0.2621 & 0.0592 & 0.1589 & 0.3147 & 0.9419 & 0.9350 \\
    \midrule
    \multirow{4}{*}{\rotatebox[origin=c]{90}{\textbf{\textit{All}}}} & (a) $T_{opt,\text{FMS}}=5$ & 0.5545 & \textbf{0.2722} & \textbf{0.2688} & \textbf{0.0761} & \textbf{0.1839} & \textbf{0.3224} & \textbf{0.9418} & \textbf{0.9396} \\
     & (b) $T_{opt,\text{FMS}}=10$ & 0.6619 & \underline{0.2461} & 0.2671 & \underline{0.0690} & 0.1799 & \underline{0.3193} & 0.9415 & 0.9361 \\
     & (c) $T_{opt,\text{FMS}}=15$ (Ours, Full) & \textbf{0.6813} & 0.2371 & \underline{0.2677} & 0.0671 & \underline{0.1832} & 0.3182 & \underline{0.9417} & \underline{0.9379} \\
     & (d) $T_{opt,\text{FMS}}=20$ & \underline{0.6783} & 0.2367 & 0.2665 & 0.0666 & 0.1795 & 0.3164 & 0.9413 & 0.9367 \\
    \bottomrule
    \end{tabular}
\end{table}

\subsection{Ablation study on the number of OCAL denoising steps}
We perform an ablation study on the number of early denoising steps $T_{opt,\text{OCAL}}$ during which OCAL is applied (as for FMS, 15 of 50 in our setting) to analyze its impact on generation quality.
We report the quantitative results in Table \ref{tab:ablation_ocal_optim_steps} below.
Across the four settings, IoU, LTA, and OGS vary by at most 0.006, 0.002, and 0.002 over \textit{\textbf{All}}.
IoU and OGS are highest at 20 steps (0.2421 and 0.0679 over \textit{\textbf{All}}, 0.005 and 0.001 above 15 steps), while MF, LTA, and MGS are highest at 15 steps over \textit{\textbf{All}} (0.6813 and 0.2677) and fall at 20 (0.6736 and 0.2657).
We select $T_{opt,\text{OCAL}} = 15$, where MF and LTA peak over \textit{\textbf{All}}, since the IoU and OGS gained at 20 steps are within 0.005 and 0.001.

\begin{table}[!htb]
\caption{
        \textbf{Ablation study on the number of OCAL denoising steps.} We report performance as a function of $T_{opt, \text{OCAL}}$, showing that 15 steps provide the best overall trade-off, achieving the highest motion fidelity while maintaining stable object grounding and alignment.
    }
  \label{tab:ablation_ocal_optim_steps}
  \centering
  \fontsize{8pt}{9pt}\selectfont
  \setlength{\tabcolsep}{4pt}
  \begin{tabular}{cl*{8}{c}}
    \toprule
    & \textbf{Setting} & MF $\uparrow$ & IoU $\uparrow$ & LTA $\uparrow$ & OGS $\uparrow$ & MGS $\uparrow$ & GTA $\uparrow$ & CTC $\uparrow$ & DTC $\uparrow$ \\
    \midrule
    \multirow{4}{*}{\rotatebox[origin=c]{90}{\textbf{\textit{Caption}}}} & (a) $T_{opt,\text{OCAL}}=5$ & 0.6730 & \underline{0.2657} & \textbf{0.2790} & \underline{0.0766} & 0.1998 & 0.3213 & \underline{0.9424} & 0.9396 \\
     & (b) $T_{opt,\text{OCAL}}=10$ & \underline{0.6770} & 0.2655 & \textbf{0.2790} & 0.0764 & \underline{0.2001} & 0.3201 & 0.9422 & \underline{0.9407} \\
     & (c) $T_{opt,\text{OCAL}}=15$ (Ours, Full) & \textbf{0.6875} & 0.2638 & 0.2776 & 0.0763 & 0.1970 & \textbf{0.3224} & \textbf{0.9426} & \textbf{0.9439} \\
     & (d) $T_{opt,\text{OCAL}}=20$ & 0.6757 & \textbf{0.2695} & \underline{0.2779} & \textbf{0.0772} & \textbf{0.2025} & \underline{0.3222} & 0.9417 & \underline{0.9423} \\
    \midrule
    \multirow{4}{*}{\rotatebox[origin=c]{90}{\textbf{\textit{Subject}}}} & (a) $T_{opt,\text{OCAL}}=5$ & 0.6771 & \underline{0.2440} & \underline{0.2607} & \textbf{0.0674} & 0.1833 & \textbf{0.3179} & 0.9404 & 0.9283 \\
     & (b) $T_{opt,\text{OCAL}}=10$ & \underline{0.6781} & 0.2400 & \underline{0.2607} & 0.0665 & 0.1804 & \underline{0.3167} & \underline{0.9407} & \underline{0.9290} \\
     & (c) $T_{opt,\text{OCAL}}=15$ (Ours, Full) & \textbf{0.6818} & 0.2403 & \textbf{0.2617} & 0.0667 & \textbf{0.1896} & 0.3150 & \textbf{0.9410} & \textbf{0.9335} \\
     & (d) $T_{opt,\text{OCAL}}=20$ & 0.6762 & \textbf{0.2444} & 0.2578 & \underline{0.0673} & \underline{0.1840} & 0.3146 & 0.9406 & 0.9285 \\
    \midrule
    \multirow{4}{*}{\rotatebox[origin=c]{90}{\textbf{\textit{Scene}}}} & (a) $T_{opt,\text{OCAL}}=5$ & \underline{0.6761} & \underline{0.2091} & 0.2594 & 0.0582 & 0.1564 & 0.3150 & 0.9412 & 0.9358 \\
     & (b) $T_{opt,\text{OCAL}}=10$ & \textbf{0.6802} & 0.2051 & 0.2588 & 0.0570 & 0.1555 & \underline{0.3164} & \textbf{0.9425} & \textbf{0.9367} \\
     & (c) $T_{opt,\text{OCAL}}=15$ (Ours, Full) & 0.6746 & 0.2073 & \textbf{0.2637} & \underline{0.0584} & \textbf{0.1628} & \textbf{0.3173} & \underline{0.9415} & \underline{0.9362} \\
     & (d) $T_{opt,\text{OCAL}}=20$ & 0.6689 & \textbf{0.2124} & \underline{0.2613} & \textbf{0.0592} & \underline{0.1585} & 0.3156 & \underline{0.9415} & 0.9351 \\
    \midrule
    \multirow{4}{*}{\rotatebox[origin=c]{90}{\textbf{\textit{All}}}} & (a) $T_{opt,\text{OCAL}}=5$ & 0.6754 & \underline{0.2396} & \underline{0.2663} & \underline{0.0674} & 0.1799 & \underline{0.3181} & 0.9414 & 0.9346 \\
     & (b) $T_{opt,\text{OCAL}}=10$ & \underline{0.6784} & 0.2369 & 0.2662 & 0.0666 & 0.1788 & 0.3177 & \textbf{0.9418} & \underline{0.9355} \\
     & (c) $T_{opt,\text{OCAL}}=15$ (Ours, Full) & \textbf{0.6813} & 0.2371 & \textbf{0.2677} & 0.0671 & \textbf{0.1832} & \textbf{0.3182} & \underline{0.9417} & \textbf{0.9379} \\
     & (d) $T_{opt,\text{OCAL}}=20$ & 0.6736 & \textbf{0.2421} & 0.2657 & \textbf{0.0679} & \underline{0.1818} & 0.3175 & 0.9413 & 0.9354 \\
    \bottomrule
    \end{tabular}
\end{table}

\subsection{Ablation study on OCAL block position}
We perform an ablation study on the OCAL block position parameter $B$ to analyze its effect on motion fidelity and object grounding.
We report the quantitative results in Table \ref{tab:ablation_ocal_block_position} below.
$B=10$ gives the highest MF across \textit{\textbf{Caption}}, \textit{\textbf{Subject}}, and \textit{\textbf{All}} settings, and its IoU, LTA, and OGS are within 0.002 of the best setting ($B=15$) over \textit{\textbf{All}}, with stable CTC and DTC.
At $B=5$, MF stays close (0.6737 over \textit{\textbf{All}}) but grounding falls to the lowest values in the table (IoU 0.2108, OGS 0.0593).
At $B=15$, IoU, LTA, and OGS stay within 0.002 of $B=10$ over \textit{\textbf{All}} while MF falls to 0.6687; at $B=20$ both fall (MF 0.6648, IoU 0.2246, OGS 0.0632), so applying OCAL late costs motion first and then grounding.
Based on these observations, we choose $B=10$ as it provides the best trade-off between high motion fidelity, accurate object grounding, and temporal consistency.

\begin{table}[!htb]
\caption{
    \textbf{Ablation study on OCAL block position.}
    Setting $B=10$ achieves the best overall trade-off, yielding the highest motion fidelity while preserving strong object grounding and stable temporal consistency across different prompt settings.
  }
  \label{tab:ablation_ocal_block_position}
  \centering
  \fontsize{8pt}{9pt}\selectfont
  \setlength{\tabcolsep}{4pt}
  \begin{tabular}{cl*{8}{c}}
    \toprule
    & \textbf{Setting} & MF $\uparrow$ & IoU $\uparrow$ & LTA $\uparrow$ & OGS $\uparrow$ & MGS $\uparrow$ & GTA $\uparrow$ & CTC $\uparrow$ & DTC $\uparrow$ \\
    \midrule
    \multirow{4}{*}{\rotatebox[origin=c]{90}{\textbf{\textit{Caption}}}} & (a) $B=5$ & \underline{0.6785} & 0.2305 & 0.2768 & 0.0669 & 0.1720 & 0.3164 & 0.9422 & 0.9419 \\
     & (b) $B=10$ (Ours, Full) & \textbf{0.6875} & \textbf{0.2638} & \underline{0.2776} & \textbf{0.0763} & \textbf{0.1970} & \textbf{0.3224} & \underline{0.9426} & \underline{0.9439} \\
     & (c) $B=15$ & 0.6739 & \underline{0.2560} & \textbf{0.2795} & \underline{0.0744} & 0.1920 & \underline{0.3196} & 0.9422 & \underline{0.9439} \\
     & (d) $B=20$ & 0.6746 & 0.2532 & 0.2751 & 0.0732 & \underline{0.1925} & 0.3140 & \textbf{0.9428} & \textbf{0.9453} \\
    \midrule
    \multirow{4}{*}{\rotatebox[origin=c]{90}{\textbf{\textit{Subject}}}} & (a) $B=5$ & \underline{0.6662} & 0.2166 & 0.2604 & 0.0601 & 0.1627 & \underline{0.3144} & 0.9408 & 0.9308 \\
     & (b) $B=10$ (Ours, Full) & \textbf{0.6818} & \textbf{0.2403} & \underline{0.2617} & \textbf{0.0667} & \textbf{0.1896} & \textbf{0.3150} & 0.9410 & 0.9335 \\
     & (c) $B=15$ & 0.6558 & \underline{0.2369} & \textbf{0.2626} & \underline{0.0658} & \underline{0.1763} & 0.3130 & \textbf{0.9414} & \underline{0.9356} \\
     & (d) $B=20$ & 0.6502 & 0.2231 & 0.2598 & 0.0617 & 0.1648 & 0.3140 & \underline{0.9411} & \textbf{0.9358} \\
    \midrule
    \multirow{4}{*}{\rotatebox[origin=c]{90}{\textbf{\textit{Scene}}}} & (a) $B=5$ & \textbf{0.6765} & 0.1854 & 0.2588 & 0.0509 & 0.1373 & 0.3169 & \underline{0.9424} & \underline{0.9373} \\
     & (b) $B=10$ (Ours, Full) & \underline{0.6746} & \underline{0.2073} & \textbf{0.2637} & \underline{0.0584} & \underline{0.1628} & 0.3173 & 0.9415 & 0.9362 \\
     & (c) $B=15$ & \textbf{0.6765} & \textbf{0.2223} & \underline{0.2617} & \textbf{0.0617} & \textbf{0.1677} & \textbf{0.3198} & 0.9416 & \textbf{0.9410} \\
     & (d) $B=20$ & 0.6697 & 0.1974 & 0.2607 & 0.0547 & 0.1463 & \underline{0.3184} & \textbf{0.9432} & 0.9342 \\
    \midrule
    \multirow{4}{*}{\rotatebox[origin=c]{90}{\textbf{\textit{All}}}} & (a) $B=5$ & \underline{0.6737} & 0.2108 & 0.2653 & 0.0593 & 0.1575 & 0.3159 & \underline{0.9418} & 0.9367 \\
     & (b) $B=10$ (Ours, Full) & \textbf{0.6813} & \underline{0.2371} & \underline{0.2677} & \underline{0.0671} & \textbf{0.1832} & \textbf{0.3182} & 0.9417 & 0.9379 \\
     & (c) $B=15$ & 0.6687 & \textbf{0.2384} & \textbf{0.2679} & \textbf{0.0673} & \underline{0.1787} & \underline{0.3174} & 0.9417 & \textbf{0.9402} \\
     & (d) $B=20$ & 0.6648 & 0.2246 & 0.2652 & 0.0632 & 0.1677 & 0.3155 & \textbf{0.9424} & \underline{0.9384} \\
    \bottomrule
    \end{tabular}
\end{table}

\subsection{Size Regularizer under Zoom}
\label{app:size_reg_zoom}
Our size regularizer $\lambda_{size}^{(i)}$ in our OCAL sums over the whole clip and is therefore one scalar per object: large for a small object and small for a large one, so that small objects are not drowned out during optimization.
Being a single number per clip, it cannot track an object that grows or shrinks over time; the per-frame masks of our OCAL do that.
We verify this on the four DAVIS videos with the largest within-clip change in ground-truth object area (rallye, motocross-jump, mbike-trick, and car-turn, with area changes of $155\times$, $20\times$, $13.6\times$, and $8.9\times$), generating each with our method with and without $\lambda_{size}$ under the three prompts.
The correlation over time between the generated and the source object area, averaged over the four videos and three prompts, is 0.68 with $\lambda_{size}$ and 0.70 without, which is indistinguishable, as the definition predicts.
Size tracking itself is strong even under extreme zoom (0.91 on rallye), but it is independent of $\lambda_{size}$.

\clearpage
\section{Generalization to CogVideoX-2B and Wan2.1}
\label{app:generalization}
To demonstrate the model-agnostic nature and zero-shot compatibility of our MotionGrounder, we extend it to two additional architectures: the (1) CogVideoX-2B \citep{cogvideox2025yang} and the 1.3B variant of (2) Wan2.1 \citep{wan2025wanopenadvancedlargescale}.

For the CogVideoX-2B architecture, we integrate FMS into transformer block $B=15$ and OCAL into the transformer block $B=5$ with corresponding loss weights set to $\lambda_\text{FMS}=0.5$ and $\lambda_\text{OCAL}=0.75$.
For the WAN architecture, we integrate FMS into the self-attention layer of transformer block $B=10$ and OCAL into the cross-attention layer of block $B=5$ with corresponding loss weights set to $\lambda_\text{FMS}=2.0$ and $\lambda_\text{OCAL}=1.0$.
Since Wan2.1 conditions on text through cross-attention, there is no video-to-text direction and OCAL runs in the text-to-video direction only.

For both architectures, we activate FMS and OCAL during the first 30\% of the denoising process, where we perform 5 optimization steps per denoising step.
During optimization, we employ a linearly decayed learning rate, decreasing from 0.002 to 0.001.
Additionally, the temperature parameter in Eq.~\ref{eqn:cross_frame_attn} is set to $\tau=2$.

Quantitative comparisons are provided in Table~\ref{tab:generalization}, showing performance gains over the base CogVideoX-2B and Wan2.1 backbones.
As shown in Table~\ref{tab:generalization}, the integration of our MotionGrounder yields a substantial improvement in MF across all prompt categories for both architectures.
Specifically, the overall MF score for CogVideoX-2B increases from 0.4024 to 0.5807 while Wan2.1 increases from 0.4623 to 0.5881, a gain that holds in every prompt setting.
This indicates that FMS transfers the dynamic characteristics of the reference video to the latent space.
Our full method also raises IoU and OGS, by 6\% and 7\% on CogVideoX-2B and by 14\% and 15\% on Wan2.1-1.3B, while LTA stays within 0.005 of the base models over All videos.
GTA, CTC, and DTC stay within 0.012 of the base models on both backbones, so motion is transferred without visible flicker or loss of the backbones' generative quality.

\begin{figure}[!htb]
    \centering
    \includegraphics[width=0.95\linewidth]{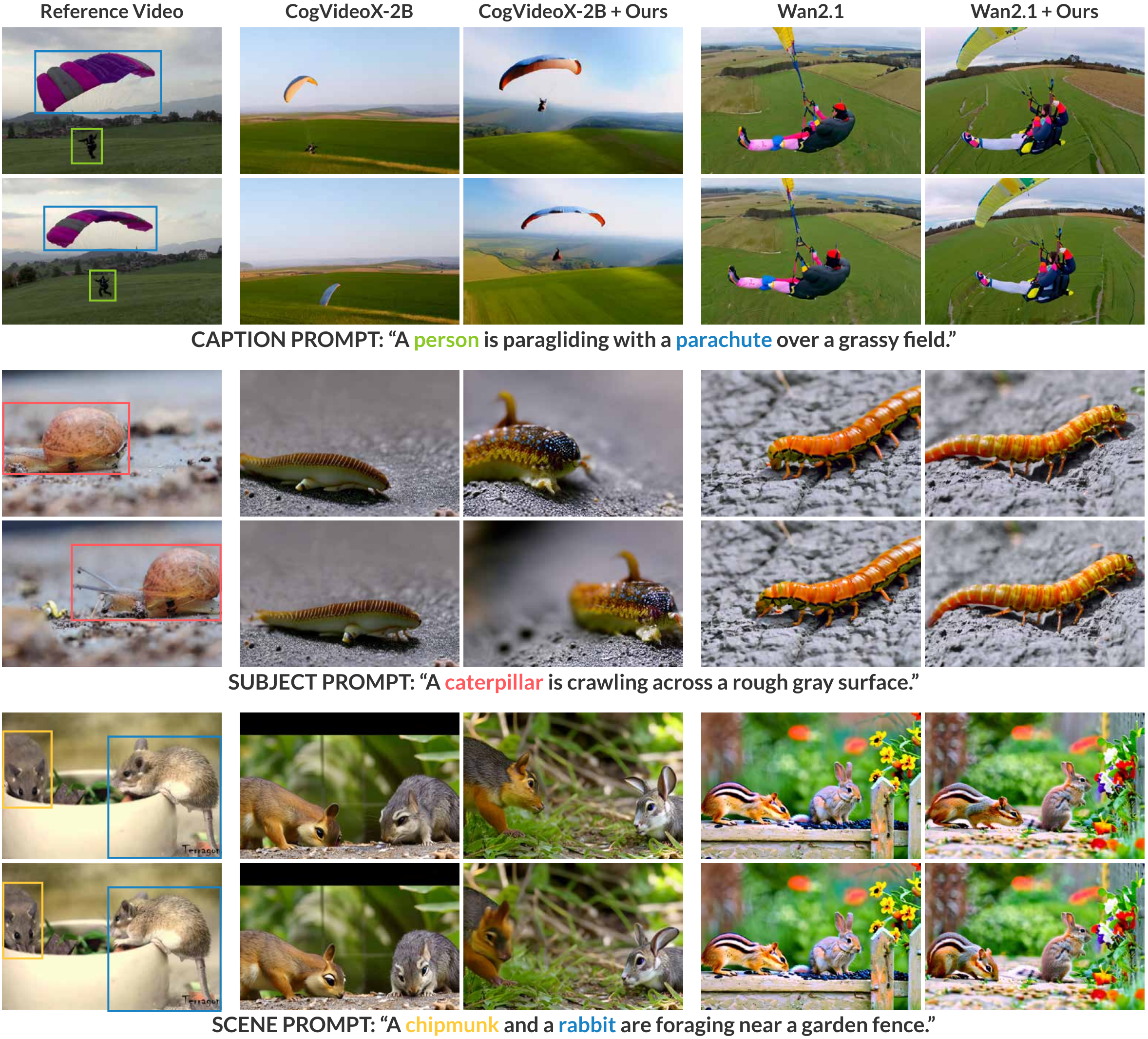}
\vspace{-7.6pt}
    \caption{
        \textbf{Qualitative generalization results on CogVideoX-2B and Wan2.1 backbones.} Compared to the original backbones, our MotionGrounder produces motion trajectories that better follow the reference video while following the target captions.
    }
    \label{fig:generalization}
\end{figure}

Fig.~\ref{fig:generalization} presents qualitative results of integrating our MotionGrounder to CogVideoX-2B and Wan2.1.
Compared to the original backbones, our MotionGrounder produces motion trajectories that more closely follow the reference video while keeping the target appearance.
These results agree with the MF gains in Table~\mbox{%
\ref{tab:generalization}}%
.
The grounding gain of our method depends on the backbone: in IoU, from the base model to ours, Wan2.1-1.3B goes from 0.1662 to 0.1894 and CogVideoX-2B from 0.1763 to 0.1868, whereas CogVideoX-5B goes from 0.1593 to 0.2371 (Table~\mbox{%
\ref{tab:ablation_main}}%
), where OCAL runs in both attention directions.

\begin{table}[!htb]
\caption{
        \textbf{Quantitative generalization results on CogVideoX-2B and Wan2.1 backbones.} We integrate MotionGrounder into the CogVideoX-2B \citep{cogvideox2025yang} and Wan2.1-1.3B \citep{wan2025wanopenadvancedlargescale} architectures to evaluate its zero-shot generalization capabilities. Our method raises MF on both backbones and raises IoU and OGS, while LTA, GTA, CTC, and DTC stay within 0.012 of the base models. The better of each pair of rows is in \textbf{bold}.
    }
  \label{tab:generalization}
  \centering
  \fontsize{8pt}{9pt}\selectfont
  \setlength{\tabcolsep}{4pt}
  \begin{tabular}{cl*{7}{c}}
    \toprule
    & \textbf{Model} & MF $\uparrow$ & IoU $\uparrow$ & LTA $\uparrow$ & OGS $\uparrow$ & GTA $\uparrow$ & CTC $\uparrow$ & DTC $\uparrow$ \\
    \midrule
    \multirow{4}{*}{\rotatebox[origin=c]{90}{\textbf{\textit{Caption}}}} & CogVideoX-2B & \underline{0.4141} & \underline{0.2006} & \textbf{0.2782} & \underline{0.0573} & \textbf{0.3212} & \textbf{0.9444} & \textbf{0.9301} \\
     & CogVideoX-2B + Ours & \textbf{0.5683} & \textbf{0.2020} & \underline{0.2766} & \textbf{0.0582} & \underline{0.3167} & \underline{0.9404} & \underline{0.9231} \\
    \cmidrule(lr){2-8}
     & Wan2.1 & \underline{0.4661} & \underline{0.1867} & \textbf{0.2790} & \underline{0.0543} & \textbf{0.3230} & \textbf{0.9443} & \textbf{0.9615} \\
     & Wan2.1 + Ours & \textbf{0.6050} & \textbf{0.2045} & \underline{0.2786} & \textbf{0.0592} & \underline{0.3191} & \underline{0.9406} & \underline{0.9565} \\
    \midrule
    \multirow{4}{*}{\rotatebox[origin=c]{90}{\textbf{\textit{Subject}}}} & CogVideoX-2B & \underline{0.3952} & \underline{0.1768} & \textbf{0.2611} & \underline{0.0480} & \textbf{0.3157} & \textbf{0.9423} & \textbf{0.9209} \\
     & CogVideoX-2B + Ours & \textbf{0.5900} & \textbf{0.1814} & \underline{0.2542} & \textbf{0.0488} & \underline{0.3137} & \underline{0.9394} & \underline{0.9134} \\
    \cmidrule(lr){2-8}
     & Wan2.1 & \underline{0.4632} & \underline{0.1642} & \underline{0.2670} & \underline{0.0453} & \underline{0.3267} & \textbf{0.9459} & \textbf{0.9662} \\
     & Wan2.1 + Ours & \textbf{0.5829} & \textbf{0.1972} & \textbf{0.2704} & \textbf{0.0555} & \textbf{0.3305} & \underline{0.9398} & \underline{0.9467} \\
    \midrule
    \multirow{4}{*}{\rotatebox[origin=c]{90}{\textbf{\textit{Scene}}}} & CogVideoX-2B & \underline{0.3980} & \underline{0.1515} & \textbf{0.2579} & \underline{0.0404} & \textbf{0.3189} & \textbf{0.9432} & \textbf{0.9266} \\
     & CogVideoX-2B + Ours & \textbf{0.5837} & \textbf{0.1770} & \underline{0.2528} & \textbf{0.0488} & \underline{0.3152} & \underline{0.9410} & \underline{0.9183} \\
    \cmidrule(lr){2-8}
     & Wan2.1 & \underline{0.4574} & \underline{0.1476} & \underline{0.2640} & \underline{0.0404} & \underline{0.3257} & \textbf{0.9453} & \textbf{0.9658} \\
     & Wan2.1 + Ours & \textbf{0.5764} & \textbf{0.1666} & \textbf{0.2656} & \textbf{0.0461} & \textbf{0.3308} & \underline{0.9413} & \underline{0.9542} \\
    \midrule
    \multirow{4}{*}{\rotatebox[origin=c]{90}{\textbf{\textit{All}}}} & CogVideoX-2B & \underline{0.4024} & \underline{0.1763} & \textbf{0.2657} & \underline{0.0486} & \textbf{0.3186} & \textbf{0.9433} & \textbf{0.9259} \\
     & CogVideoX-2B + Ours & \textbf{0.5807} & \textbf{0.1868} & \underline{0.2612} & \textbf{0.0519} & \underline{0.3152} & \underline{0.9403} & \underline{0.9183} \\
    \cmidrule(lr){2-8}
     & Wan2.1 & \underline{0.4623} & \underline{0.1662} & \underline{0.2700} & \underline{0.0466} & \underline{0.3251} & \textbf{0.9452} & \textbf{0.9645} \\
     & Wan2.1 + Ours & \textbf{0.5881} & \textbf{0.1894} & \textbf{0.2715} & \textbf{0.0536} & \textbf{0.3268} & \underline{0.9406} & \underline{0.9524} \\
    \bottomrule
    \end{tabular}
\end{table}

\section{Evaluation on the DAVIS Benchmark of DiTFlow}
\label{app:davis}
To test generalization beyond our benchmark, we evaluate on the DAVIS subset released by DiTFlow \mbox{%
\citep{ditflow2025pondaven}}%
, using its prompts.
We use the 43 sequences with complete ground-truth masks, which IoU and OGS require, and 32 of them are not in our benchmark.
All methods use the hyperparameters of Appx.~\mbox{%
\ref{app:hyperparameters}}%
.

Table~\mbox{%
\ref{tab:davis} }\hskip0pt%
shows that our MotionGrounder achieves the best MF, IoU, and MGS.
Our MF lead over DMT is narrow (0.7615 vs.\ 0.7609 with ground-truth masks).
MotionClone obtains a higher OGS (0.0870 vs.\ 0.0854, or 0.0860 with SAM3.1 masks), but a much lower MF (0.6311 vs.\ 0.7615).
Our LTA (0.2713) stays at the backbone's level (0.2711), slightly below the baselines (0.2720 to 0.2755).
With masks generated automatically by SAM3.1 \mbox{%
\citep{sam32026carion}}%
, our results match those with ground-truth masks.

The component rows follow the pattern of Table~\mbox{%
\ref{tab:ablation_main}}%
.
FMS alone improves motion (MF 0.4959 to 0.7440) but barely changes grounding.
OCAL alone improves grounding (IoU 0.1858 to 0.4035) but transfers little motion (MF 0.5841).
Only their combination improves both.

This benchmark is mostly single-object: 40 of the 43 sequences have one annotated object.
It therefore tests generalization to data we did not design, while our multi-object claim rests on the by-count results in Table~\mbox{%
\ref{tab:main_quantitative_comparison}}%
.

\begin{table}[!htb]
  \caption{\textbf{Quantitative comparison on the DAVIS benchmark of DiTFlow} (43 sequences with complete ground-truth masks).
    Best results among the compared methods are in \textbf{bold} and second best are \underline{underlined}.
    Component rows are shown for reference.}
  \label{tab:davis}
  \centering
  \fontsize{8pt}{9pt}\selectfont
  \setlength{\tabcolsep}{2pt}

\begin{tabular}{l*{8}{c}}
    \toprule
    \textbf{Method} & MF $\uparrow$ & IoU $\uparrow$ & LTA $\uparrow$ & OGS $\uparrow$ & MGS $\uparrow$ & GTA $\uparrow$ & CTC $\uparrow$ & DTC $\uparrow$ \\
    \midrule
    DMT & 0.7609 & 0.2632 & \textbf{0.2755} & 0.0757 & 0.2021 & 0.3243 & 0.9341 & 0.9196 \\
    MOFT & 0.6878 & 0.2772 & 0.2720 & 0.0793 & 0.2088 & 0.3223 & 0.9361 & 0.9076 \\
    MotionClone & 0.6311 & 0.2993 & \underline{0.2751} & \textbf{0.0870} & 0.2246 & \underline{0.3247} & \textbf{0.9388} & \textbf{0.9360} \\
    ConMo & 0.7281 & 0.2372 & 0.2734 & 0.0685 & 0.1867 & \textbf{0.3252} & 0.9355 & 0.9206 \\
    DiTFlow & 0.7484 & 0.2866 & 0.2745 & 0.0822 & 0.2161 & 0.3245 & \underline{0.9376} & 0.9248 \\
    \midrule
    \textit{Backbone only} & 0.4959 & 0.1858 & 0.2711 & 0.0538 & 0.1242 & 0.3226 & 0.9445 & 0.9339 \\
    \textit{+ FMS only} & 0.7440 & 0.2303 & 0.2658 & 0.0655 & 0.1759 & 0.3188 & 0.9407 & 0.9372 \\
    \textit{+ OCAL only} & 0.5841 & 0.4035 & 0.2783 & 0.1164 & 0.2868 & 0.3227 & 0.9430 & 0.9454 \\
    \midrule
    Ours (GT masks) & \underline{0.7615} & \underline{0.2999} & 0.2713 & 0.0854 & \underline{0.2304} & 0.3236 & 0.9360 & \underline{0.9274} \\
    Ours (SAM3.1 masks) & \textbf{0.7617} & \textbf{0.3007} & 0.2719 & \underline{0.0860} & \textbf{0.2341} & 0.3239 & 0.9368 & 0.9265 \\
    \bottomrule
  \end{tabular}

\end{table}

\newpage
\section{Comparison with FlowMotion}
\label{app:flowmotion}
FlowMotion \mbox{%
\citep{flowmotion2026wang} }\hskip0pt%
is a training-free method for flow-matching T2V models and the only one of MotionFlow, MultiMotion, and FlowMotion with public code, so it is the one we compare against.
Since it is built for Wan2.1-1.3B and CogVideoX-5B is not a flow-matching model, we compare on Wan2.1-1.3B, running the released FlowMotion code and our method with the configuration of Appx.~\mbox{%
\ref{app:generalization}}%
.
Shared sampling settings are matched to FlowMotion's defaults (21 frames, 50 denoising steps, the same seed, captions, and videos, a classifier-free guidance scale of 6, and an empty negative prompt), which our method was not tuned for.

Over the full benchmark, ours reaches MF 0.5888 / IoU 0.1921 / OGS 0.0544 against 0.5603 / 0.2303 / 0.0628 for FlowMotion, and a DTC of 0.9473 against 0.8874; Table~\mbox{%
\ref{tab:flowmotion} }\hskip0pt%
gives the split by object count.
We lead motion fidelity and temporal consistency, and FlowMotion leads grounding on this backbone; the ungrounded object-level MF of the two is close (0.356 against 0.339 for ours), so the MGS gap comes from the IoU gap, not from how objects move.

The gap is backbone-specific.
Wan2.1 conditions on text through cross-attention, so there is no video-to-text direction and OCAL runs in the text-to-video direction only (Appx.~\mbox{%
\ref{app:generalization}}%
).
Our full method lifts IoU by 14\% on Wan2.1-1.3B (0.1662 to 0.1894, Table~\ref{tab:generalization}) against 49\% on CogVideoX-5B (0.1593 to 0.2371, Table~\mbox{%
\ref{tab:ablation_main}}%
), where our grounding claim rests, and by 6\% on the bidirectional CogVideoX-2B (Table~\mbox{%
\ref{tab:generalization}}%
), so the gap is not explained by the missing direction alone (Appx.~\mbox{%
\ref{app:ocal_no_fms}}%
).

\begin{table}[!htb]
  \caption{
    \textbf{Comparison with FlowMotion on Wan2.1-1.3B, by object count.} Both methods run on the same backbone with matched sampling settings (21 frames, 50 steps, same seed, captions, and videos). The better of the two is in \textbf{bold}. We lead MF at three of four counts and LTA, GTA, CTC, and DTC at every count, while FlowMotion leads IoU and OGS.
  }
  \label{tab:flowmotion}
  \centering
  \fontsize{8pt}{9pt}\selectfont
  \setlength{\tabcolsep}{2pt}

  \begin{tabular}{lc*{8}{c}}
    \toprule
    \textbf{Method} & $N$ (\# of videos) & MF $\uparrow$ & IoU $\uparrow$ & LTA $\uparrow$ & OGS $\uparrow$ & MGS $\uparrow$ & GTA $\uparrow$ & CTC $\uparrow$ & DTC $\uparrow$ \\
    \midrule
    FlowMotion & $1$ (21) & 0.5573 & \textbf{0.2782} & 0.2636 & \textbf{0.0774} & \textbf{0.1987} & 0.3125 & 0.9227 & 0.8757 \\
    Ours & & \textbf{0.5910} & 0.2481 & \textbf{0.2742} & 0.0719 & 0.1743 & \textbf{0.3256} & \textbf{0.9404} & \textbf{0.9527} \\
    \midrule
    FlowMotion & $2$ (16) & \textbf{0.6117} & \textbf{0.2724} & 0.2632 & \textbf{0.0732} & \textbf{0.1993} & 0.3189 & 0.9249 & 0.8927 \\
    Ours & & 0.6066 & 0.1939 & \textbf{0.2697} & 0.0544 & 0.1383 & \textbf{0.3302} & \textbf{0.9376} & \textbf{0.9423} \\
    \midrule
    FlowMotion & $3$ (10) & 0.5071 & \textbf{0.1471} & 0.2611 & \textbf{0.0392} & 0.1000 & 0.3121 & 0.9178 & 0.8895 \\
    Ours & & \textbf{0.5798} & 0.1442 & \textbf{0.2665} & 0.0388 & \textbf{0.1028} & \textbf{0.3243} & \textbf{0.9396} & \textbf{0.9506} \\
    \midrule
    FlowMotion & ${\geq}4$ (5) & 0.5144 & \textbf{0.0613} & 0.2494 & \textbf{0.0158} & \textbf{0.0412} & 0.3246 & 0.9301 & 0.9150 \\
    Ours & & \textbf{0.5404} & 0.0470 & \textbf{0.2580} & 0.0122 & 0.0320 & \textbf{0.3464} & \textbf{0.9354} & \textbf{0.9344} \\
    \bottomrule
  \end{tabular}

\end{table}

\clearpage
\section{MotionGrounder with Automatically Generated Masks}
\label{app:noisy_masks}
To evaluate the robustness of our MotionGrounder to noisy automatically generated masks, we assess its performance using masks produced by off-the-shelf segmentation models.
Specifically, we generate object masks using SAM-Track \citep{samtrack2023cheng} and SAM3 \citep{sam32026carion}, and conduct experiments on the entire dataset using the hyperparameters described in Sec.~\ref{sec:experiments}.

Table~\ref{tab:noisy_masks} presents the quantitative results obtained using both clean and noisy masks, alongside the performance of other baseline methods.
For the noisy mask results, we additionally report the percentage difference relative to the clean mask scores below the corresponding noisy mask scores.
Table~\ref{tab:noisy_masks} reveals the following:

\begin{itemize}
    \item First, the overall degradation caused by noisy masks is minimal.
    Most metrics decrease by less than 4\%, with several metrics exhibiting negligible changes below 1\%.
    On average, SAM-Track and SAM3 only incur performance drops of -0.64\% and -1.15\%, respectively, across the seven metrics other than MGS.
    MGS drops by 4.14\% and 5.40\%, while object-level MF barely changes ($+2.07\%$ and $-0.29\%$), so the drop reflects placement rather than motion.
    Interestingly, some metrics slightly improve when using noisy masks, such as MF for SAM-Track (+0.01\%) and CTC for SAM3 (+0.04\%).
    This suggests that our MotionGrounder does not heavily depend on accurate masks and generalizes well to realistic segmentation outputs.
    \item Second, the results reveal that motion quality is largely preserved under noisy mask conditions.
    Both SAM-Track and SAM3 variants maintain MF scores of at least 0.6755, CTC scores of at least 0.9414, and DTC scores of at least 0.9350, remaining competitive with or superior to prior methods.
    This indicates that our MotionGrounder can preserve motion fidelity and temporal consistency even when object localization is imperfect.
    \item Third, our MotionGrounder outperforms prior methods in every grounding metric even under noisy masks.
    Using SAM-Track and SAM3 masks, our method still achieves the best, second-best, and third-best scores in IoU, LTA, OGS, and MGS, indicating that our proposed grounding mechanism, OCAL, remains effective despite imperfect masks.
\end{itemize}

\begin{table}[!htb]
  \caption{
    \textbf{Quantitative comparison using automatically generated masks.}
    Our MotionGrounder remains robust under noisy masks generated by off-the-shelf segmentation models, maintaining strong motion fidelity, grounding accuracy, and temporal consistency with only minor performance degradation compared to using clean masks.
    The best, second-best, and third-best scores are in \textbf{bold}, \underline{underlined}, and \textit{italic}, respectively.
  }
  \label{tab:noisy_masks}
  \centering
  \fontsize{8pt}{9pt}\selectfont
  \setlength{\tabcolsep}{2pt}
  \begin{tabular}{l*{8}{c}}
    \toprule
    \textbf{Method} (\textbf{\textit{All}}) & MF $\uparrow$ & IoU $\uparrow$ & LTA $\uparrow$ & OGS $\uparrow$ & MGS $\uparrow$ & GTA $\uparrow$ & CTC $\uparrow$ & DTC $\uparrow$ \\
    \midrule
    DMT & \textbf{0.6886} & 0.2057 & 0.2656 & 0.0571 & 0.1574 & 0.3155 & 0.9348 & 0.9057 \\
    MOFT & 0.5450 & 0.1988 & 0.2630 & 0.0553 & 0.1385 & 0.3138 & 0.9387 & 0.8907 \\
    MotionClone & 0.5526 & 0.2081 & 0.2646 & 0.0581 & 0.1463 & 0.3127 & 0.9392 & 0.9345 \\
    ConMo & 0.6503 & 0.1967 & 0.2642 & 0.0553 & 0.1483 & 0.3168 & 0.9366 & 0.8982 \\
    DiTFlow & 0.6637 & 0.1868 & 0.2657 & 0.0520 & 0.1361 & 0.3156 & 0.9399 & 0.9259 \\
    \midrule
    Ours (Clean Masks) & \textit{0.6813} & \textbf{0.2371} & \textbf{0.2677} & \textbf{0.0671} & \textbf{0.1832} & \textbf{0.3182} & \underline{0.9417} & \textbf{0.9379} \\
    Ours (SAM-Track) & \underline{0.6814} & \underline{0.2337} & \underline{0.2666} & \underline{0.0658} & \underline{0.1757} & \textit{0.3171} & \textit{0.9414} & \textit{0.9350} \\
    \quad vs.\ clean masks (\%) & $+0.01$ & $-1.47$ & $-0.39$ & $-1.95$ & $-4.14$ & $-0.35$ & $-0.03$ & $-0.30$ \\
    Ours (SAM3) & 0.6755 & \textit{0.2305} & \textit{0.2659} & \textit{0.0648} & \textit{0.1733} & \underline{0.3177} & \textbf{0.9421} & \underline{0.9365} \\
    \quad vs.\ clean masks (\%) & $-0.86$ & $-2.80$ & $-0.65$ & $-3.51$ & $-5.40$ & $-0.16$ & $+0.04$ & $-0.15$ \\
    \bottomrule
  \end{tabular}
\end{table}

\begin{figure}[!htb]
    \centering
    \includegraphics[width=0.8\linewidth]{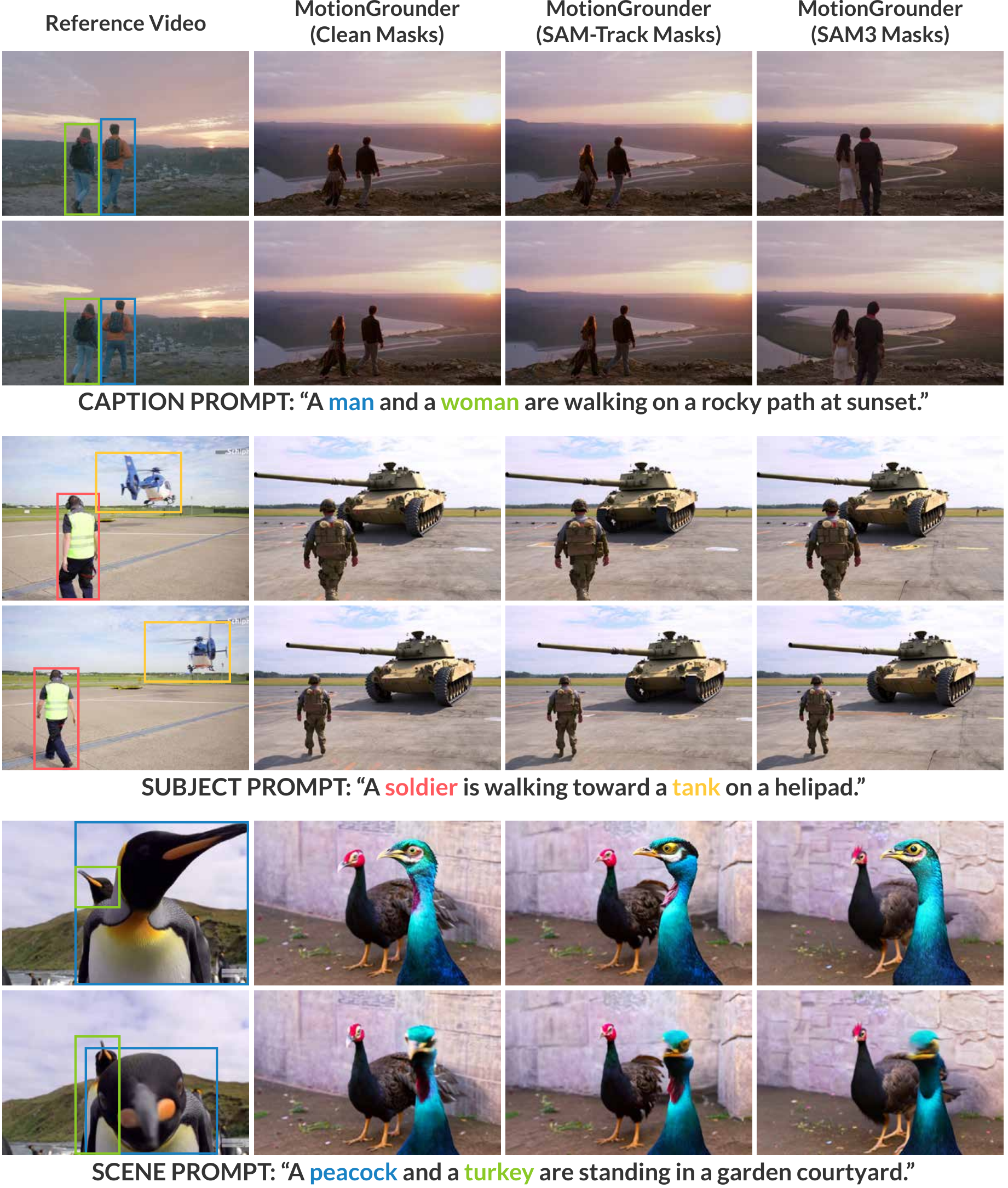}
\vspace{-8.0pt}
    \caption{
        \textbf{Qualitative comparison using automatically generated masks.} Our MotionGrounder produces visually consistent results across clean, SAM-Track, and SAM3 masks, with only minor differences in appearance and motion transfer quality despite segmentation noise.
    }
    \label{fig:noisy_masks}
\end{figure}

Fig.~\ref{fig:noisy_masks} provides a qualitative comparison between results obtained using clean and noisy masks.
As shown in Fig.~\ref{fig:noisy_masks}, the overall appearance and scene composition remain consistent when using SAM-Track and SAM3 masks instead of clean masks.
The generated videos preserve subject identity and spatial layout across all settings, indicating that our MotionGrounder is robust to segmentation noise in practical scenarios.
Although slight variations occasionally appear, these differences have minimal impact on the overall visual quality and motion transfer performance.
Overall, our quantitative and qualitative results suggest that our MotionGrounder is robust to segmentation noise and can be reliably applied in practical settings where manually annotated masks are unavailable.

\clearpage
\section{Runtime and GPU Memory Comparison}
\label{app:runtime}

We report total runtime (seconds), decomposed into motion signal preparation, optimization, and remaining denoising steps, along with GPU memory usage (GB) in Table~\ref{tab:runtime_and_memory}, averaged over three runs of one 24-frame $480 \times 720$ video with 50 denoising steps on an NVIDIA A6000 GPU.

\textbf{Fast motion signal preparation.}
Our Flow-based Motion Signal (FMS) achieves the fastest motion signal preparation among all compared methods.
This efficiency arises from constructing motion cues directly from optical flows without requiring inversion or denoising with the DiT backbone.
As a result, our MotionGrounder with FMS alone achieves runtime comparable to existing baselines while avoiding the expensive signal extraction procedures used by DMT, MOFT, MotionClone, ConMo, and DiTFlow.

\textbf{Minimal OCAL overhead.}
Adding Object-Caption Alignment Loss (OCAL) introduces explicit object-level control and grounding with only a negligible increase in total runtime.
Specifically, our full method requires $\sim 493$ s compared to $\sim 490$ s for FMS only, indicating that the additional optimization cost remains negligible despite the improved grounding capability.
Overall, our MotionGrounder keeps a runtime comparable to prior methods while leading IoU, LTA, and OGS over \textit{\textbf{All videos}} (Table~\ref{tab:main_quantitative_comparison} in the main paper).

\textbf{Scalability to number of objects.}
Our MotionGrounder maintains nearly identical runtime and memory usage as the number of objects increases (from 1 to 6).
This demonstrates that OCAL enables fine-grained multi-object grounding without introducing noticeable per-object computational growth, even with object-level attention refinement and optimization.
In contrast, existing methods rely on a single global caption and do not provide explicit multi-object control.

\begin{table}[!htb]
  \caption{
        \textbf{Runtime and memory usage comparison.} We report total runtime (seconds) decomposed into motion signal preparation, optimization, and remaining denoising steps, along with GPU memory usage (GB). Our MotionGrounder matches the runtime of the optimization-based baselines while leading every metric except MF over \textit{\textbf{All videos}} (Table~\mbox{%
\ref{tab:main_quantitative_comparison}}\hskip0pt%
).
    }
  \label{tab:runtime_and_memory}
  \centering
  \fontsize{8pt}{9pt}\selectfont
  \setlength{\tabcolsep}{4pt}
    \begin{tabular}{l*{5}{c}}
      \toprule
      \multirow{2}{*}{\textbf{Method}} & Signal & Optimi- & Rest of & Total & Memory \\
      & Prep.\ $\downarrow$ & zation $\downarrow$ & Steps $\downarrow$ & Runtime $\downarrow$ & (GB) $\downarrow$ \\

      \midrule

      CogVideoX-5B \citep{cogvideox2025yang}
        & 0.00 & 0.00 & 182.96 & 182.96 & 30.97 \\
      DMT \citep{dmt2023yatim}
        & 5.31 & 342.24 & 130.62 & 478.17 & 30.97 \\
      MOFT \citep{moft2024xiao}
        & 7.56 & 342.89 & 130.06 & 480.51 & 30.97 \\
      MotionClone \citep{motionclone2024ling}
        & 5.51 & 496.28 & 130.46 & 632.25 & 37.42 \\
      ConMo \citep{conmo2025gao}
        & 5.28 & 346.21 & 131.11 & 482.59 & 30.97 \\
      DiTFlow \citep{ditflow2025pondaven}
        & 5.87 & 352.52 & 130.49 & 488.88 & 37.42 \\

      \midrule

      Ours, FMS Only - 1 Object
        & 0.86 & 358.96 & 130.93 & 490.75 & 32.17 \\
      Ours, OCAL Only - 1 Object
        & 0.00 & 245.94 & 130.17 & 376.11 & 30.97 \\
      Ours, Full - 1 Object
        & 0.88 & 361.21 & 131.58 & 493.67 & 32.17 \\
      Ours, Full - 2 Objects
        & 0.81 & 361.02 & 131.26 & 493.09 & 32.17 \\
      Ours, Full - 3 Objects
        & 0.71 & 360.64 & 131.00 & 492.35 & 32.17 \\
      Ours, Full - 4 Objects
        & 0.79 & 360.64 & 130.98 & 492.41 & 32.17 \\
      Ours, Full - 5 Objects
        & 0.85 & 360.75 & 130.94 & 492.53 & 32.17 \\
      Ours, Full - 6 Objects
        & 0.78 & 360.89 & 131.00 & 492.67 & 32.17 \\

      \bottomrule
    \end{tabular}
\end{table}

\clearpage
\section{Limitations and Future Work}
\label{app:limitation}
\begin{figure}[!htb]
    \centering
    \includegraphics[width=1.0\linewidth]{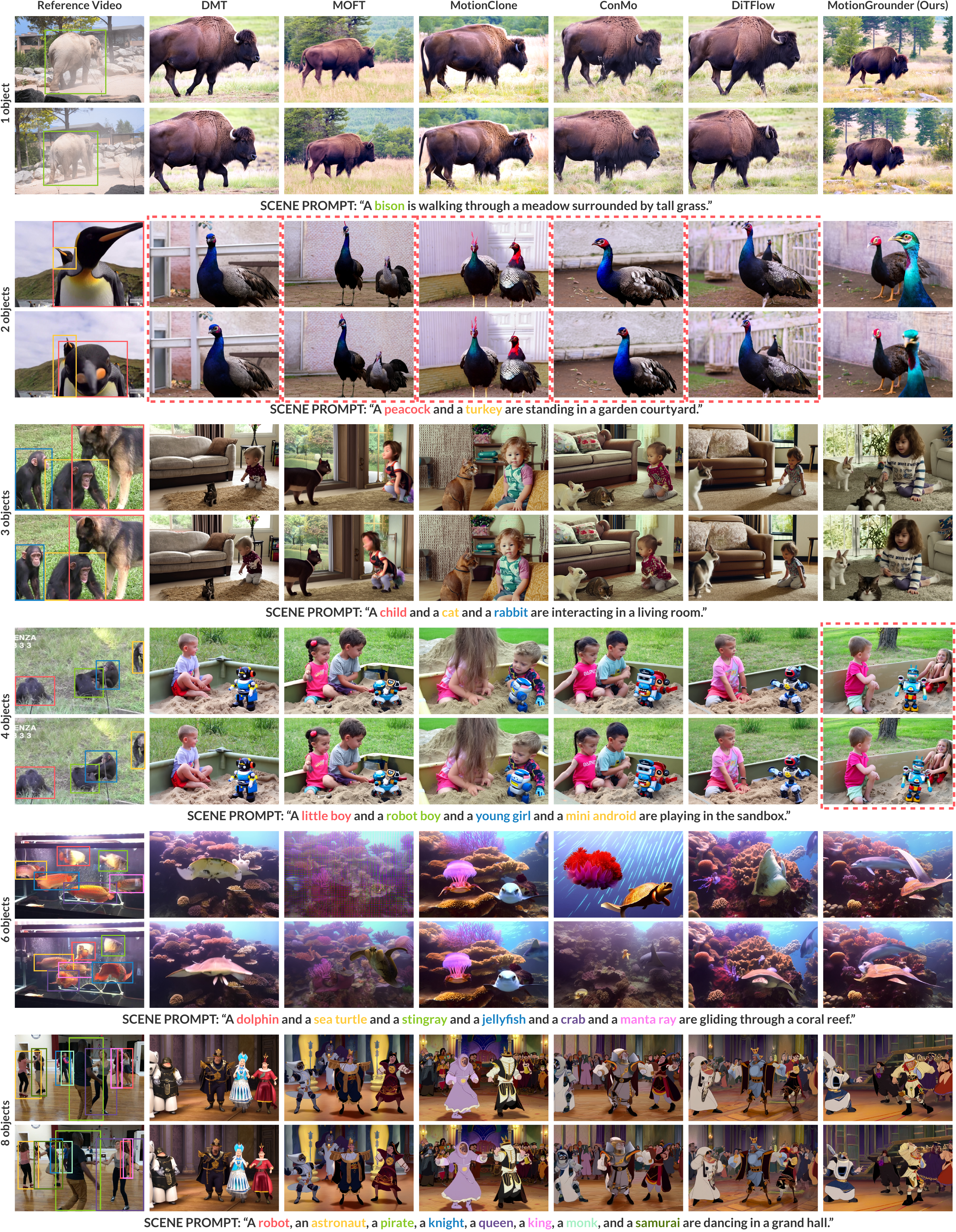}
\vspace{-18.7pt}
    \caption{
        \textbf{Limitations.} An excessive number of objects leads to degraded object consistency and weaker motion--object alignment in our MotionGrounder, a challenge also observed in existing motion transfer methods. Red dashed lines mark where each method first fails.
    }
    \label{fig:supp_limitations}
\end{figure}

\textbf{Number of objects.}
Fig.~\ref{fig:supp_limitations} examines robustness under increasing scene complexity by progressively adding more objects to the target caption.
Existing methods already fail in simple multi-object scenarios with only two objects, often missing specified entities or producing spatial misalignment where objects are generated in incorrect target regions.
The points at which these failures first occur are marked by red dashed lines in Fig.~\ref{fig:supp_limitations}.
In contrast, our MotionGrounder remains effective with up to three objects, maintaining coherent motion transfer and consistent object--caption grounding.
When four or more objects are present, performance degrades as dense object interactions increase ambiguity in motion assignment and grounding.
Beyond four objects, all methods, including our MotionGrounder, fail, indicating a shared limitation rather than one specific to our method.

\textbf{Grounding at high counts.}
Fig.~\mbox{%
\ref{fig:supp_limitations} }\hskip0pt%
also compares the methods at the highest object counts, and the numbers agree with it: at $N{\geq}3$ we lead every object-grounding metric (IoU, OGS, and MGS) against the best baseline at that count (IoU 0.1428 vs.\ 0.1354, OGS 0.0392 vs.\ 0.0360, MGS 0.1095 vs.\ 0.1011; Table~\mbox{%
\ref{tab:per_count}}%
). In the four-object row, we generate three of the four objects, losing the mini android, and each of the three is placed where it belongs; MOFT, MotionClone, and ConMo also generate three of the four, yet none of them grounds those objects correctly, while DMT and DiTFlow generate only two. What separates the methods there is not how many objects appear but whether each one lands where its caption says it should.

\textbf{Latent resolution.}
The object-count ceiling arises from the DiT-based LDM formulation, where input frames are heavily compressed by the VAE into low-resolution latents that are further patchified and processed at limited spatial resolution by the DiT.
Consequently, small objects and their corresponding masks are often lost or poorly represented, hindering accurate generation and control.
Operating in the pixel domain or in higher-resolution latent spaces could mitigate this issue, but at the cost of increased computational complexity.
Overall, these results highlight that scaling grounded motion transfer to highly cluttered, multi-object scenes remains an open research challenge.

\textbf{Motion granularity.}
FMS, like the attention-based motion signals we compare against, supervises trajectories on the $30 \times 45$ latent patch grid, so at $480 \times 720$ one supervised unit spans roughly $16 \times 16$ pixels.
Motion finer than a patch, such as the dog's tongue licking the chimpanzee in the second column of Fig.~\mbox{%
\ref{fig:main_figure}}%
, cannot be transferred by any of these signals.
What transfers is each object's coarse path and articulated motion at the scale of limbs and body.

\textbf{Distance between source and target.}
Scores drift as the target prompt departs from the source: from the \textit{\textbf{Caption}} to the \textit{\textbf{Subject}} to the \textit{\textbf{Scene}} setting, our MF falls from 0.6875 to 0.6818 to 0.6746 and our OGS from 0.0763 to 0.0667 to 0.0584 (Tables~\mbox{%
\ref{tab:per_prompt_caption}}\hskip0pt%
--\mbox{%
\ref{tab:per_prompt_scene}}%
, Appx.~\mbox{%
\ref{app:per_count}}%
).
The further the target object departs from the source in size, shape, and motion pattern, the more its path deviates, since neither loss enforces the original shape.
Every baseline drifts the same way, and our IoU and OGS remain the highest in each setting (Tables~\mbox{%
\ref{tab:per_prompt_caption}}\hskip0pt%
--\mbox{%
\ref{tab:per_prompt_scene}}%
), so the drift is a property of the task rather than of our method.
It is also the price of the shape flexibility shown in Appx.~\mbox{%
\ref{app:shape}}%
.

\textbf{Source motion.}
Our OGS falls as source object motion rises (Spearman $\rho = -0.43$, $p = 0.002$ over the 52 videos; Appx.~\mbox{%
\ref{app:near_static}}%
), so our grounding advantage is largest on sources with less object motion and shrinks as motion grows; the baselines show no such correlation because their grounding is uniformly lower.

\textbf{Backbone dependence.}
The grounding gain of our method depends on the backbone (Appx.~\mbox{%
\ref{app:generalization}}%
): our full method lifts IoU by 49\% on CogVideoX-5B, by 6\% on the bidirectional CogVideoX-2B, and by 14\% on Wan2.1-1.3B, where OCAL runs in the text-to-video direction only.
On Wan2.1-1.3B, FlowMotion \mbox{%
\citep{flowmotion2026wang} }\hskip0pt%
grounds better than ours while we retain higher motion fidelity and temporal consistency (Table~\mbox{%
\ref{tab:flowmotion}}%
, Appx.~\mbox{%
\ref{app:flowmotion}}%
).

\textbf{External models.}
Our MotionGrounder also relies on external optical flow and mask models.
Automatically generated masks change every score by at most 5.4\% (Table~\mbox{%
\ref{tab:noisy_masks}}%
, Appx.~\mbox{%
\ref{app:noisy_masks}}%
) and match ground-truth masks on DAVIS (Table~\mbox{%
\ref{tab:davis}}%
, Appx.~\mbox{%
\ref{app:davis}}%
).

\textbf{Future work.}
We see the open problem less as designing a better loss than as choosing the granularity of motion control deliberately: a patch-level signal carries coarse trajectories, whereas a finer, part-aware signal would be needed to carry articulation.

\clearpage
\section{User Study}
\label{app:user_study}
\subsection{Protocol}
\begin{figure}[!htb]
    \centering
    \includegraphics[width=0.98\linewidth]{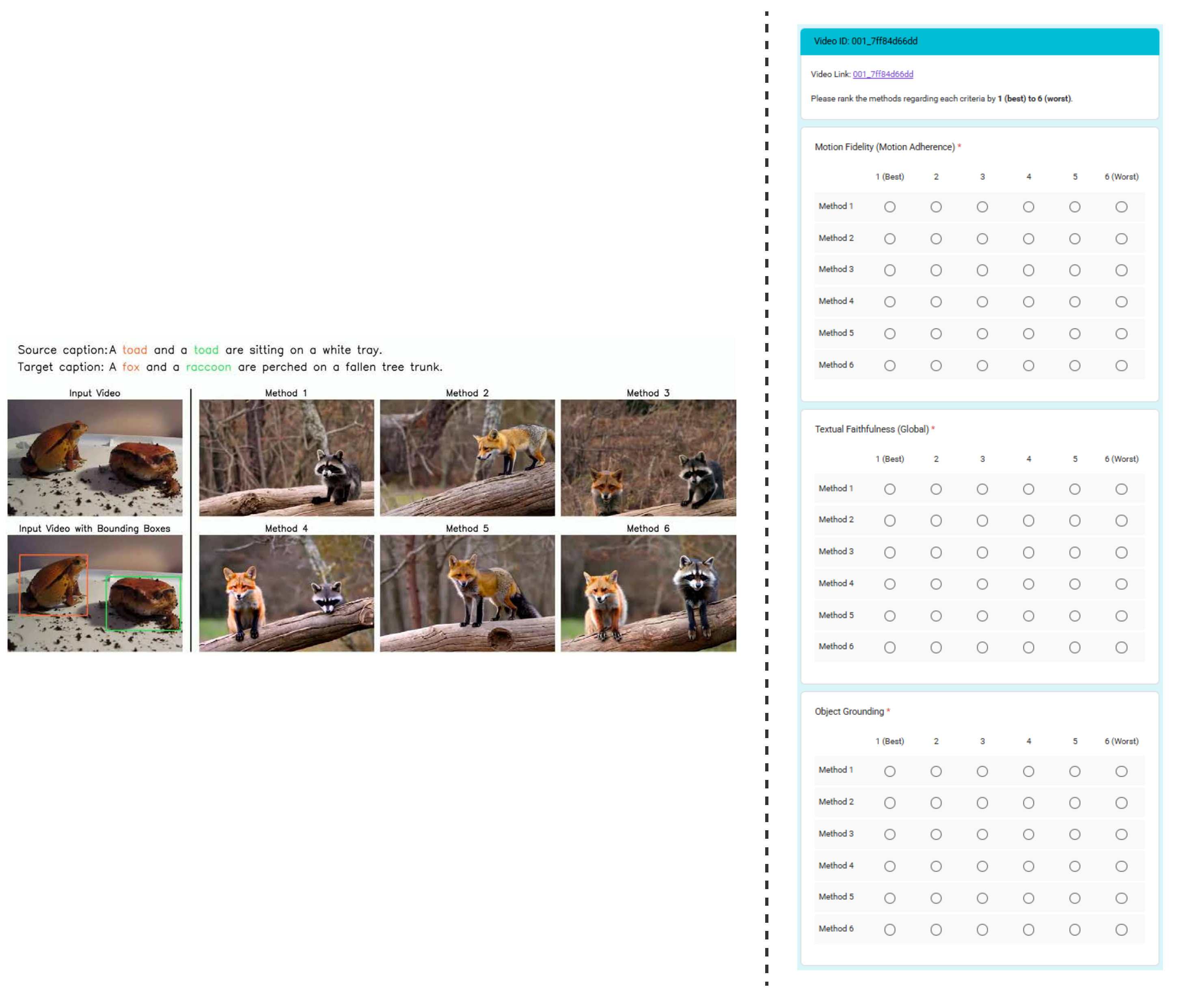}
\vspace{-8.3pt}
    \caption{
        Our user study interface and questionnaire form.
    }
    \label{fig:supp_user_study}
\end{figure}

For our user study, we select 50 generated videos from our method along with baseline results from DMT \citep{dmt2023yatim}, MOFT \citep{moft2024xiao}, MotionClone \citep{motionclone2024ling}, ConMo \citep{conmo2025gao}, and DiTFlow \citep{ditflow2025pondaven}, covering scenes from the \textit{Caption}, \textit{Subject}, and \textit{Scene} prompt settings with varying numbers of objects.
We ask 49 participants to rank the videos according to:
\begin{itemize}
    \item \textbf{Motion Adherence (MA)}: Rank the videos according to how naturally they replicate the reference motion;
    \item \textbf{Global Textual Faithfulness (GTF)}: Rank the videos that best reflect the overall target caption, considering the global scene and content rather than individual object placement; and
    \item \textbf{Object Grounding (OG)}: Rank the videos for object grounding accuracy, i.e., correctly generating each target object in its proper location relative to the source video.
\end{itemize}
Before starting, participants are shown examples of high- and low-quality videos for each criterion to guide their evaluations.
Fig.~\ref{fig:supp_user_study} presents our user study interface and questionnaire form.
The rankings split by object count are reported in Table~\mbox{%
\ref{tab:user_study_per_count} }\hskip0pt%
of the main paper.

\subsection{Statistical Analyses}
\textbf{Mean Rank.}
The human evaluation results in Table \ref{tab:user_study_statistical_analyses} demonstrate that our MotionGrounder consistently achieves the highest performance across all three evaluation dimensions.
In terms of Mean Rank, our method significantly outperforms all baselines, notably reaching a sub-3.0 rank in every category.
This consistent ranking suggests that human raters frequently identify MotionGrounder as the primary or secondary choice, while other baselines are clustered in the 3.5 to 3.8 range, struggling to maintain consistent top-tier placement.

\textbf{Borda Count.}
The Borda Count analysis further validates the consistent preference for our MotionGrounder across all evaluation dimensions.
Unlike the Mean Rank, which only reflects average ordinal placement, the Borda Count captures the overall consensus strength of human judgments by rewarding methods that are consistently preferred across comparisons.
As shown in Table~\ref{tab:user_study_statistical_analyses}, our MotionGrounder achieves the highest Borda Count in MA (3.01), GTF (3.03), and OG (3.17), substantially outperforming all baselines.
In contrast, the second-best method, DiTFlow, attains notably lower scores of 2.60, 2.60, and 2.54, respectively.
The particularly large margin in OG aligns with our superior OGS rank, indicating that users consistently favor our method for preserving accurate object-level motion associations.
These results demonstrate that our MotionGrounder not only achieves better average rankings but also exhibits stronger consensus preference across evaluators.

\begin{table}[!htb]
  \caption{
    \textbf{User study results.} We evaluate human preference across three dimensions: Motion Adherence (MA), Global Textual Faithfulness (GTF), and Object Grounding (OG). We report the Object Grounding Score (OGS) Rank alongside the Mean Rank ($\downarrow$) and Average Borda Count ($\uparrow$) to demonstrate consistent superiority across both ordinal and consensus-based metrics.
  }
  \label{tab:user_study_statistical_analyses}
  \centering
  \fontsize{8pt}{9pt}\selectfont
  \setlength{\tabcolsep}{4pt}
    \begin{tabular}{lcccccccc}
      \toprule

      \multirow{2}{*}{\textbf{Methods}} & \multicolumn{3}{c}{\textbf{Mean Rank} $\downarrow$} & \textbf{OGS} & \multicolumn{3}{c}{\textbf{Borda Count} $\uparrow$} \\
      \cmidrule(lr){2-4} \cmidrule(lr){6-8}
      & MA & GTF & OG & \textbf{Rank} $\downarrow$ & MA & GTF & OG \\

      \midrule

      DMT \citep{dmt2023yatim}
        & 3.65 & 3.70 & 3.72 & 3.68 & 2.35 & 2.30 & 2.28 \\
      MOFT \citep{moft2024xiao}
        & 3.81 & 3.78 & 3.81 & 3.74 & 2.19 & 2.22 & 2.19 \\
      MotionClone \citep{motionclone2024ling}
        & 3.59 & 3.56 & 3.59 & 3.61 & 2.41 & 2.44 & 2.41 \\
      ConMo \citep{conmo2025gao}
        & 3.56 & 3.59 & 3.58 & \underline{3.52} & 2.44 & 2.41 & 2.42 \\
      DiTFlow \citep{ditflow2025pondaven}
        & \underline{3.40} & \underline{3.40} & \underline{3.46} & 3.75 & \underline{2.60} & \underline{2.60} & \underline{2.54} \\
      MotionGrounder (Ours)
        & \textbf{2.99} & \textbf{2.97} & \textbf{2.83} & \textbf{2.70} & \textbf{3.01} & \textbf{3.03} & \textbf{3.17} \\
      \bottomrule
    \end{tabular}
\end{table}

\end{document}